\documentclass[conference]{IEEEtran}

\usepackage{times}

\usepackage{multicol}
\usepackage[bookmarks=true,colorlinks=true,urlcolor=red,pagebackref]{hyperref}

\newcommand\inputpgf[2]{{
\let\pgfimageWithoutPath\pgfimage
\renewcommand{\pgfimage}[2][]{\pgfimageWithoutPath[##1]{#1/##2}}
\input{#1/#2}
}}

\usepackage{tabularx}
\newcolumntype{Y}{>{\centering\arraybackslash}X} %

\usepackage{caption}

\usepackage{comment}
\usepackage{siunitx}
\usepackage{relsize}
\usepackage{ifthen}
\usepackage[colorinlistoftodos]{todonotes}

\usepackage[caption=false]{subfig}

\usepackage[vlined,ruled,linesnumbered]{algorithm2e}
\usepackage{graphics} %
\usepackage{rotating}
\usepackage{color}
\usepackage{enumerate}
\usepackage[T1]{fontenc}
\usepackage{psfrag}
\usepackage{epsfig} %
\usepackage{booktabs}
\usepackage{graphicx,url}
\usepackage{multirow}
\usepackage{array}
\usepackage{latexsym}
\usepackage{amsfonts}
\usepackage{amsmath}
\usepackage{amssymb}
\usepackage{xstring}
\usepackage[noend]{algorithmic}
\usepackage{multirow}
\usepackage{xcolor}
\usepackage{prettyref}
\usepackage{flexisym}
\usepackage{bigdelim}
\usepackage{breqn} %
\usepackage{listings}
\let\labelindent\relax
\usepackage{enumitem}
\usepackage{xspace}
\usepackage{bm}
\graphicspath{{./figures/}}
\usepackage{tikz}
\usetikzlibrary{matrix,calc}

\usepackage{mdwlist}

\let\itemize\undefined
\makecompactlist{itemize}{stditemize}

\usepackage{amsthm,amsmath}
\newtheoremstyle{mystyle}%
  {}%
  {}%
  {\itshape}%
  {}%
  {\bfseries}%
  {.}%
  { }%
  {\thmname{#1}\thmnumber{ #2}\thmnote{ (#3)}}%
\theoremstyle{mystyle}

\newrefformat{prob}{Problem\,\ref{#1}}
\newrefformat{def}{Definition\,\ref{#1}}
\newrefformat{sec}{Section\,\ref{#1}}
\newrefformat{sub}{Section\,\ref{#1}}
\newrefformat{prop}{Proposition\,\ref{#1}}
\newrefformat{app}{Appendix\,\ref{#1}}
\newrefformat{alg}{Algorithm\,\ref{#1}}
\newrefformat{cor}{Corollary\,\ref{#1}}
\newrefformat{thm}{Theorem\,\ref{#1}}
\newrefformat{lem}{Lemma\,\ref{#1}}
\newrefformat{fig}{Fig.\,\ref{#1}}
\newrefformat{tab}{Table\,\ref{#1}}

\newtheorem{theorem}{Theorem}
\newtheorem{problem}{Problem}

\newtheorem{corollary}[theorem]{Corollary}

\newtheorem{proposition}[theorem]{Proposition}

\newcommand{\cf}{\emph{cf.}\xspace}

\newcommand{\bdmath}{\begin{dmath}}
\newcommand{\edmath}{\end{dmath}}
\newcommand{\beq}{\begin{equation}}
\newcommand{\eeq}{\end{equation}}
\newcommand{\bdm}{\begin{displaymath}}
\newcommand{\edm}{\end{displaymath}}
\newcommand{\bea}{\begin{eqnarray}}
\newcommand{\eea}{\end{eqnarray}}
\newcommand{\beal}{\beq \begin{array}{ll}}
\newcommand{\eeal}{\end{array} \eeq}
\newcommand{\beas}{\begin{eqnarray*}}
\newcommand{\eeas}{\end{eqnarray*}}
\newcommand{\ba}{\begin{array}}
\newcommand{\ea}{\end{array}}
\newcommand{\bit}{\begin{itemize}}
\newcommand{\eit}{\end{itemize}}
\newcommand{\ben}{\begin{enumerate}}
\newcommand{\een}{\end{enumerate}}

\newcommand{\calB}{{\cal B}}

\newcommand{\calL}{{\cal L}}

\newcommand{\calN}{{\cal N}}

\newcommand{\calS}{{\cal S}}

\newcommand{\setal}{~\emph{et~al.}\xspace}
\newcommand{\eg}{\emph{e.g.,}\xspace}
\newcommand{\ie}{\emph{i.e.,}\xspace}
\newcommand{\myParagraph}[1]{{\bf #1.}\xspace}

\newcommand{\M}[1]{{\bm #1}} %
\renewcommand{\boldsymbol}[1]{{\bm #1}}

\newcommand{\hide}[1]{}
\newcommand{\wrt}{w.r.t.\xspace}

\newcommand{\hiddenText}{{\color{gray} hidden text.}}
\newcommand{\hideWithText}[1]{\hiddenText}

\newcommand{\kron}{\otimes}

\newcommand{\subject}{\text{ subject to }}
\DeclareMathOperator*{\argmax}{arg\,max}
\DeclareMathOperator*{\argmin}{arg\,min}

\newcommand{\norm}[1]{\left\| #1 \right\|}

\newcommand{\prob}[1]{{\mathbb P}\left(#1\right)}
\newcommand{\tran}{^{\mathsf{T}}}

\newcommand{\trace}[1]{\mathrm{tr}\left(#1\right)}

\newcommand{\rank}[1]{\mathrm{rank}\left(#1\right)}

\newcommand{\inv}{^{-1}}

\newcommand{\ones}{{\mathbf 1}}
\newcommand{\zero}{{\mathbf 0}}
\newcommand{\eye}{{\mathbf I}}
\newcommand{\vect}[1]{\left[\begin{array}{c}  #1  \end{array}\right]}

\newcommand{\Real}[1]{ { {\mathbb R}^{#1} } }

\newcommand{\SOthree}{\ensuremath{\mathrm{SO}(3)}\xspace}

\newcommand{\MA}{\M{A}}
\newcommand{\MB}{\M{B}}

\newcommand{\MG}{\M{G}}
\newcommand{\MM}{\M{M}}

\newcommand{\MP}{\M{P}}
\newcommand{\MQ}{\M{Q}}

\newcommand{\MR}{\M{R}}

\newcommand{\MH}{\M{H}}

\newcommand{\MX}{\M{X}}
\newcommand{\MY}{\M{Y}}

\newcommand{\vh}{\boldsymbol{h}} 
\newcommand{\vb}{\boldsymbol{b}}
\newcommand{\vc}{\boldsymbol{c}}

\newcommand{\ve}{\boldsymbol{e}}

\newcommand{\vg}{\boldsymbol{g}}

\newcommand{\vr}{\boldsymbol{r}}
\newcommand{\vs}{\boldsymbol{s}}

\newcommand{\vv}{\boldsymbol{v}}
\newcommand{\vt}{\boldsymbol{t}}
\newcommand{\vxx}{\boldsymbol{x}} 
\newcommand{\vy}{\boldsymbol{y}}

\newcommand{\vepsilon}{\boldsymbol{\epsilon}}

\newcommand{\scenario}[1]{{\smaller \sf#1}\xspace}

\newcommand{\blue}[1]{{\color{blue}#1}}
\newcommand{\green}[1]{{\color{green}#1}}
\newcommand{\red}[1]{{\color{red}#1}}

\newcommand{\linkToPdf}[1]{\href{#1}{\blue{(pdf)}}}
\newcommand{\linkToPpt}[1]{\href{#1}{\blue{(ppt)}}}
\newcommand{\linkToCode}[1]{\href{#1}{\blue{(code)}}}
\newcommand{\linkToWeb}[1]{\href{#1}{\blue{(web)}}}
\newcommand{\linkToVideo}[1]{\href{#1}{\blue{(video)}}}
\newcommand{\linkToMedia}[1]{\href{#1}{\blue{(media)}}}
\newcommand{\award}[1]{\xspace} %

\renewcommand{\subject}{\text{s.t.}}

\newcommand{\barcsq}{\barc^2}

\newcommand{\tls}{\scenario{TLS}}
\newcommand{\gm}{\scenario{GM}}

\newcommand{\gnc}{\scenario{GNC}}

\newcommand{\vectorize}[1]{\ensuremath{\mathrm{vec}\left(#1 \right) }}

\newcommand{\sym}{\mathcal{S}}

\newcommand{\pd}{\sym_{++}}

\renewcommand{\norm}[1]{\left\| #1 \right\|}

\newcommand{\omitted}[1]{}

\newcommand{\bmat}{\left[ \begin{array}}
\newcommand{\emat}{\end{array}\right]}

\newcommand{\edit}[1]{#1\xspace}

\newrefformat{eq}{Eq.\,\ref{#1}}
\newcommand{\subMeas}[1]{\calS} %

\newcommand{\domainX}{\mathbb{X}}

\newcommand{\apollo}{\scenario{ApolloScape}}
\newcommand{\apolloCar}{\scenario{ApolloCar3D}}

\newcommand{\pascal}{\scenario{PASCAL3D+}}

\newcommand{\fsdp}{f_{\mathrm{SDP}}}
\newcommand{\fest}{f_{\mathrm{est}}}

\newcommand{\barvy}{\bar{\vy}}

\newcommand{\barMB}{\bar{\MB}}
\newcommand{\vrhomo}{\tilde{\vr}}

\newcommand{\nrShapes}{K}

\newcommand{\inthr}{\varepsilon}
\renewcommand{\barcsq}{\varepsilon^2}
\newcommand{\MHtl}{\bar{\MH}}

\newcommand{\name}{\scenario{PACE$^\star$}}
\newcommand{\nameLong}{shaPe and pose estimAtion for Category-level pErception}%
\newcommand{\nameRobust}{\scenario{PACE\#}}
\newcommand{\nameRobustApolloDepths}{\scenario{PACE\#-ApolloDepths}}
\newcommand{\nameRobustGTDepths}{\scenario{PACE\#-GTDepths}}
\newcommand{\nameRobustGTKeypoints}{\scenario{PACE\#-GTKeypoints}}

\newcommand{\altern}{\scenario{Altern}}

\newcommand{\irlsgm}{\scenario{IRLS-GM}}
\newcommand{\irlstls}{\scenario{IRLS-TLS}}
\newcommand{\cliquename}{\scenario{Clique-PACE$^\star$}}
\newcommand{\aka}{\emph{a.k.a.}}

\newcommand{\toCheck}[1]{{#1}}
\newcommand{\toCheckTwo}[1]{{#1}}
\newcommand{\outPaceSharp}{$70-90\%$}
\newcommand{\outGNC}{$50-60\%$}
\newcommand{\weight}{\omega}

\newcommand{\best}[1]{{\bf #1}}
\newcommand{\secondBest}[1]{{#1}}
\usepackage[font=small,labelfont=bf]{caption}

\usepackage[numbers,sort&compress]{natbib}
\usepackage{siunitx}

 \newcommand{\isExtended}[2]{{#1}\xspace} %

\isExtended{
}
{ \usepackage{xr}
\externaldocument{supplementaryMaterial}
}

\begin{document}
\graphicspath{{./figures/}}

\title{Optimal Pose and Shape Estimation for Category-level 3D Object Perception}

\author{Jingnan Shi, Heng Yang, Luca Carlone \\
Laboratory for Information \& Decision Systems (LIDS) \\
Massachusetts Institute of Technology \\
\{jnshi, hankyang, lcarlone\}@mit.edu
}

\makeatletter

\maketitle

\begin{tikzpicture}[overlay, remember picture]
\path (current page.north east) ++(-3.2,-0.4) node[below left] {
This paper has been accepted for publication at the 2021 Robotics: Science and Systems Conference.
};
\end{tikzpicture}
\begin{tikzpicture}[overlay, remember picture]
\path (current page.north east) ++(-6.5,-0.8) node[below left] {
Please cite the paper as: J. Shi, H. Yang, and L. Carlone,
};
\end{tikzpicture}
\begin{tikzpicture}[overlay, remember picture]
\path (current page.north east) ++(-1.5,-1.2) node[below left] {
``Optimal Pose and Shape Estimation for Category-level 3D Object Perception,'' \emph{Robotics: Science and Systems (RSS)}, 2021.
};
\end{tikzpicture}

\begin{abstract}
We consider a \emph{category-level perception} problem, where one is given 
3D sensor data picturing an object of a given category (\eg a car), and has to reconstruct 
 the pose and shape of the object despite intra-class variability (\ie different car models have different shapes). 
 We consider an \emph{active shape model}, where ---for an 
object category--- we are given a library of potential CAD models describing objects in that category,
and 
we adopt a standard formulation where pose and shape estimation are formulated as a non-convex optimization. 
Our first contribution is to provide the first \emph{certifiably optimal} solver for pose and shape estimation. 
In particular, we show that rotation estimation can be decoupled from the estimation of the object translation and shape, and 
we demonstrate that (i) the optimal object rotation can be computed via a tight (small-size) semidefinite relaxation, 
and (ii) the translation and shape parameters can be computed in closed-form given the rotation.
Our second contribution is to add an outlier rejection layer to our solver, hence making it robust to a large 
number of misdetections. 
Towards this goal, 
we wrap our optimal solver in a robust estimation scheme based on \emph{graduated non-convexity}.
To further enhance robustness to outliers, we also develop the first graph-theoretic formulation to prune outliers 
in category-level perception, which removes outliers via convex hull and maximum clique computations; 
the resulting approach is robust to \toCheck{\outPaceSharp} outliers.
Our third contribution is an extensive experimental evaluation. Besides providing an ablation study on 
a simulated dataset and on the \pascal dataset, we combine our solver 
with a deep-learned keypoint detector, and show that the resulting approach \toCheckTwo{improves over the state of the art in vehicle pose estimation in the \apollo~datasets}.
\end{abstract}

\IEEEpeerreviewmaketitle

\section{Introduction}
\label{sec:intro}

Robotics applications, from self-driving cars to domestic robotics, demand
robots to be able to identify and estimate the 
pose and shape of objects in the environment. In self-driving applications, for instance, the perception system needs to 
estimate the poses of other vehicles in the surroundings, identify traffic lights and traffic signs, and 
detect pedestrians~\cite{Gupta20itsc-atom,Kolotouros19cvpr-shapeRec}. Similarly, domestic applications require estimating the location and shape of objects to support more effective interaction and manipulation~\cite{Manuelli19-kpam,Gao21-kpam2,Pavlakos17icra-semanticKeypoints}. 
Object pose estimation is made harder by the large intra-class shape variability of common objects: 
for instance, the shape of a car largely varies depending on the model (\eg take a station wagon versus a Smart). 

Despite the fast-paced progress, reliable 3D object pose estimation remains a challenge, 
as witnessed by recent self-driving car accidents caused by 
misdetections~\cite{teslaAccidentTruck,McCausland-UberCrash}.  
Deep learning has been making great strides in enabling robots to detect objects;
popular tools such as YOLO~\cite{Redmon16cvpr} and Mask-RCNN~\cite{He17iccv-maskRCNN} have made object detection possible on commodity hardware and with reasonable performance for in-distribution test data. 
However, detections are typically at the level of \mbox{categories (\eg car vs. pedestrian) rather than} 
at the level of instances (\eg a specific car model) and ---with the current methods--- 
enabling instance-level detections would require an unreasonably large amount of labeled data and 
computation (\eg to scale to million of potential instances). 
In turn, category-level perception renders the use of standard tools for pose estimation (from point cloud registration~\cite{Yang20tro-teaser,Horn87josa,Bustos18pami-GORE} to 2D-3D pose estimation~\cite{Zheng2013ICCV-revisitPnP,Kneip2014ECCV-UPnP,Schweighofer2008bmvc-SOSforPnP}) ineffective, since they rely on the knowledge of the shape of the object.

\begin{figure}[t]
    \centering
    \includegraphics[width=0.99\columnwidth]{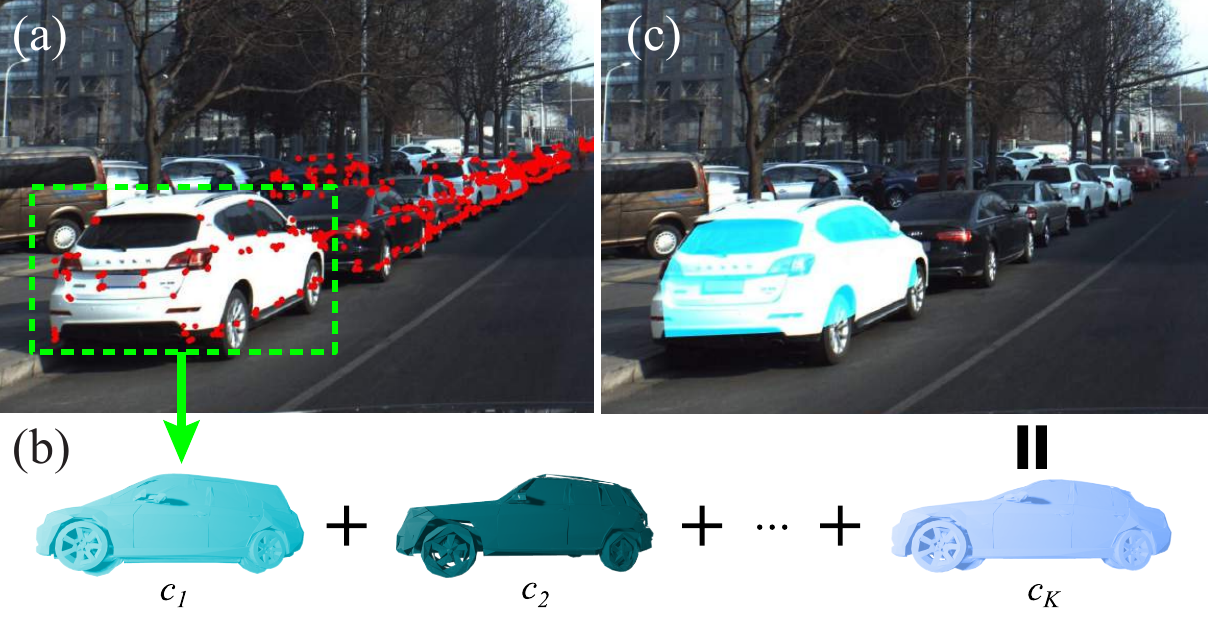}
    \caption{We propose 
    the first certifiably optimal approach to estimate the 3D pose and shape of 
    objects %
    from 3D keypoint detections (red points in (a)). 
    Our approach estimates pose and shape using an overcomplete library of CAD models (b) and
    is robust to 
    up to \toCheck{\outPaceSharp} outliers in the detections.
    (c) The approach \toCheckTwo{is more accurate than the state of the art} on the \apollo dataset~\cite{Wang19pami-apolloscape}. \label{fig:example} \vspace{-5mm}}
\end{figure}

These limitations have triggered robotics and computer vision 
research on category-level 3D object pose estimation. 
Traditional methods include the popular 
\emph{active shape model}~\cite{Cootes95cviu,Zhou15cvpr,Yang20cvpr-shapeStar}, where
one attempts to estimate the pose and shape of an object given a large database of 3D CAD models.
Despite its popularity (\eg the model is also used in human shape estimation and face detection~\cite{Zhou15cvpr}), pose estimation with active shape models leads to a non-convex optimization 
 problem and local solvers get stuck in poor solutions, and are sensitive to outliers~\cite{Zhou15cvpr,Yang20cvpr-shapeStar}.
More recently, research effort has been devoted to end-to-end learning-based 3D pose estimation 
with encouraging results in human pose estimation~\cite{Kolotouros19cvpr-shapeRec} and vehicle pose estimation~\cite{Chabot17-deepMANTA,Ke20-gsnet,Lopez19-vehicle,Kundu18-3dRCNN,Suwajanakorn18-latent3Dkeypoints}; 
these approaches still require a large amount of 3D labeled data, \mbox{which is hard to obtain in the wild.}

\myParagraph{Contribution} 
We address the shortcomings of existing approaches for pose and shape estimation based on the active shape model and propose the first approach that can compute optimal pose and shape estimates and is resilient to a large number of outliers.
We consider a {category-level perception} problem, where one is given 
3D keypoint detections of an object belonging to a given category (\eg detections of the wheels, rear-view mirrors, and other interest points of a car), and has to reconstruct 
 the pose and shape of the object despite intra-class variability.
 We assume the availability of 
a library of CAD models of objects in that category; 
such a library is typically available, since CAD models are extensively used in the design, manufacturing, and simulation of 3D objects.  

Our first contribution is \name, the first \emph{certifiably optimal} solver for 3D-3D pose and shape estimation. 
In particular, we show that ---despite the non-convexity of the problem--- rotation estimation can be decoupled from the estimation of object translation and shape, and 
we demonstrate that (i) the optimal object rotation can be computed via a tight (small-size) semidefinite relaxation, 
and (ii) the translation and shape parameters can be computed in closed form given the rotation.

Our second contribution is to equip \name with an outlier rejection scheme.
 Towards this goal, we extend existing tools for outlier rejection to category-level perception.
 In particular, we build on~\cite{Yang20tro-teaser} (which assumes the shape to be known) and  
 (i) show how to extend the graph-theoretic outlier pruning in~\cite{Yang20tro-teaser} to the case in which the shape is unknown, and (ii) apply a \emph{graduated non-convexity}~\cite{Yang20ral-GNC} scheme for robust estimation. 
 The resulting approach is named \nameRobust.

Our third contribution is an extensive experimental evaluation. %
We provide an ablation study on 
a simulated dataset and on the \pascal dataset, and show that \name is more accurate than 
iterative solvers, while \nameRobust~\toCheckTwo{dominates} other robust solvers and is robust to \toCheck{\outPaceSharp} outliers.
Finally, we integrate our solver in a realistic system ---including a deep-learned keypoint detector--- and show 
that the resulting approach \toCheckTwo{improves over the state of the art 
in vehicle pose estimation}
 in the \apollo~\cite{Wang19pami-apolloscape} driving datasets (Fig.~\ref{fig:example}).

\section{Related Work}
\label{sec:relatedWork}

{\bf Early approaches}  for category-level perception focus on 2D problems, where one has to locate objects 
---from human faces~\cite{Pantic00-facialExp} to resistors~\cite{Cootes95cviu}--- in images.
Classical approaches include \emph{active contour models}~\cite{Kass87ijcv,Chan99-activeContour} and 
\emph{active shape models}~\cite{Cootes95cviu,Cootes94-activeShapeMedical,Belongie02-shapeContext}.
 These works use techniques like PCA to build a library of 2D landmarks from training data, and then use iterative optimization
 algorithms to estimate the 2D object locations in the images, rather than estimating 3D poses. 

The landscape of category-level perception has been recently reshaped by the rapid adoption of 
{\bf convolutional networks}~\cite{Lecun98IEEE-CNN,Krizhevsky12nips-alexNet,Simonyan15-vgg}.
Pipelines using deep learning have seen great successes in areas such as human pose estimation~\cite{Toshev14-deepPose,Newell16-stackedHourglass,Tompson14-cnnHumanPose,He17iccv-maskRCNN, Martinez17-3DHumanPose},
and pose estimation of household objects~\cite{Manuelli19-kpam,Gao21-kpam2,Pavlakos17icra-semanticKeypoints}.
With the growing interest in self-driving vehicles, %
 research has also focused on jointly estimating vehicle shape and pose~\cite{Chabot17-deepMANTA,Ke20-gsnet,Lopez19-vehicle,Kundu18-3dRCNN,Suwajanakorn18-latent3Dkeypoints}.
Many open-source driving datasets have also been released for benchmarking~\cite{Wang19pami-apolloscape,Caesar20-nuscenes,Sun20-waymoDataset}.

For methods that aim to recover both the 3D shapes and poses of the objects of interests, a common paradigm is 
to use {\bf end-to-end methods}.
Usually, an encoder-decoder network is used to first convert input images to some latent representations,
and then use a decoder to map the latent representation back to 3D space~\cite{Groueix18arxiv-atlasnet,Richter18CVPR-MatryoshkaNetwork,Tatarchenko17iccv-OGN}.
Alternatively, recent work~\cite{Burchfiel19-probabilisticCategory} trains CNNs with generative representations of 3D objects to predict probabilistic distribution of object poses.
An additional alignment loss can also be incorporated into the network to regress for pose directly~\cite{Avetisyan19ICCV-e2eCADAlign, Manhardt19-2dlifting, Manhardt20-cps}.
One drawback of such approaches is that it is difficult for neural networks to learn the necessary 3D structure of the object on a per-pixel basis.
As shown in~\cite{Tatarchenko19CVPR-singleViewReconLimitation}, such networks can be outperformed by methods trained on model recognition and retrieval only.

{\bf Multi-stage methods} form another major paradigm for category-level perception.
Such approaches first recover the position of semantic keypoints~\cite{Pavlakos17icra-semanticKeypoints} in the images with neural networks,
and then recover the 3D pose of the object by solving a geometric optimization problem~\cite{Oberweger18-heatmapPose,Pavlakos17icra-semanticKeypoints,Peng19CVPR-PVNet,Mo19tr-orcVIO,Hou20-mobilepose}.
In some works, a canonical coordinate space is predicted by a network instead of relying on geometric reasoning~\cite{Wang19-normalizedCoordinate,Feng20-convCategory,Li20-categoryArticulated,Chen20-learnCanonicalShape}.
Lim\setal~\cite{Lim13iccv} establish 2D-3D correspondences between images and textureless CAD models by using HOG descriptors, and render edgemaps of the CAD models.
Chabot\setal~\cite{Chabot17-deepMANTA} use a two-staged approach to first regress a set of 2D part coordinates, and then choose the best corresponding 3D template and use PnP to solve for the 3D pose.
Pavlakos\setal~\cite{Pavlakos17icra-semanticKeypoints} use a stacked hourglass neural network~\cite{Newell16-stackedHourglass} for 2D semantic keypoint detection, and then 
employ block coordinate descent to resolve the object pose.
Zhou\setal~\cite{Zhou15cvpr,Zhou17pami-shapeEstimationConvex} propose a convex relaxation for jointly optimizing 3D shape parameters and object pose from 2D keypoints under a weak perspective camera model. Yang and Carlone~\cite{Yang20cvpr-shapeStar} apply the moment/sums-of-squares hierarchy~\cite{Blekherman12Book-sdpandConvexAlgebraicGeometry,Lasserre01siopt-LasserreHierarchy,Wang21SIOPT-tssos} to develop tighter relaxations than~\cite{Zhou15cvpr} but lead to semidefinite programs whose size grows with the number of CAD models. Probabilistic guarantees are studied in~\cite{Wangni21SIIMS-StatisticallyProvable2D3D}.

Our work belongs to the class of multi-stage methods, 
but we assume to have access to depth information for the semantic keypoints (\ie we consider a 3D-3D estimation setup~\cite{Shi20tr-robin}). Depth information %
is readily available in many robotics problems via direct sensing (\eg RGB-D or stereo)
 or algorithms (\eg mono depth techniques~\cite{Eigen14nips-monodepth,Lasinger19arXiv-robustMonocularDepthEstimation}). 
As we will show in Section~\ref{sec:optimalSolver-category-level}, the use of depth information allows us to mathematically \emph{decouple} the estimation of rotation from object translation and shape parameters, which leads to the first certifiably optimal solver that runs in a fraction of a second even in the presence of thousands of CAD models.

\section{Problem Statement: \\3D-3D Category-Level Perception}
\label{sec:problemStatement}

\myParagraph{Active Shape Model} We consider the problem of estimating the 3D pose $(\MR,\vt)$ and shape of an object, where $\MR \in \SOthree$ and 
$\vt \in \Real{3}$ are the unknown 3D rotation and translation of the object, respectively. 
We assume the object shape to be partially specified: we are given a library
of 3D CAD models $\calB_k$, $k=1,\ldots,\nrShapes$, and assume that the unknown object shape $\calS$ 
(modeled as a collection of 3D points) can be written 
as a combination of the given CAD models. 
More formally, each point $\vs(i)$ of the 
shape $\calS$ can be written as: 
\bea
\label{eq:activeShape}
\vs(i) = \textstyle\sum_{k=1}^{\nrShapes} c_{k} \vb_k(i)
\eea
where $\vb_k(i)$ is a given point belonging to the CAD model $\calB_k$;
the \emph{shape parameters} $\vc \triangleq [c_1 \ldots c_\nrShapes]\tran$ are unknown, and 
the entries of $\vc$ are assumed to be non-negative and sum up to 1 ($\vc \geq 0, \sum_{k=1}^{\nrShapes} c_{k} = 1$).
For instance, if ---upon estimation--- 
the vector $\vc$ has the $l$-th entry equal to 1 and the remaining entries equal to zero in~\eqref{eq:activeShape}, then the 
estimated shape of the object matches the $l$-th CAD model in the library; therefore, the estimation of the 
shape parameters $\vc$ can be understood as a fine-grained classification of the object among the instances in the library. However, the model is even more expressive, since it allows the object shape to be a convex 
combination of CAD models, which enables the active shape model~\eqref{eq:activeShape} to interpolate between different shapes in the library.

\myParagraph{Measurements} 
Towards the goal of estimating the object pose and shape, we are given a set of $N$ 3D keypoint detections.
 These are noisy measurements of 3D points belonging to the object and are commonly obtained using learning-based 
 semantic keypoint detectors applied to RGB-D or RGB+Lidar data (\eg~\cite{Pavlakos17icra-semanticKeypoints}). 
 Each measurement $\vy(i)$ ($i=1, \ldots, N$) is described by the following generative model:
\begin{equation}
\label{eq:generativeModel}
\vy(i)=\MR \textstyle\sum_{k=1}^{\nrShapes}c_{k} \vb_{k}(i) + \vt + \vepsilon (i) \qquad i=1, \ldots, N
\end{equation}
where the measurement $\vy(i)$ pictures a 3D point on the object (written as a linear combination $\sum_{k=1}^{\nrShapes}c_{k} \vb_{k}(i)$ of the shapes in the library as in~\eqref{eq:activeShape}), after these are rotated and translated 
according to the 3D pose $(\MR,\vt)$ of the object, and where $\vepsilon (i)$ represents measurement noise. 
Intuitively, each measurement corresponds to a noisy measurement of a semantic feature of the object 
(\eg wheel center or rear-view mirrors of a car) and each $\vb_{k}(i)$ corresponds to the feature location for a specific CAD model. We are now ready to state the 3D-3D category-level perception problem.

\begin{problem}[3D-3D Category-Level Perception]\label{prob:statement}
Compute the 3D pose $(\MR,\vt)$ and shape $(\vc)$ of an object given $N$ 3D keypoint measurements in the 
form~\eqref{eq:generativeModel}, possibly corrupted by outliers, \ie measurements with large error $\vepsilon (i)$.  
\end{problem}

\section{Certifiably Optimal Solver \\ for 3D-3D Category-Level Perception}
\label{sec:optimalSolver-category-level}

This section shows how to solve Problem~\ref{prob:statement} in the outlier-free case, where 
the noise $\vepsilon (i)$ is assumed to follow a zero-mean Gaussian distribution (we generalize the formulation
to deal with outliers in Section~\ref{sec:outlierRejection-category-level}). 
A standard formulation for the pose and shape estimation problem leads to the following 
\emph{regularized non-linear least squares} problem:
\bea
\label{eq:probOutFree-3D3Dcatlevel}
  \hspace{-5mm} \min_{\substack{\MR \in \SOthree, \\ \vt \in \Real{3}, \vc \in \Real{\nrShapes} } } & \displaystyle 
   \hspace{-2mm} \sum_{i=1}^{N} w_i \left\| \vy(i) - \MR \sum_{k=1}^{\nrShapes} c_{k} \vb_{k}(i) - \vt \right\|^{2} + \lambda \norm{\vc}^2 \\
  \subject & \ones\tran \vc  = 1  \nonumber%
\eea
where the first summand in the objective minimizes the residual error \wrt the generative model~\eqref{eq:generativeModel} ($w_i \geq 0, i=1,\dots,N$ are given weights), %
and the second term provides an $\ell_2$ regularization (\aka~\emph{Tikhonov regularization}~\cite{Tikhonov13book-numericalIllposed}) of the shape coefficients $\vc$ (controlled by the 
user-specified parameter $\lambda \geq 0$); the constraint $\ones\tran \vc  = 1$ (where $\ones$ is a vector with all entries equal to 1) forces the shape coefficients to
sum-up to 1; in this section, we drop the constraint that $\vc$ has to be nonnegative for mathematical convenience. 
Numerically, the regularization term ensures the problem is well-posed regardless of 
the number of shapes in the library (otherwise, the problem would be under-constrained when 
$K$ is large). 
From the probabilistic standpoint, problem~\eqref{eq:probOutFree-3D3Dcatlevel} is a \emph{maximum a posteriori}
estimator assuming that the keypoints measurement noise follows a zero-mean Gaussian with covariance $\frac{1}{w_i} \eye_3$ (where $\eye_3$ is the 3-by-3 identity matrix) and we have a zero-mean Gaussian prior with covariance $\frac{1}{\lambda}$ over the shape parameters $\vc$ (proof in Appendix~\ref{sec:app-mapOutlierFree}).

Problem~\eqref{eq:probOutFree-3D3Dcatlevel} is non-convex due to the product between rotation $\MR$ and shape 
parameters $\vc$ in the objective, and due to the nonconvexity of the constraint set $\SOthree$ the rotation $\MR$ 
is required to belong to, see \eg~\cite{Hartley13ijcv,Rosen18ijrr-sesync}. Therefore, existing approaches based on 
 local search~\cite{Lin14eccv-modelFitting,Gu06cvpr-faceAlignment,Ramakrishna12eccv-humanPose} are prone to converge to local minima corresponding to incorrect estimates.

\myParagraph{Results Overview} The rest of this section provides the first certifiably optimal algorithm to solve Problem~\eqref{eq:probOutFree-3D3Dcatlevel}. Towards this goal we show that 
(i) the translation $\vt$ in~\eqref{eq:probOutFree-3D3Dcatlevel} can be solved in closed form given the 
rotation and shape parameters (Section~\ref{sec:translation-cat-level}), 
(ii) the shape parameters $\vc$ can be solved in closed form given the rotation  (Section~\ref{sec:shape-cat-level}), 
and (iii) the rotation can be estimated (independently on shape and translation) using a 
tight semidefinite relaxation (Section~\ref{sec:rotation-cat-level}). 
This sequence of results leads to an optimal solver for pose and shape summarized in Section~\ref{sec:summary-cat-level}.

\subsection{Closed-form Translation Estimation}
\label{sec:translation-cat-level}

From simple inspection of~\eqref{eq:probOutFree-3D3Dcatlevel}, we observe that the vector $\vt$ is unconstrained and appears quadratically in the cost function, \ie from the standpoint of $\vt$, eq.~\eqref{eq:probOutFree-3D3Dcatlevel} is a \emph{linear} least squares problem. Therefore, for any choice of $\MR$ and $\vc$, the optimal translation 
can be computed in closed-form as:

\vspace{-3mm}
\bea
\label{eq:optTran}
\vt^{\star} (\MR,\vc) = \vy_w - \MR \textstyle\sum_{k=1}^{\nrShapes} c_{k} \vb_{k,w} 
\eea

\vspace{-2mm}
\noindent
where

\vspace{-2mm}
\small{
\begin{align}
\label{eq:weightedCentroids}
\!\!\vy_w \!\triangleq \!\frac{1}{(\sum_{i=1}^N \!w_i)} \!\sum_{i=1}^N \!w_i \vy(i), 
\;\;\;\; 
\vb_{k,w} \!\triangleq \!\frac{1}{(\sum_{i=1}^N \!w_i)} \!\sum_{i=1}^N  \!w_i \vb_k(i), \!\!\!
\end{align}
}

\normalsize
\noindent
are the weighted centroids of $\vy(i)$ and $\vb_k(i)$'s. 
This manipulation is common in related work, \eg~\cite{Zhou15cvpr,Yang20cvpr-shapeStar}.

\subsection{Closed-form Shape Estimation}
\label{sec:shape-cat-level}

Substituting the optimal translation~\eqref{eq:optTran} (as a function of $\MR$ and $\vc$) 
back into the cost function~\eqref{eq:probOutFree-3D3Dcatlevel}, we obtain an optimization problem that only 
depends on $\MR$ and $\vc$:
\bea\label{eq:translation-free-problem-1}
  \hspace{-3mm} \min_{\substack{\MR \in \SOthree, \vc \in \Real{\nrShapes}} } \!\!&\!\! \sum_{i=1}^{N} \left\Vert \bar{\vy}(i)  - \MR \sum_{k=1}^{\nrShapes} c_{k} \bar{\vb}_{k}(i) \right\Vert^{2} + \lambda \norm{\vc}^2 \\
  \subject & \ones\tran \vc - 1 = 0 \nonumber
\eea
where

\vspace{-7mm}
\begin{align}
  \bar{\vy}(i) \triangleq \sqrt{w_i} (\vy(i) - \vy_w), \ \ 
  \bar{\vb}_{k}(i) \triangleq \sqrt{w_i} (\vb_{k}(i) - \vb_{k,w}),
\end{align}
are the (weighted) relative positions of $\vy(i)$ and $\vb_k(i)$ \wrt~their corresponding weighted centroids. 
Using the fact that the $\ell_2$ norm is invariant to rotation,
problem~\eqref{eq:translation-free-problem-1} is equivalent to:
\bea\label{eq:translation-free-problem}
\hspace{-3mm} \min_{\substack{\MR \in \SOthree, \vc \in \Real{\nrShapes}} } \!\!&\!\! \sum_{i=1}^{N} \left\Vert \MR\tran \bar{\vy}(i)  - \sum_{k=1}^{\nrShapes} c_{k} \bar{\vb}_{k}(i) \right\Vert^{2} + \lambda \norm{\vc}^2 \\
  \subject & \ones\tran \vc - 1 = 0 \nonumber
\eea

We can further simplify the expression by adopting the following matrix notations:

\vspace{-5mm}
\begin{align}
\bar{\vy} &= \left( \bar{\vy}(1)^{\tran}, \ldots, \bar{\vy}(N)^{\tran} \right)^{\tran} 
\in \Real{3N} 
\\
\bar{\MB} &= \begin{bmatrix}
\bar{\vb}_{1}(1) & \cdots & \bar{\vb}_{\nrShapes}(1) \\
\vdots & \ddots & \vdots \\
\bar{\vb}_{1}(N) & \cdots & \bar{\vb}_{\nrShapes}(N)
\end{bmatrix} 
\in \Real{3N \times \nrShapes} 
\end{align}
which allows rewriting~\eqref{eq:translation-free-problem} in the following compact form:
\bea \label{eq:clsofc}
  \min_{\substack{\MR \in \SOthree, \vc \in \Real{\nrShapes}} } & \left\Vert \barMB \vc -  
(\eye_N \kron \MR\tran) \barvy
  \right\Vert^{2} + \lambda \norm{\vc}^2 \\
  \subject & \ones\tran \vc - 1 = 0 \nonumber
\eea
Now the reader can again recognize that ---for any choice of $\MR$--- 
problem~\eqref{eq:clsofc} is a linearly-constrained linear least squares problem in $\vc$, 
which admits a closed-form solution.

\begin{proposition}[Optimal Shape]\label{prop:shapeEstimation}
For any choice of rotation $\MR$, the optimal shape parameters that solve~\eqref{eq:clsofc} can be 
computed in closed-form as:
\bea \label{eq:optimalvcofR}
\vc^{\star} (\MR) = 2\MG \barMB\tran (\eye_N \kron \MR\tran) \barvy + \vg
\eea
where we defined the following constant matrices and vectors:
\bea
\MHtl \triangleq 2(\barMB\tran\barMB + \lambda \eye_K)  \label{eq:inverseofdensematrix}\\
\MG \triangleq \MHtl\inv - \frac{\MHtl\inv \ones \ones\tran \MHtl\inv }{\ones\tran \MHtl\inv \ones},\quad 
\vg \triangleq \frac{\MHtl\inv \ones}{\ones\tran \MHtl\inv \ones}
\eea
\end{proposition}

\subsection{Certifiably Optimal Rotation Estimation}
\label{sec:rotation-cat-level}

Substituting the optimal shape parameters~\eqref{eq:optimalvcofR} (as a function of $\MR$) 
back into the cost function~\eqref{eq:clsofc}, we obtain an optimization problem that only 
depends on $\MR$:
\bea \label{eq:nonconvexR}
\min_{\MR \in \SOthree}  \norm{\MM (\eye_N \kron \MR\tran)\barvy + \vh}^2
\eea
where the matrix $\MM \in \Real{(3N+\nrShapes)\times 3N}$ and 
vector $\vh \in \Real{3N+\nrShapes}$ are defined as:
\bea
\MM \triangleq \bmat{c}
2\barMB \MG \barMB\tran - \eye_{3N} \\
2 \sqrt{\lambda} \MG \barMB\tran
\emat, \qquad  
\vh \triangleq \bmat{c}
\barMB \vg \\ \sqrt{\lambda} \vg
\emat .
\eea
Problem~\eqref{eq:nonconvexR} is a quadratic optimization over the non-convex set $\SOthree$. 
It is known that the set $\MR \in \SOthree$ can be described as a set of quadratic equality constraints, see \eg~\cite{Tron15rssws3D-dualityPGO3D,Rosen18ijrr-sesync} or~\cite[Lemma 5]{Yang20cvpr-shapeStar}. 
Therefore, we can succinctly rewrite~\eqref{eq:nonconvexR} as a 
\emph{quadratically constrained quadratic program} (QCQP).

\begin{proposition}[Optimal Rotation]\label{prop:optRotation}
The category-level rotation estimation problem~\eqref{eq:nonconvexR} can be equivalently written as a
\emph{quadratically constrained quadratic program} (QCQP):
\bea
\label{eq:categoryQCQP}
\min_{\vrhomo \in \Real{10}} & \vrhomo\tran \MQ \vrhomo \\
\subject & \vrhomo\tran \MA_i \vrhomo = 0, \forall i = 1,\dots, 15 \nonumber
\eea
where $\vrhomo \triangleq [1,\vectorize{\MR}\tran]\tran \in \Real{10}$ is a vector stacking all the entries of the unknown 
rotation $\MR$ in~\eqref{eq:nonconvexR} (with an additional unit element), 
$\MQ \in \sym^{10}$ is a symmetric constant matrix (expression given in Appendix~\ref{sec:app-rotEst-details}), and 
$\MA_i \in \sym^{10}, i=1,\dots,15$ are the constant matrices that define the quadratic constraints 
describing the set $\SOthree$~\cite[Lemma 5]{Yang20cvpr-shapeStar}.
\end{proposition}

While a QCQP is still a non-convex problem, it admits a standard semidefinite relaxation, described below.

\begin{corollary}[Shor's Semidefinite Relaxation]\label{cor:optRotation-relax}
The following semidefinite program (SDP) is a convex relaxation of~\eqref{eq:categoryQCQP}.
\bea
\label{eq:categoryQCQPrelax}
\min_{\MX \in \sym^{10}} & \trace{\MQ \MX} \\
\subject & \trace{\MA_0 \MX} = 1,  \nonumber\\
& \trace{\MA_i \MX} = 0, \forall i=1,\dots,15  \nonumber\\
 & \MX \succeq 0  \nonumber
\eea
Moreover, when the optimal solution $\MX^\star$ of~\eqref{eq:categoryQCQPrelax} has rank 1, it can 
be factored as $\MX^\star = \vect{1 \\ \vectorize{\MR^\star}} [1 \; \vectorize{\MR^\star}]$ where $\MR^\star$ 
is the optimal rotation minimizing~\eqref{eq:nonconvexR}. 
\end{corollary}

The rationale behind using the relaxation~\eqref{eq:categoryQCQPrelax} is threefold: 
(i) similar to related quadratic problems over \SOthree~\cite{Rosen18ijrr-sesync,Yang20tro-teaser,Yang19iccv-QUASAR,Briales16iros,Eriksson18cvpr-strongDuality}, the relaxation~\eqref{eq:categoryQCQPrelax} empirically produces rank-1 ---and hence \emph{optimal}--- solutions in common problems; 
(ii) even when the relaxation is not tight, the problem allows computing how suboptimal the resulting estimate is;
(iii) the relaxation entails solving a small semidefinite program ($10 \times 10$ matrix size, and $16$ linear equality constraints), hence it can be solved in milliseconds 
using standard interior-point methods (\eg~MOSEK~\cite{mosek} interfaced via CVX~\cite{CVXwebsite} or CVXPY~\cite{Diamond16cvxpy}).  
The proposed solution falls in the class of \emph{certifiable algorithms}~(see \cite{Bandeira15arxiv}  
and Appendix A in~\cite{Yang20tro-teaser}), since it allows solving a hard (non-convex) problem efficiently and with 
provable a posteriori guarantees.
\subsection{Summary}
\label{sec:summary-cat-level}

The results in this section suggest a simple algorithm to compute a certifiably optimal solution to the 
original pose and shape estimation problem~\eqref{eq:probOutFree-3D3Dcatlevel}: 
(i) we fist compute the optimal rotation $\MR^\star$ using the semidefinite relaxation~\eqref{eq:categoryQCQPrelax} (which 
is independent from the translation and shape parameters); 
(ii) we retrieve the optimal shape $\vc^{\star} (\MR^\star)$ given the 
optimal rotation using~\eqref{eq:optimalvcofR}. 
Finally, we retrieve the optimal translation $\vt^{\star} (\MR^\star,\vc^\star)$ %
using~\eqref{eq:optTran}. We call the resulting algorithm \name (\emph{\nameLong}).

\section{Increasing Robustness via Outlier Pruning and Graduated Non-Convexity}
\label{sec:outlierRejection-category-level}

This section extends the optimal solver presented in the previous section to deal with the case where some of the  
measurements are outliers, \ie some measurements in~\eqref{eq:generativeModel} have unexpectedly large noise.
In such as case, problem~\eqref{eq:probOutFree-3D3Dcatlevel} (even when solved to optimality) does not 
return an accurate estimate since the quadratic cost in~\eqref{eq:probOutFree-3D3Dcatlevel} implicitly assumes the 
measurement noise to be a zero-mean Gaussian. 
This section first presents a pre-processing that filters out gross outliers from the measurements using 
a graph-theoretic pruning (Section~\ref{sec:robin-category-level}). Then, we show that the optimal solver (\name) can 
be easily re-used in a robust estimation framework based on graduated non-convexity~\cite{Yang20ral-GNC} (Section~\ref{sec:gnc-category-level}).

\subsection{Outlier Pruning for Category-Level Perception}
\label{sec:robin-category-level}

We use a graph-theoretic approach to prune outliers, 
similar to~\cite{Mangelson18icra,Yang20tro-teaser,Enqvist09iccv,Shi20tr-robin}. 
The key idea is to check if pairs of 3D keypoints can be mutually compatible (\ie can possibly be 
simultaneously inliers) and model pair-wise compatibility as edges in a graph where the nodes are the 
3D keypoints. Then, since the inliers are all mutually compatible, they must form a large clique in the graph 
and can be retrieved by computing the maximum clique.
Our main novelty is to develop an efficient mutual compatibility test for category-level perception, 
while related work has focused on known shapes~\cite{Yang20tro-teaser,Enqvist09iccv,Shi20tr-robin}. 

\begin{figure}[t]
    \centering
    \includegraphics[width=0.9\columnwidth]{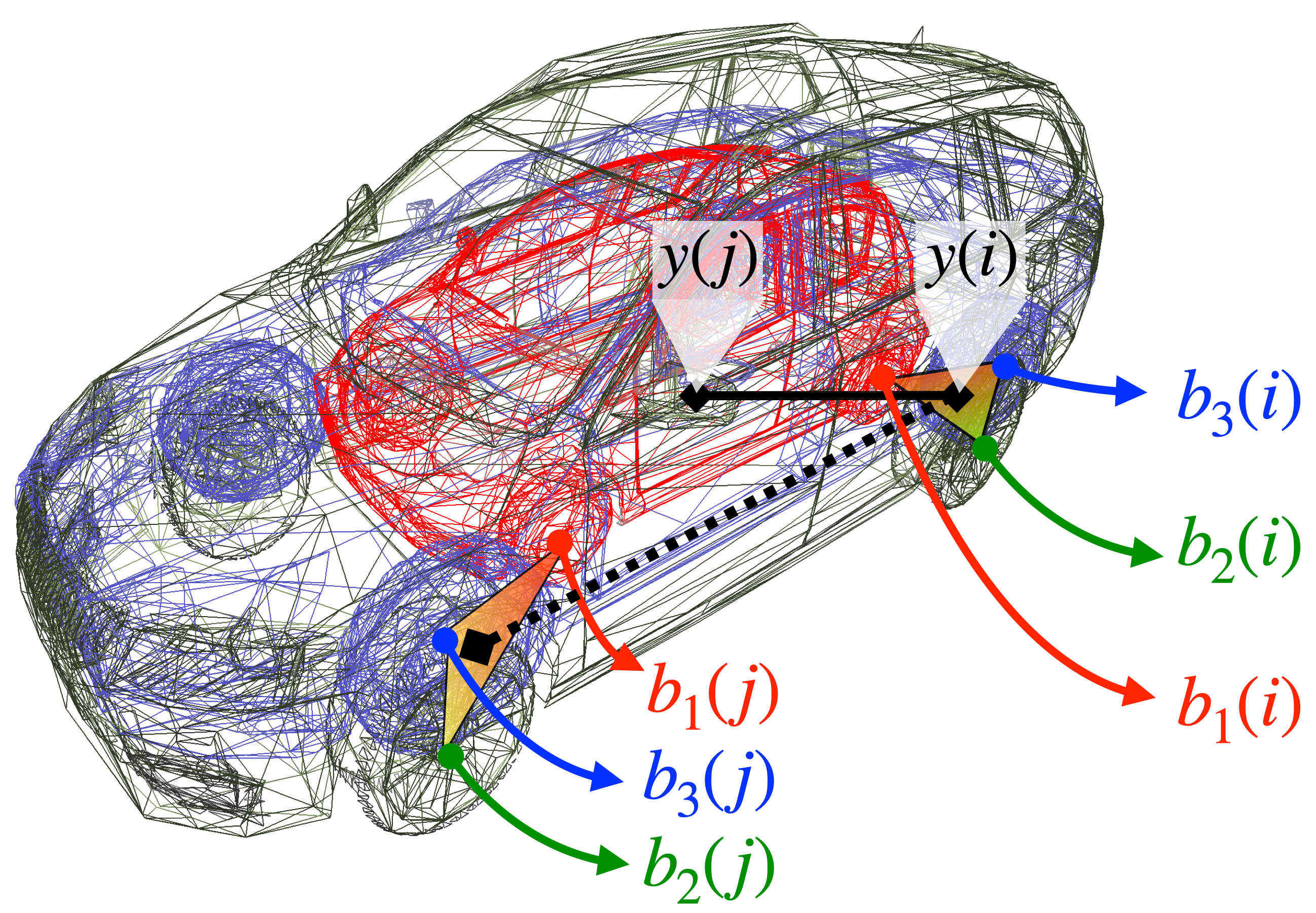}  
     \vspace{-1mm}
    \caption{Example of compatibility test %
    with 3 CAD models of cars (red, dark green, blue, indexed from 1 to 3). (Noiseless) inliers (\eg the detection of the back wheel $\vy(i)$ in the figure) 
    must fall in the convex hull of the corresponding points on the CAD models (\eg triangle $\vb_{1}(i)-\vb_{2}(i)-\vb_{3}(i)$ encompassing the back wheel positions across CAD models).
    This restricts the relative distance between two inliers and allows filtering out outliers.
     For instance, the dashed black line shows a distance that is compatible with the location of the convex hulls, while 
     the solid black line is too short compared to the relative position of the wheels (for any car model) and 
     allows pointing out that there is an outlier (\ie $\vy(j)$ in the figure).
     \label{fig:compatibility} \vspace{-5mm}}  
\end{figure}

\myParagraph{Mutually Compatible Measurements}
The goal here is to design a boolean condition that allows asserting if two measurements can be 
both inliers for any choice of pose and shape parameters. 
The challenge is that such a condition should not depend on the pose and shape parameters, which are unknown.
Therefore, we show how to manipulate the model~\eqref{eq:generativeModel} to obtain a condition that do not 
depend on $\MR$, $\vt$, and $\vc$. 
Towards this goal, let us call $\inthr$ the maximum error for a measurement to be called an inlier.
In other words, a measurement in~\eqref{eq:generativeModel} is an inlier if $\|\vepsilon(i)\|\leq \inthr$.

A pair of inliers $i$ and $j$ in eq.~\eqref{eq:generativeModel} must satisfy
$\|\vepsilon(i)\|\leq \inthr$ and $\|\vepsilon(j)\|\leq \inthr$. 
Taking the difference between measurement $i$ and $j$ in~\eqref{eq:generativeModel} leads to:
\bea
\vy(j) - \vy(i)= 
\MR \sum_{k=1}^{\nrShapes}c_{k} ( \vb_{k}(j) - \vb_{k}(i)) + (\vepsilon (j) - \vepsilon (i)) \nonumber
\eea 
where the translation cancels out in the subtraction. 
Now taking the $\ell_2$ norm of both members: %
\bea
\| \vy(j) - \vy(i) \|= 
\| \MR \sum_{k=1}^{\nrShapes}c_{k} ( \vb_{k}(j) - \vb_{k}(i)) + (\vepsilon (j) - \vepsilon (i)) \| \nonumber
\eea 
Using the triangle inequality and observing that $\|\vepsilon(i)\|\leq \inthr$ and $\|\vepsilon(j)\|\leq \inthr$ 
imply $\| \vepsilon (j) - \vepsilon (i) \| \leq 2 \inthr$:
\bea
- 2 \inthr \leq 
\| \vy(j) - \vy(i) \| - \| \MR \sum_{k=1}^{\nrShapes}c_{k} ( \vb_{k}(j) - \vb_{k}(i)) \| 
\leq  2 \inthr
\eea
Now observing that the $\ell_2$ norm is invariant to rotation and rearranging the terms:
\bea
\| \textstyle\sum_{k=1}^{\nrShapes}c_{k} ( \vb_{k}(j) - \vb_{k}(i)) \| - 2 \inthr \leq 
\;\; \| \vy(j) - \vy(i) \|   \;\;
\leq  \\ 
\| \textstyle\sum_{k=1}^{\nrShapes}c_{k} ( \vb_{k}(j) - \vb_{k}(i)) \| + 2 \inthr \nonumber
\eea
Considering the extreme cases over the set of possible shape coefficients:

\vspace{-5mm}
\bea \label{eq:definebminbmax}
\overbrace{\min_{ \vc \geq 0, \ones\tran \vc = 1} \|  \sum_{k=1}^{\nrShapes}c_{k} ( \vb_{k}(j) - \vb_{k}(i)) \|}^{b_{ij}^\min} - 2 \inthr \leq 
\| \vy(j) - \vy(i) \| 
\leq \\
 \underbrace{ \max_{\vc \geq 0, \ones\tran \vc = 1} \| \sum_{k=1}^{\nrShapes}c_{k} ( \vb_{k}(j) - \vb_{k}(i)) \|}_{ b_{ij}^\max } + 2 \inthr \nonumber
\eea
Since $\sum_{k=1}^{\nrShapes}c_{k} \vb_{k}(j)$ is a convex combinations of the points $\vb_{k}(j)$ ($k=1\ldots,\nrShapes$) and hence lies in the convex hull of such points, the term $\|  \sum_{k=1}^{\nrShapes}c_{k} ( \vb_{k}(j) - \vb_{k}(i)) \|$ represents the distance between two (unknown) points in the two
convex hulls defined by the set of points $\vb_{k}(j)$ and $\vb_{k}(i)$ ($k=1\ldots,\nrShapes$) (Fig.~\ref{fig:compatibility}).
 The minimum $b_{ij}^\min$ and the maximum 
$b_{ij}^\max$ over the convex hulls can be easily computed, either in closed form or via small convex programs (details in Appendix~\ref{sec:app-bmin-bmax}). 
Therefore,
a pair of inliers must satisfy:
 \bea
 \label{eq:compatible-pair-cat-level}
 b_{ij}^\min - 2 \inthr \leq 
\| \vy(j) - \vy(i) \| 
\leq 
b_{ij}^\max + 2 \inthr
\eea 
Note that $b_{ij}^\min$ and $b_{ij}^\max$ only depend on the given library and can be pre-computed.
Any pair of measurements that do not satisfy~\eqref{eq:compatible-pair-cat-level} cannot be simultaneously inliers 
for problem~\eqref{eq:generativeModel}.

\myParagraph{Largest Set of Compatible Measurements}
The compatibility test~\eqref{eq:compatible-pair-cat-level} checks if a pair of measurements can be together in the inlier set. Therefore, after testing compatibility between every pair of keypoints, we can find inliers by searching for the largest set of 
mutually compatible measurements. References~\cite{Mangelson18icra,Yang20tro-teaser,Enqvist09iccv,Shi20tr-robin} 
have already established that the largest set of mutually compatible measurements can be found by computing 
the maximum clique 
of a graph where nodes correspond to the 3D keypoints and an edge connects nodes $i$ and $j$ is the corresponding measurements satisfy the compatibility test~\eqref{eq:compatible-pair-cat-level}.
While we refer the reader to those papers for details, 
here we observe that such graph-theoretic approach has been shown to remove a large amount of gross outliers~\cite{Yang20tro-teaser} (while preserving all inliers). We will handle the remaining outliers using
 graduated non-convexity as discussed in the next section. In our experiments, we show that 
 while graduated non-convexity can be robust to up to \toCheck{\outGNC~outliers}, the addition of this graph-theoretic outlier 
 pruning boosts robustness to \toCheck{\outPaceSharp~outliers}.

\subsection{Graduated Non-Convexity for Category-Level Perception}
\label{sec:gnc-category-level}

While the graph-theoretic outlier pruning in the previous section 
is able to filter out a large fraction of gross outliers (without even computing an estimate), 
in this section we show how to use the remaining measurements (potentially still contaminated by a few outliers) 
to compute an accurate pose and shape estimate.
 Towards this goal, we use a standard robust estimation framework, 
 and we optimize the resulting optimization using graduated non-convexity (GNC)~\cite{Yang20ral-GNC}. 

As prescribed by standard robust estimation, 
we re-gain robustness to outliers by replacing
the squared $\ell_2$ norm in~\eqref{eq:probOutFree-3D3Dcatlevel} with a robust loss function $\rho$:
\bea
\label{eq:robust-3D3Dcatlevel}
  \hspace{-5mm} \min_{\substack{\MR \in \SOthree, \\ \vt \in \Real{3}, \vc \in \Real{\nrShapes} } } & 
  \displaystyle 
   \hspace{-3mm} \sum_{i=1}^{N} \rho \left( \left\| \vy(i) \!-\! \MR \sum_{k=1}^{\nrShapes} c_{k} \vb_{k}(i) \!-\! \vt \right\| \right) \!+\! \lambda \norm{\vc}^2 \hspace{-3mm} \nonumber\\
  \subject & \ones\tran \vc  = 1  %
\eea
While GNC can be applied to a broad class of loss functions~\cite{Yang20ral-GNC}, here 
we consider a truncated least square loss $\rho(r) = \min(r^2, \barcsq)$ which minimizes the squared 
residuals whenever they are below $\barcsq$ (note: the constant $\inthr$ is the same inlier threshold of the previous section) or becomes constant otherwise. Such cost function can be written by 
using auxiliary slack variables $\rho(r) = \min(r^2, \barcsq) = \min_{\weight\in\{0,1\}} \weight r^2 + (1-\weight)\barcsq$~\cite{Yang20ral-GNC}, 
hence allowing to rewrite~\eqref{eq:robust-3D3Dcatlevel} as:
\bea \label{eq:robust-3D3Dcatlevel2}
  \hspace{-3mm} \min_{\substack{\MR \in \SOthree, \\ \vt \in \Real{3}, \vc \in \Real{\nrShapes} \\ \weight_i\in\{0,1\} \forall i } } & 
  \displaystyle 
   \hspace{-3mm} \sum_{i=1}^{N} \weight_i \left\| \vy(i) \!-\! \MR \sum_{k=1}^{\nrShapes} c_{k} \vb_{k}(i) \!-\! \vt \right\|^2 
   \!\!
   + 
   \!(1\!-\!\weight_i)\barcsq  \!+\! \lambda \norm{\vc}^2 \hspace{-3mm} \nonumber \hspace{-3mm}\\
  \subject & \ones\tran \vc  = 1 \hspace{-3mm} %
\eea
In~\eqref{eq:robust-3D3Dcatlevel2}, when $\weight_i = 1$, the $i$-th measurement is considered an inlier and the 
cost minimizes the corresponding squared residual; when $\weight_i = 0$, the cost becomes independent of $\vy(i)$ hence the 
corresponding measurement is rejected as an outlier. Therefore, problem~\eqref{eq:robust-3D3Dcatlevel2} simultaneously 
estimates pose and shape variables $(\MR,\vt,\vc)$ while classifying inliers/outliers via the binary weights $\weight_i$ ($i=1,\ldots, N$).

Now the advantage is that we can minimize~\eqref{eq:robust-3D3Dcatlevel2} with an alternation 
scheme where we iteratively optimize  (i) over $(\MR,\vt,\vc)$ with fixed weights $\weight_i$ 
and (ii) over the weights $\weight_i$ with fixed $(\MR,\vt,\vc)$. 
This approach is convenient since the optimization over   $(\MR,\vt,\vc)$ can be solved to optimality using \name, 
while the optimization of the weights can be solved in closed form~\cite{Yang20ral-GNC}.
To improve convergence of this alternation scheme, we adopt graduated non-convexity~\cite{Blake1987book-visualReconstruction,Yang20ral-GNC,Yang20ral-GNC}, which starts with a convex approximation of the loss function in~\eqref{eq:robust-3D3Dcatlevel2} and then gradually increases the non-convexity until the original 
robust loss $\rho$ in~\eqref{eq:robust-3D3Dcatlevel2} is retrieved.

\emph{We call \nameRobust the approach applying graph-theoretic outlier pruning and then using 
GNC to retrieve a robust estimate.}

\section{Experiments}
\label{sec:experiments}
In this section, we first demonstrate the optimality of \name and the robustness of \nameRobust in simulated data and in the \pascal dataset~\cite{Xiang2014WACV-PASCAL+} (Section~\ref{sec:exp-optimality-robustness}). Then we show that \nameRobust can be integrated in a realistic perception system and achieve state-of-the-art performance on vehicle pose estimation in the \apollo dataset~\cite{Wang19pami-apolloscape} (Section~\ref{sec:exp-apollo}). 
\toCheckTwo{In both cases, our solvers outperform baseline approaches in terms of accuracy.}

\subsection{Ablation: Optimality and Robustness}
\label{sec:exp-optimality-robustness}

\newcommand{\mpwfour}{4.6cm}
\newcommand{\myhspace}{\hspace{-3.5mm}}
\begin{figure*}[t]
	\begin{center}
	\begin{minipage}{\textwidth}
	\begin{tabular}{cccc}%
		\myhspace \hspace{-3mm}
			\begin{minipage}{\mpwfour}%
			\centering%
			\includegraphics[width=\columnwidth]{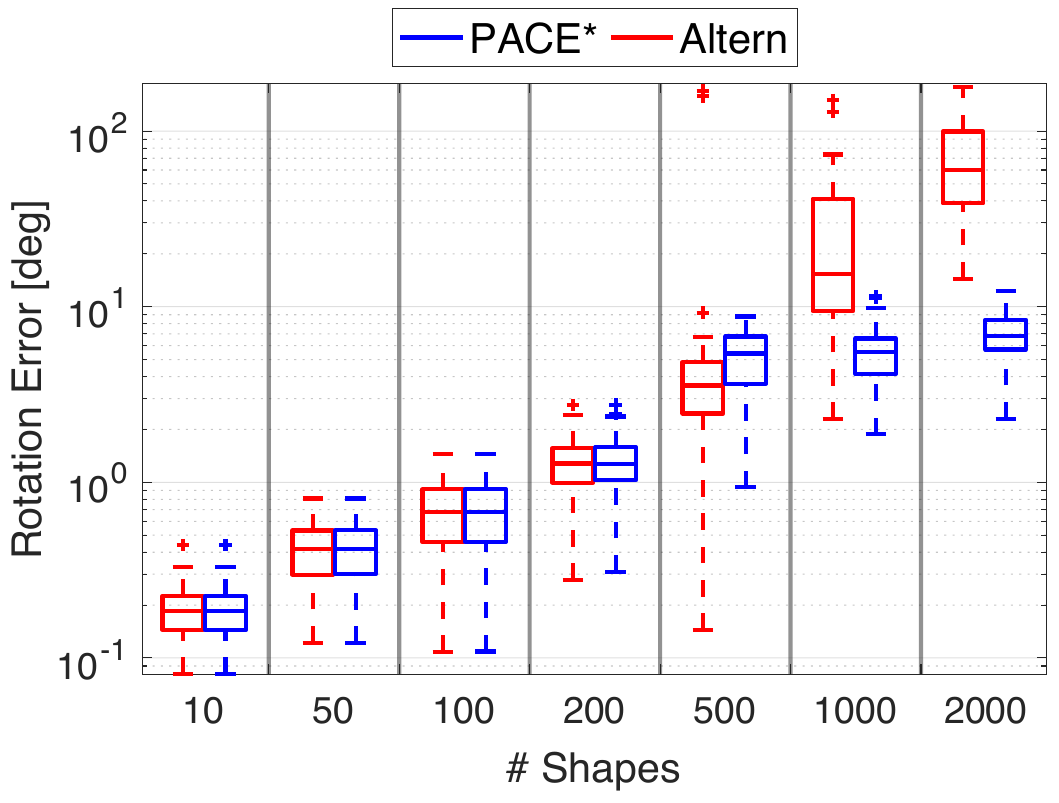}
			\end{minipage}
		&   \myhspace
			\begin{minipage}{\mpwfour}%
			\centering%
			\includegraphics[width=\columnwidth]{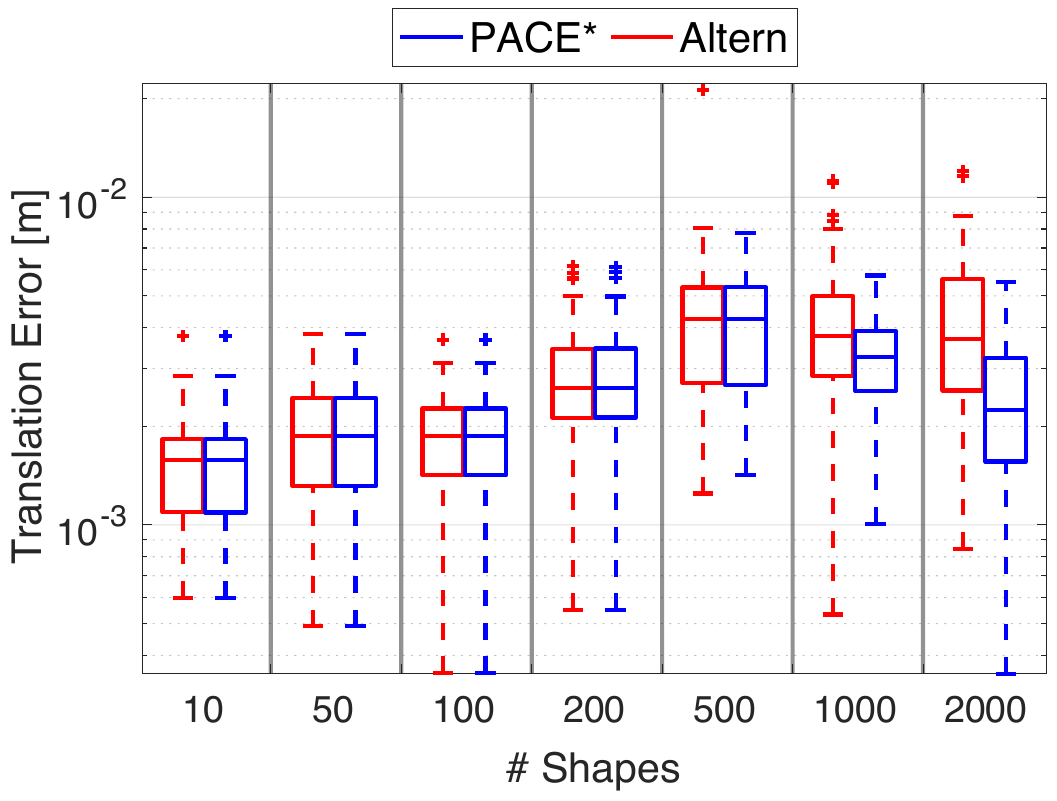}
			\end{minipage}
		&   \myhspace
			\begin{minipage}{\mpwfour}%
			\centering%
			\includegraphics[width=\columnwidth]{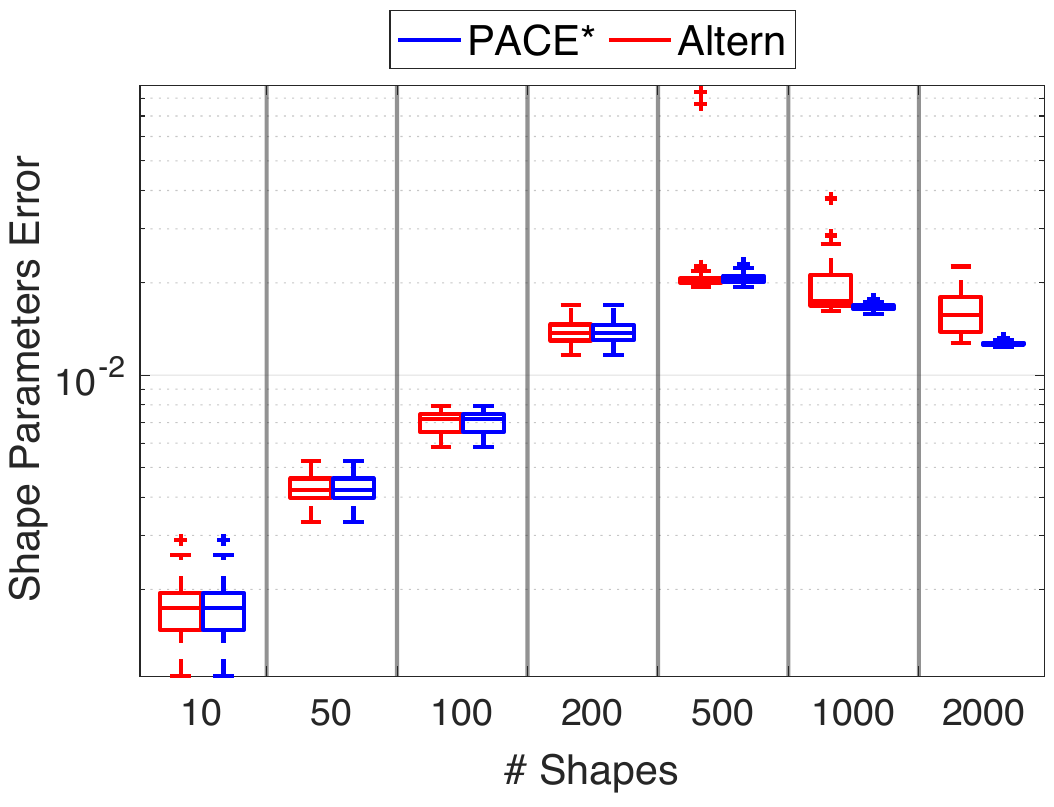}
			\end{minipage}
		&   \myhspace
			\begin{minipage}{\mpwfour}%
			\centering%
			\includegraphics[width=\columnwidth]{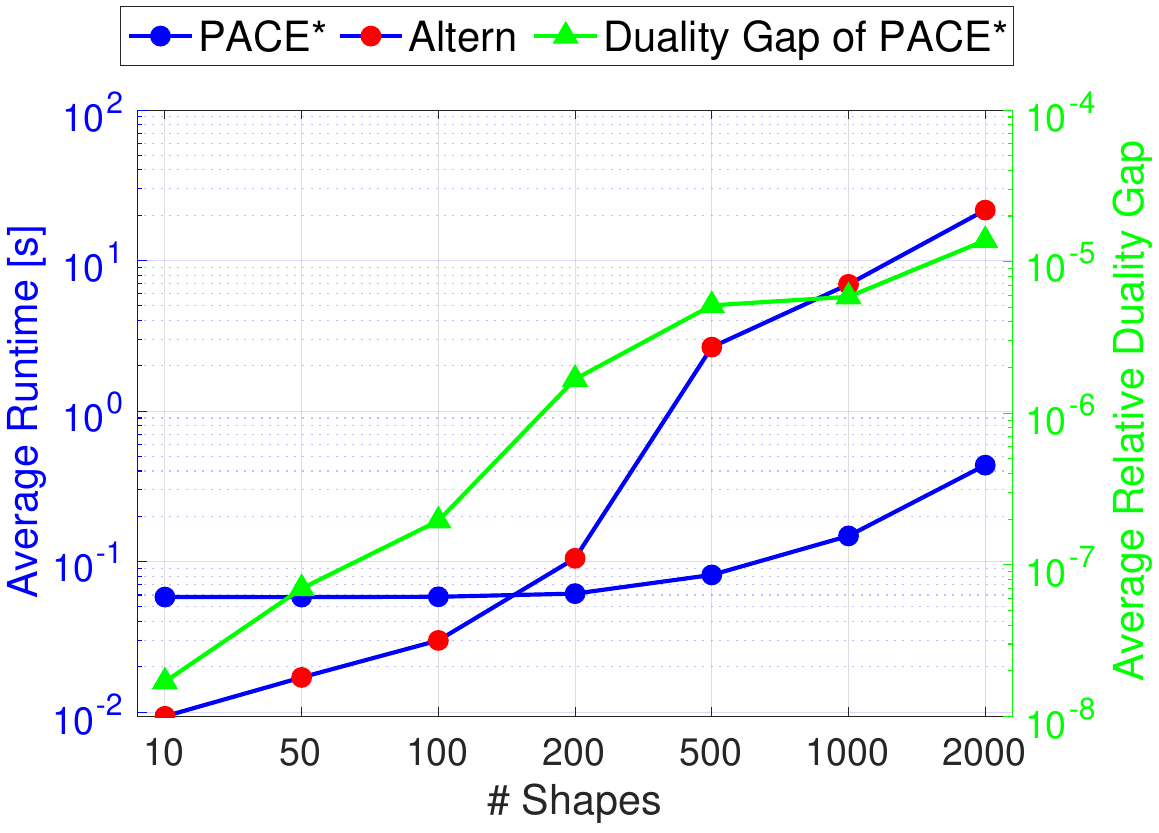}
			\end{minipage}
		\\
		\multicolumn{4}{c}{\smaller (a) Performance of the certifiably optimal solver \name~on outlier-free random simulated data: $N=100$. \vspace{1mm}}
		\\
		\myhspace \hspace{-3mm}
			\begin{minipage}{\mpwfour}%
			\centering%
			\includegraphics[width=\columnwidth]{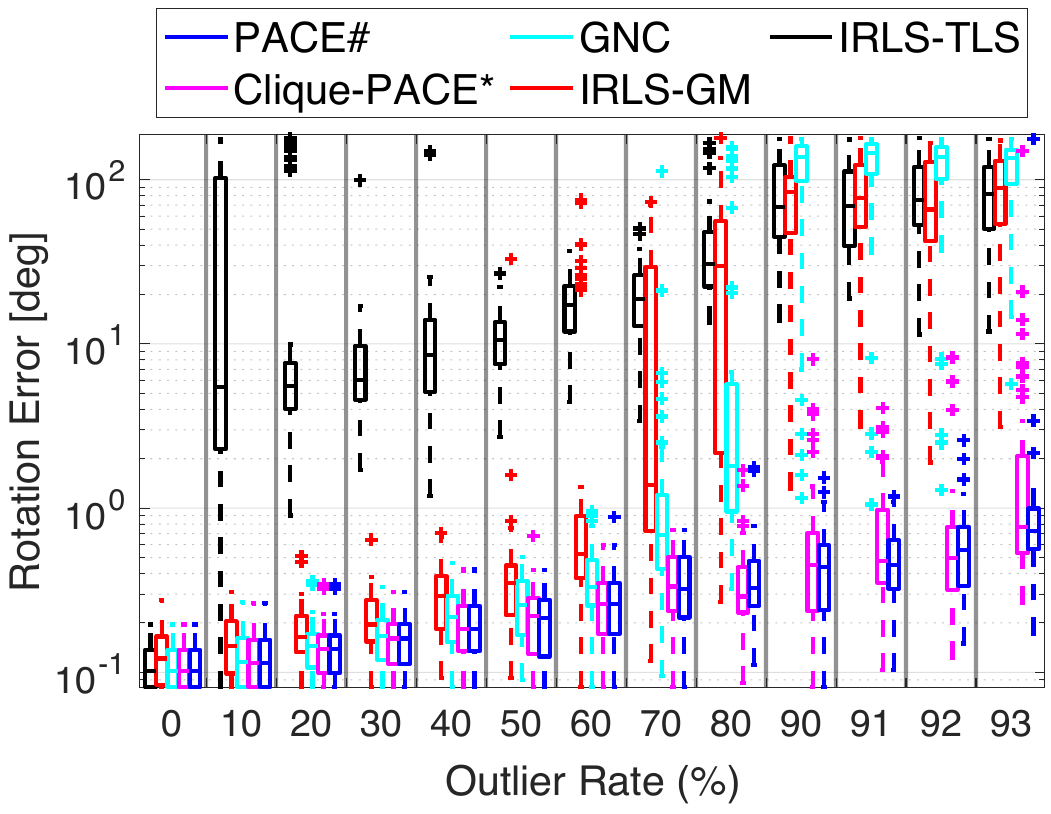}
			\end{minipage}
		&   \myhspace
			\begin{minipage}{\mpwfour}%
			\centering%
			\includegraphics[width=\columnwidth]{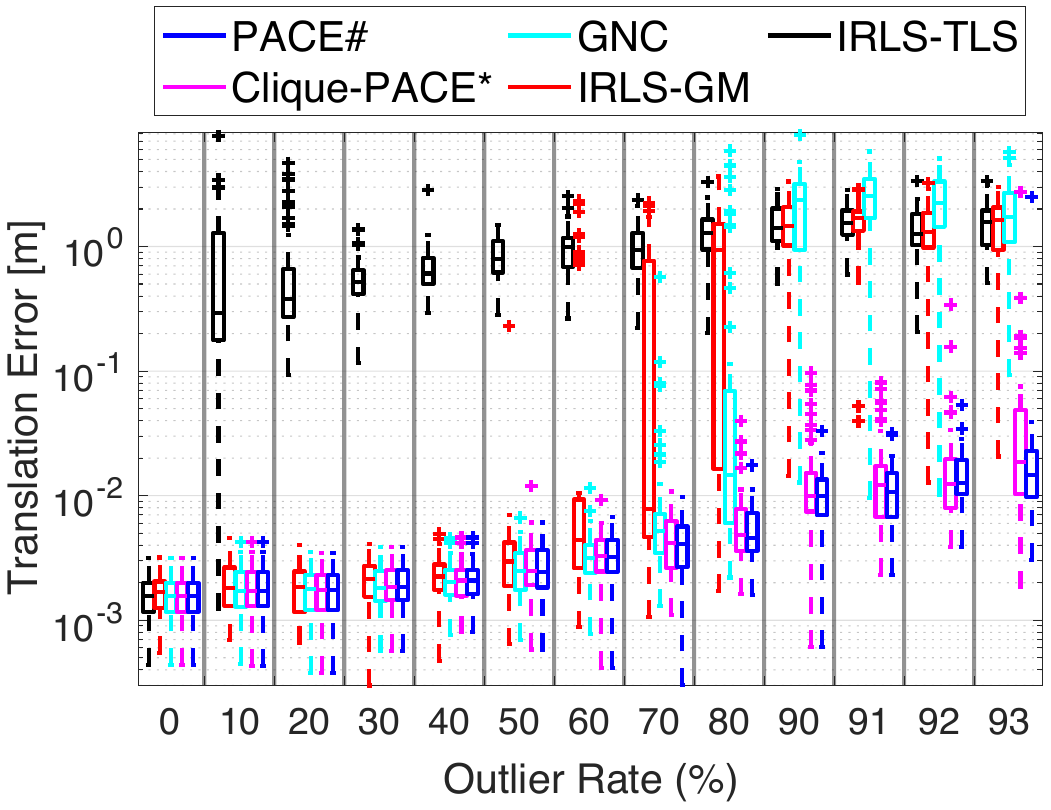}
			\end{minipage}
		&   \myhspace
			\begin{minipage}{\mpwfour}%
			\centering%
			\includegraphics[width=\columnwidth]{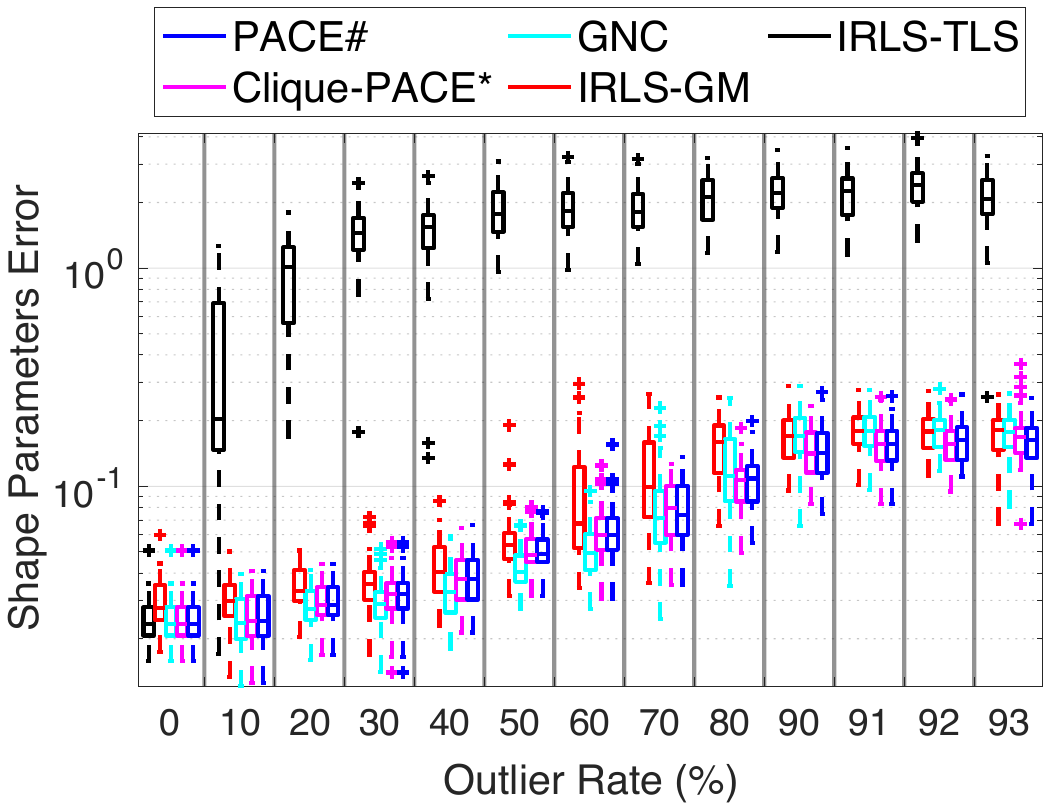}
			\end{minipage}
		&   \myhspace
			\begin{minipage}{\mpwfour}%
			\centering%
			\includegraphics[width=\columnwidth]{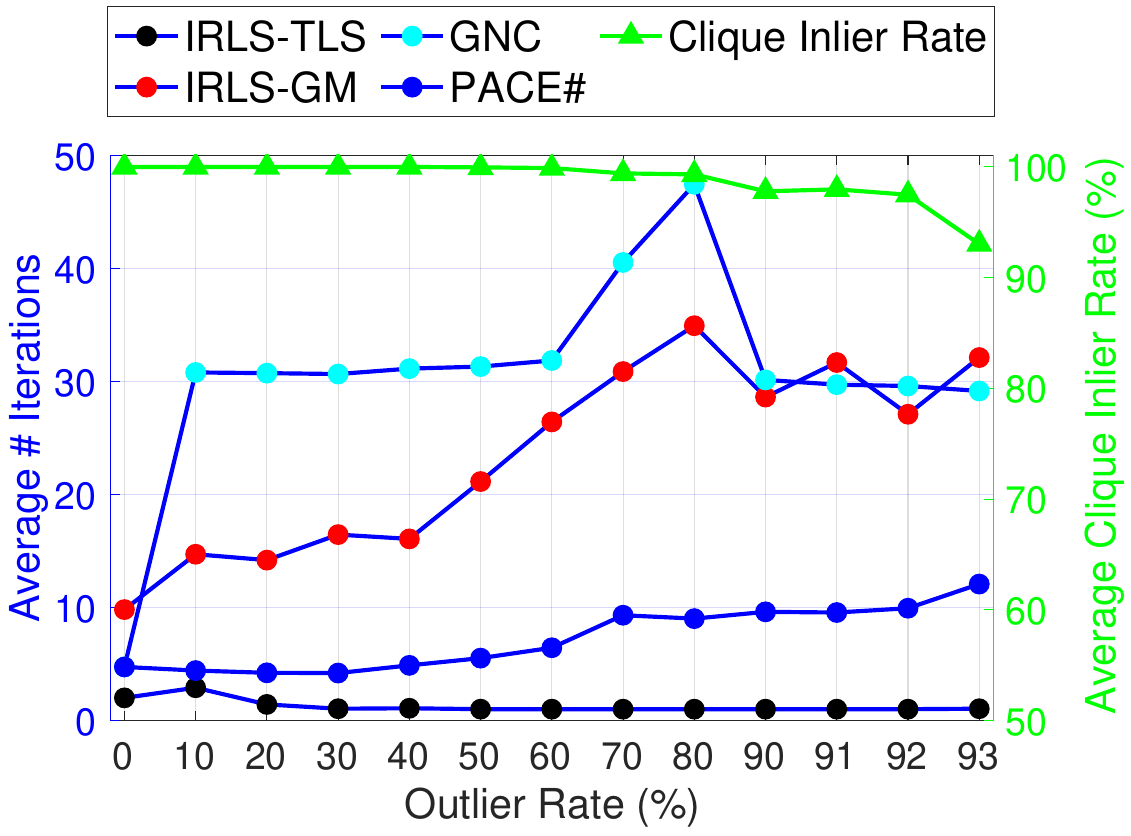}
			\end{minipage}
		\\
		\multicolumn{4}{c}{\smaller (b) Robustness of \nameRobust against increasing outliers on random simulated data: $N=100$, $K=10$, $r=0.1$. \vspace{1mm}}
		\\
		\myhspace \hspace{-3mm}
			\begin{minipage}{\mpwfour}%
			\centering%
			\includegraphics[width=\columnwidth]{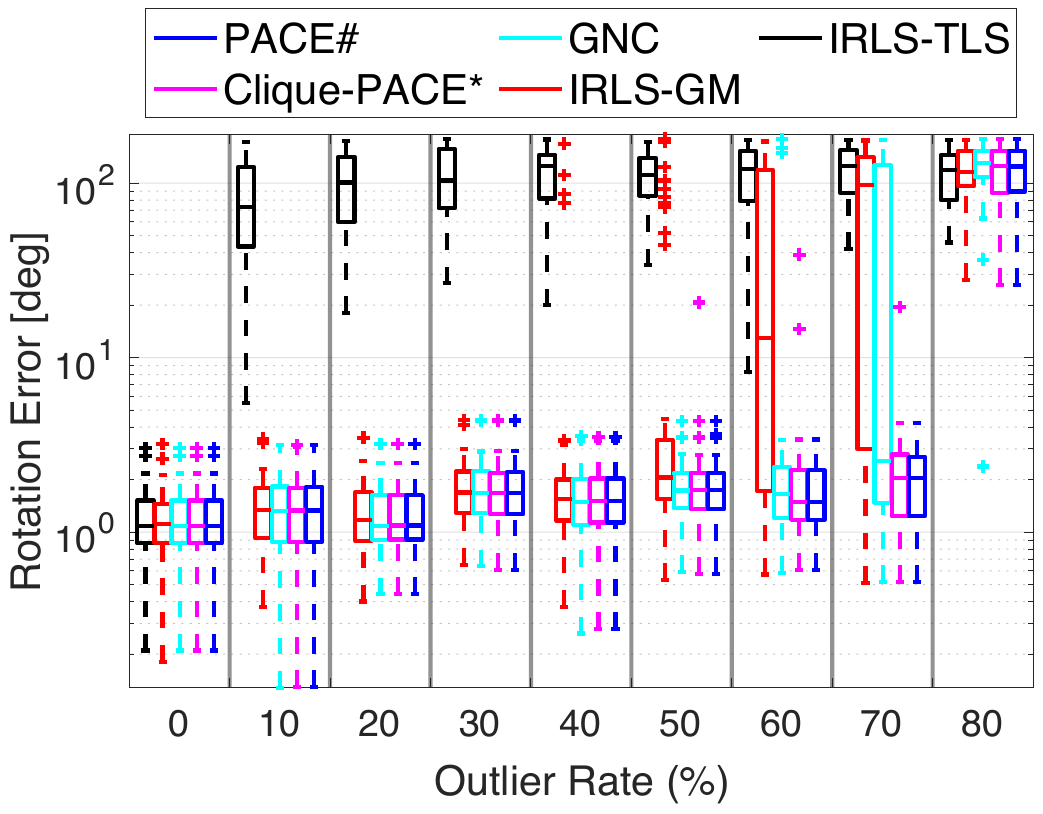}
			\end{minipage}
		&   \myhspace
			\begin{minipage}{\mpwfour}%
			\centering%
			\includegraphics[width=\columnwidth]{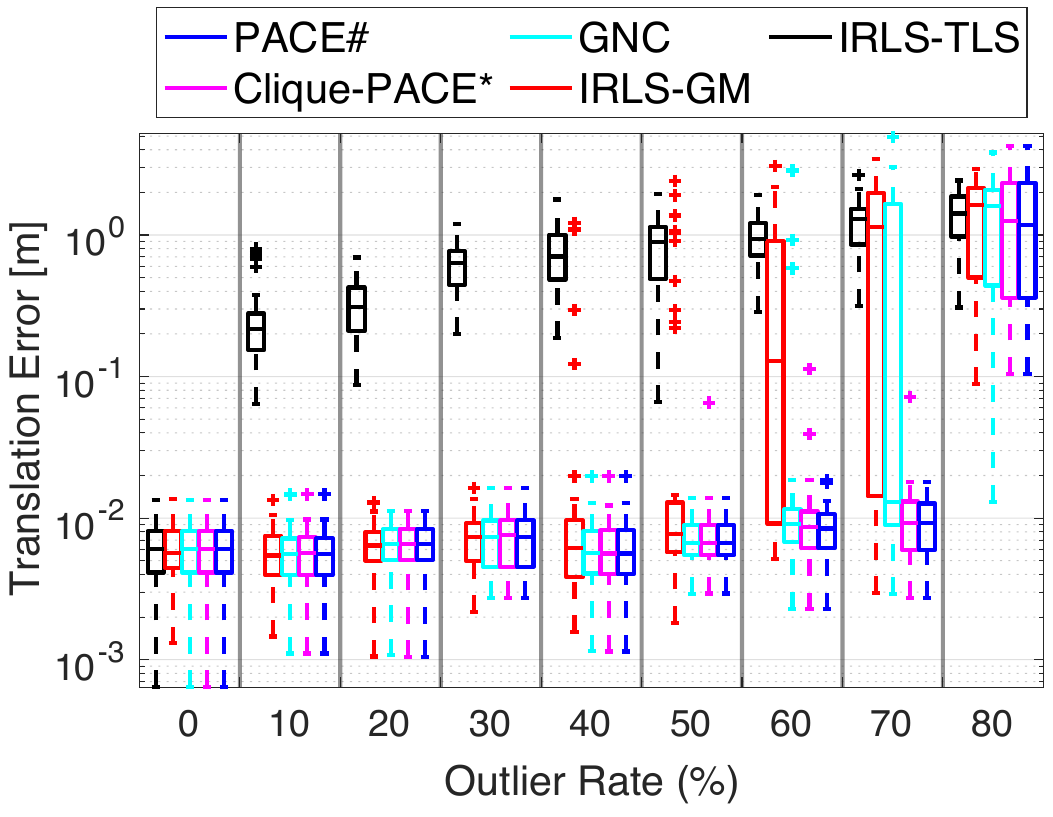}
			\end{minipage}
		&   \myhspace
			\begin{minipage}{\mpwfour}%
			\centering%
			\includegraphics[width=\columnwidth]{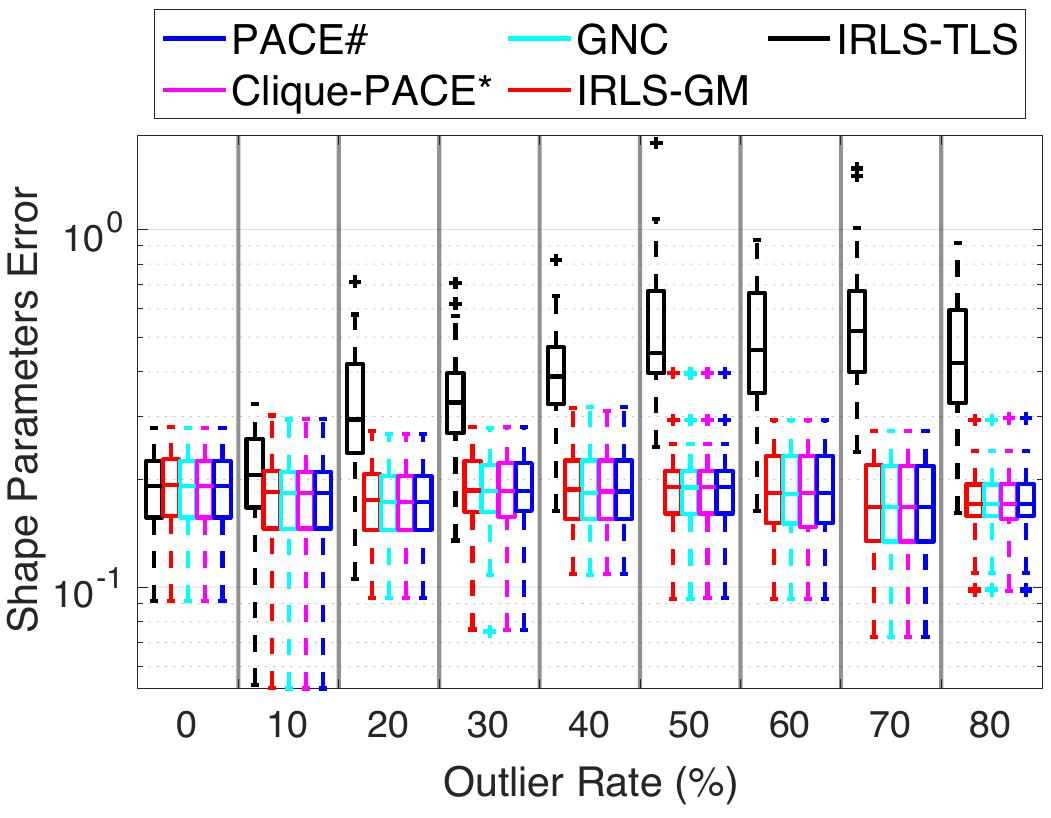}
			\end{minipage}
		&   \myhspace
			\begin{minipage}{\mpwfour}%
			\centering%
			\includegraphics[width=\columnwidth]{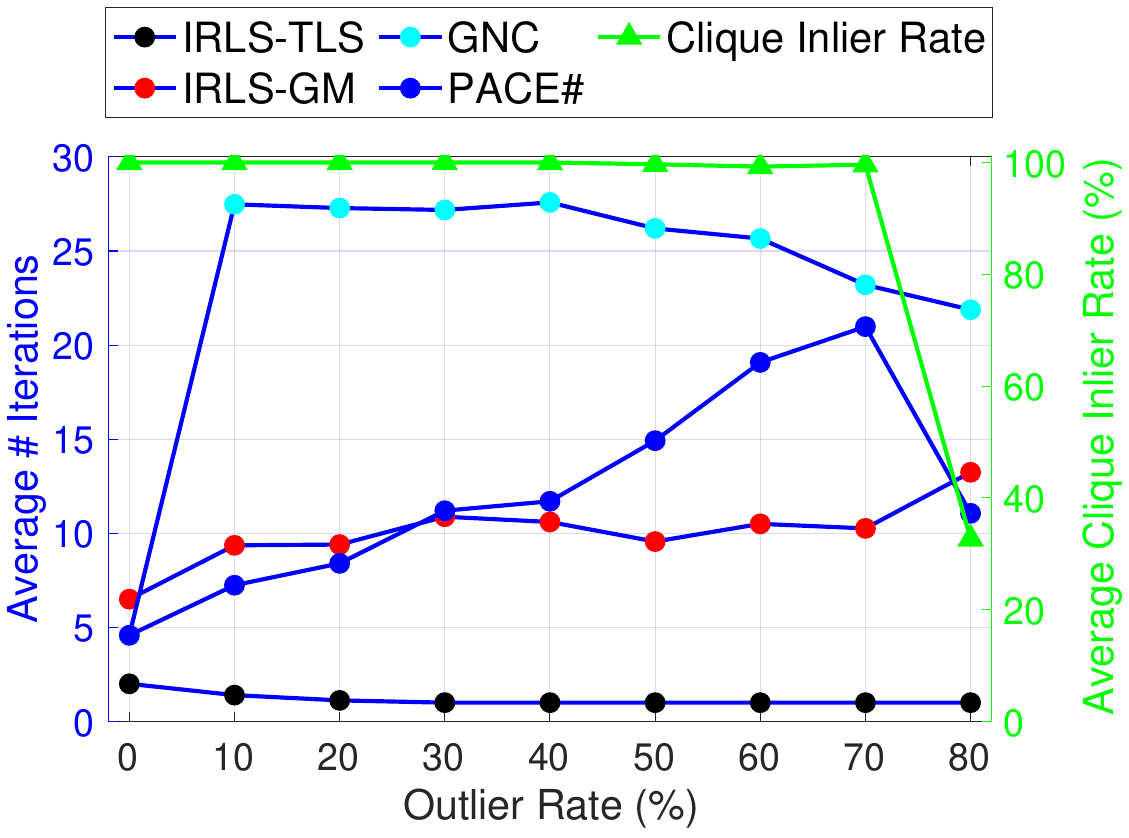}
			\end{minipage}
		\\
		\multicolumn{4}{c}{\smaller (c) Robustness of \nameRobust against increasing outliers on the \emph{car} category in the \pascal~dataset~\cite{Xiang2014WACV-PASCAL+}: $N=12$, $K=9$.\vspace{-2mm}}
	\end{tabular}
	\end{minipage} 
	\caption{Performance of \name~and \nameRobust~compared with baselines in simulated experiments. (a) \name~compared with alternating minimization (\altern) on random simulated outlier-free data with $N=100$ and $K$ increasing from $10$ to $2000$; (b) \nameRobust~along with two individual components of itself (\cliquename~and \gnc), compared with two variants of iterative reweighted least squares (\irlsgm~and \irlstls)~\cite{MacTavish15crv-robustEstimation} on random simulated outlier-contaminated data with $N=100$, $K=10$ and outlier rates up to $93\%$; (c) same as (b) but using the \emph{car} category CAD models from the \pascal~dataset~\cite{Xiang2014WACV-PASCAL+}, with $N=12$, $K=9$ and outlier rates up to $80\%$. Each boxplot and lineplot summarizes $50$ Monte Carlo random runs.
	\label{fig:simulation-robust}} 
	\vspace{-8mm} 
	\end{center}
\end{figure*}

\myParagraph{Optimality of \name} To evaluate the performance of \name in solving the outlier-free problem~\eqref{eq:probOutFree-3D3Dcatlevel}, we randomly simulate $K$ shape models $\calB_k$ whose points $\vb_k(i)$'s 
are drawn from an i.i.d. Gaussian distribution $\calN(\zero,\eye_3)$. We sample 
shape parameters $\vc$ uniformly at random in $[0,1]^\nrShapes$,
and normalize $\vc$ such that $\ones\tran\vc = 1$. Then we draw random poses $(\MR,\vt)$ \toCheck{as in~\cite{Yang20tro-teaser}} %
and generate the measurements $\vy(i)$ according to the model~\eqref{eq:generativeModel}, where the noise $\vepsilon(i)$ follows $\calN(\zero,\sigma^2\eye_3)$ with standard deviation $\sigma = 0.01$. We fix $N=100$, increase $K$ from $10$ up to $\num{2000}$ and set the regularization factor $\lambda = \sqrt{K/N}$ so that larger regularization is imposed when $K$ increases and the problem becomes more ill-posed. We compare \name with a baseline approach based on alternating minimization~\cite{Lin14eccv-modelFitting,Gu06cvpr-faceAlignment,Ramakrishna12eccv-humanPose} (details given in Appendix~\ref{sec:app-alternationApproach}) that offers no optimality guarantees (label: \altern). 

Fig.~\ref{fig:simulation-robust}(a) plots the statistics of rotation error (angular distance between estimated and ground-truth rotations), translation error, shape parameters error ($\ell_2$ distance between estimated and ground-truth translation/shape parameters), as well as average runtime and relative duality gap (details in Appendix~\ref{sec:app-sdp-relaxation-gap}).
We make the following observations: (i) \name returns accurate pose and shape estimates up to $K=2000$, while \altern starts failing at $K=500$. (ii) Although \altern is faster than \name for small $K$ (\eg~$K<200$), \name is orders of magnitude faster than \altern for large $K$. In fact, the runtime of \name only slightly increases because \name solves a fixed-size SDP regardless of the increase in $K$ (the increase in runtime is due to inversion of a dense matrix in~\eqref{eq:inverseofdensematrix}). (iii) The relaxation~\eqref{eq:categoryQCQPrelax} is empirically tight (duality gap $< 10^{-4}$), certifying global optimality of the solution returned by \name.

\myParagraph{Robustness of \nameRobust}
To test the robustness of \nameRobust on outlier-contaminated data, we follow the same data generation protocol as before, except that (i) when generating the CAD models, we follow a more realistic active shape model~\cite{Cootes95cviu} where we first generate a mean shape $\calB$ whose points $\vb(i)$'s are i.i.d. Gaussian $\calN(\zero,\eye_3)$, and then each CAD model is generated from the mean shape by: $\vb_k(i) = \vb(i) + \vv(i)$, where $\vv(i)$ follows $\calN(\zero,r^2 \eye_3)$ and represents the \emph{intra-class variation} of semantic keypoints with variation  radius $r$. (ii) we replace a fraction of the measurements $\vy(i)$ with arbitrary 3D points sampled according to $\calN(\zero,\eye_3)$ and violating the generative model~\eqref{eq:generativeModel}. We compare \nameRobust with two variants: \cliquename (\ie~after pruning outliers using maximum clique, \name is applied \emph{without} \gnc) and \gnc (\ie~\gnc is applied to problem data \emph{without} any outlier pruning), as well as two variants of the popular \emph{iterative reweighted least squares} heuristics: \irlstls and \irlsgm, where \tls and \gm denote the truncated least squares cost function and the Geman-McClure cost function~\cite{MacTavish15crv-robustEstimation}. For fair comparison, we use \name inside \nameRobust, \gnc, \irlstls, and \irlsgm when updating $(\MR,\vt,\vc)$ given fixed weights. We set $\inthr = 0.05$ for outlier pruning and \gnc. Fig.~\ref{fig:simulation-robust}(b) plots the results under increasing outlier rates up to $93\%$ when $N=100$, $K=10$ and $r=0.1$. We make the following observations: (i) \irlstls quickly fails (at $10\%$ outlier rate) due to the binary nature of the \tls cost, while \irlsgm is robust to $40\%$ outliers. (ii) \gnc alone already \toCheckTwo{outperforms} \irlstls and \irlsgm and is robust to \toCheck{$60\%$} outliers. (iii) With our maximum-clique outlier pruning, the robustness of \nameRobust is boosted to $92\%$, a level that is comparable to cases when the shapes are known (\eg~\cite{Yang20tro-teaser}). In addition, outlier pruning speeds up the convergence of \gnc (\cf~number of iterations plot for \gnc and \nameRobust in Fig.~\ref{fig:simulation-robust}(b)). (iv) Even without \gnc, the outlier pruning is so effective that \name alone is able to succeed with up to $90\%$ outliers, despite that the estimates are typically less accurate than \nameRobust. In fact, looking at the clique inlier rate plot (\green{green} lineplot in Fig.~\ref{fig:simulation-robust}(b)), the reader sees that the set of measurements after maximum clique pruning is almost free of outliers, explaining the surprising performance of \cliquename. In Appendix~\ref{sec:app-experiments}, we show extra results
for $r=0.2$ and $K=50$, which further confirm \nameRobust's robustness to $90\%$ outliers.

\myParagraph{Robustness on \pascal} For a simulation setup that is closer to realistic scenarios, we use the CAD models from the \emph{car} category in the \pascal dataset~\cite{Xiang2014WACV-PASCAL+}, which contains $K=9$ CAD models of $N=12$ semantic keypoints. We randomly sample $(\MR,\vt,\vc)$ and add noise and outliers as before, and compare the performance of \nameRobust with other baselines, as shown in Fig.~\ref{fig:simulation-robust}(c). The dominance of \nameRobust over other baselines, and the effectiveness of outlier pruning is clearly seen across the plots. \nameRobust is robust to $70\%$ outliers, while other baselines break at a much lower outlier rate. Note that at $80\%$ outlier rate, there are only two inlier semantic keypoints, making it pathological to estimate shape and pose.

\subsection{Vehicle Pose Estimation on \apollo}
\label{sec:exp-apollo}

\myParagraph{Setup and Baselines}
We evaluate \nameRobust on the \apollo dataset~\cite{Wang19pami-apolloscape,Song19-apollocar3d}.
The \apollo self-driving dataset is a large collection of multi-modal data collected in four different cities in China under varying lighting and road conditions~\cite{Wang19pami-apolloscape}.
Within the dataset, annotations are provided for different perception tasks, ranging from pixel-level semantic segmentation to dense semantic 3D point clouds for the environments.
For our experiments, we specifically use the subset of \apollo named \apolloCar.
\apolloCar consists of high-resolution (3384 $\times$ 2710) images taken from the main \apollo dataset,
with additional 2D annotations of semantic keypoints, ground truth poses, and 3D CAD models of car instances in each frame.
The dataset contains a total of 5277 images, with an average of 11.7 cars per image, and a total of 79 ground-truth CAD models~\cite{Song19-apollocar3d}.
For each car, a total of 66 semantic keypoints were labeled on 2D images.

We compare \nameRobust against DeepMANTA~\cite{Chabot17-deepMANTA}, 3D-RCNN~\cite{Kundu18-3dRCNN}, and GSNet~\cite{Ke20-gsnet}, three recent state of the art methods for 3D vehicle pose estimation.
For our experiments, we use the official splits of the \apolloCar dataset.
Namely, we use the validation split (200 images) for all the quantitative experiments shown below, consistent with the evaluation setups reported in other baseline methods.

We use the 2D semantic keypoints extracted by GSNet~\cite{Ke20-gsnet} as measurements for \nameRobust; 
in particular we use the pretrained weights from~\cite{Ke20-gsnet} \toCheckTwo{and reject keypoints with confidence less than $0.05$}.
\toCheckTwo{For each 2D semantic keypoint, we retrieve the corresponding depth from the depth images provided by \apollo; the resulting technique is labeled \nameRobustApolloDepths.
 We also provide an ablation study to assess the impact of depth and keypoint quality on \nameRobust.
  Towards this goal, we test two variants: \nameRobustGTDepths uses ground-truth depths obtained by ray-tracing the GSNet keypoints using ground-truth 3D car models,
while \nameRobustGTKeypoints uses ground-truth 2D semantic keypoints with ground-truth depths.}
While the 2D semantic keypoint annotations are provided by \apolloCar, the dataset does not provide the corresponding 3D keypoint annotations on the CAD models.
To obtain the necessary 2D-3D correspondences, we manually label the 66 3D semantic keypoints on the 79 CAD models.
We then provide this set of labeled 3D points as the shape library to \nameRobust. 
We use $\lambda = 0.5$ and \toCheckTwo{$\inthr = 0.15$} in \nameRobust.

\myParagraph{Results} %
Table~\ref{tab:apollo-stats} shows the performance of \nameRobust against various baselines.
We use two metrics called A3DP-Rel and A3DP-Abs (for both, the higher the better) following the same definitions used in~\cite{Song19-apollocar3d}.
They are measures of precision with thresholds jointly considering translation, rotation, and 3D shape similarity between estimated cars and ground truth.
A3DP-Abs uses absolute translation thresholds, whereas A3DP-Rel uses relative translation thresholds.
The \textit{mean} column represents the average A3DP-Abs/Rel over 10 different thresholds.
\textit{c-l} represents a loose criterion ($2.8$ m for translation error, $\pi/6$ rad for rotation error, and $0.5$ for shape similarity),
and \textit{c-s} represents a strict criterion ($1.4$ m for translation error, $\pi/12$ rad for rotation error, and $0.75$ for shape similarity).
\toCheckTwo{\nameRobust outperforms the baselines in terms of the \textit{mean} and  \textit{c-s} criteria;
this is partially expected since we use depth information, which is not available to the other methods.
In terms of the strict criterion \textit{c-s}, \nameRobust outperforms competitors by a large amount, confirming that it can retrieve highly accurate estimates.
\nameRobustGTDepths outperforms baselines across all criteria, suggesting that if accurate depth measurements are available,
\nameRobust can roughly double the performance of state-of-the-art methods in terms of \textit{mean} and \textit{c-s} criteria.
\nameRobustGTKeypoints shows the results produced by \nameRobust when using ground-truth keypoint detections and depths:
 this is the best potential accuracy
\nameRobust could achieve if provided with perfect keypoint detections.
In our tests, the average number of inliers produced by GSNet is 21.8\%,\footnote{We define true inliers as 2D keypoint detections such that there exists a ground-truth annotated keypoint with the same ID within a radius of 5 pixels.}
  showing that 
there is still a large margin of improvement for state-of-the-art methods in semantic keypoint detection.
}

\begin{table}[ht!]
  \centering
    \begin{minipage}[b]{0.99\linewidth}
    \centering
      \begin{tabular}{llllrrr}
      \toprule
                            & \multicolumn{3}{c}{A3DP-Rel $\uparrow$}    & \multicolumn{3}{c}{A3DP-Abs $\uparrow$}          \\
      \midrule
                            & mean & c-l  & c-s               & mean & c-l & c-s \\
      \midrule
      DeepMANTA~\cite{Chabot17-deepMANTA}  & 16.0 & 23.8 & 19.8              & \secondBest{20.1} & 30.7          & \secondBest{23.8}            \\
      3D-RCNN~\cite{Kundu18-3dRCNN}        & 10.8 & 17.8 & 11.9              & 16.4 & 29.7          & 19.8            \\
      GSNet~\cite{Ke20-gsnet}              & \secondBest{20.2} & \best{40.5} & \secondBest{19.9}              & 18.9 & \best{37.4} & 18.4            \\
      \nameRobustApolloDepths    & \toCheck{\best{25.9}} & \toCheck{\secondBest{35.7}} &  \toCheck{\best{33.7}}                         &
      \toCheck{\best{22.4}}   & \toCheck{\secondBest{34.7}}  & \toCheck{\best{31.6}}    \\
      \midrule
      \nameRobustGTDepths    & \toCheck{36.0} & \toCheck{45.4} &  \toCheck{43.6}                         &
      \toCheck{35.3}   & \toCheck{44.2}  & \toCheck{43.2}    \\
      \nameRobustGTKeypoints & \toCheck{64.5} & \toCheck{88.1} &  \toCheck{86.0}                        & \toCheck{64.3}            & \toCheck{88.1}  & \toCheck{86.1} \\
      \bottomrule
      \end{tabular}%
      \end{minipage} %
      \vspace{-1mm}
      \caption{Evaluation of \nameRobust on \apollo. Results for DeepMANTA,3D-RCNN, and GSNet are taken from~\cite{Ke20-gsnet}. 
      The best result for each column is highlighted in boldface. \label{tab:apollo-stats}\vspace{-2mm}}   
\end{table}

\toCheck{
Table~\ref{tab:timing} shows the timing breakdown for \nameRobust.
We also report the timing for the GSNet keypoint detection from~\cite{Ke20-gsnet} for completeness.
In our current implementation of \nameRobust, the max-clique pruning is in C++ and its runtime is negligible, 
while \gnc is implemented in Python.
All tests are run on a Linux
computer with an Intel i9-9920X CPU at 3.5 GHz.
}
\vspace{-2mm}

\begin{table}[ht!]
  \centering
    \begin{minipage}[b]{0.99\linewidth}\centering
        \begin{tabular}{c|c|c}
            \toprule
             \multirow{2}{*}{GSNet keypoint detection}  & \multicolumn{2}{c}{\nameRobust}  \\
              											&  Max-clique & \gnc  \\
            \midrule
            0.45 \si{s}   &  2 \si{ms} & 0.45 \si{s}  \\
            \bottomrule
        \end{tabular}
        \vspace{-2mm}
      \caption{Timing breakdown for \nameRobust.\vspace{-4mm}}
      \label{tab:timing}
    \end{minipage}%
\end{table}

\section{Conclusion}
\label{sec:conclusion}

We proposed \name, the first certifiably optimal solver for the estimation of the pose and shape of 
3D objects from 3D keypoint detections. While existing iterative methods get stuck in local minima corresponding to poor estimates, 
\name leverages a tight and fixed-size SDP relaxation to compute certifiably optimal estimates.
We also design a second algorithm, \nameRobust, that adds an outlier rejection layer to \name and 
is able to estimate accurate pose and shape parameters in the face of large amounts  of outliers (\eg~\toCheck{\outPaceSharp} of the measurements are incorrect).
 The proposed methods dominate the state of the art in terms of accuracy and robustness on both Monte Carlo
 simulations and on the \pascal dataset.
 Moreover, we show that \nameRobust can be successfully combined with deep-learned
  keypoint detectors, \toCheckTwo{and leads to  highly accurate vehicle pose estimates
 in  
  the \apollo driving datasets.}

\section*{Acknowledgments}

\edit{This work was partially funded by ARL DCIST CRA W911NF-17-2-0181, ONR RAIDER N00014-18-1-2828, NSF CAREER award
``Certifiable Perception for Autonomous Cyber-Physical Systems'', and Lincoln Laboratory's Resilient Perception in Degraded Environments program. We would like to also thank Charleen Tan for helping with labeling 3D keypoints.}

 \bibliographystyle{abbrvnat} 
\bibliography{sections/refs.bib,sections/myRefs.bib}

\isExtended{
\appendices 
\renewcommand{\theequation}{A\arabic{equation}}
\renewcommand{\thefigure}{A\arabic{figure}}
\setcounter{equation}{0}
\setcounter{figure}{0}

\section{Problem~\eqref{eq:probOutFree-3D3Dcatlevel} is a MAP Estimator when the Measurement Noise is Gaussian}
\label{sec:app-mapOutlierFree}

Here we prove that the optimization in eq.~\eqref{eq:probOutFree-3D3Dcatlevel} is a 
\emph{maximum a posteriori} (MAP) estimator when the measurement noise $\vepsilon (i)$ in~\eqref{eq:generativeModel} 
follows a zero-mean Gaussian with covariance $\frac{1}{w_i} \eye_3$ (where $\eye_3$ is the 3-by-3 identity matrix) and we have a zero-mean Gaussian prior with covariance $\frac{1}{\lambda} \eye_K $ over the shape parameters $\vc$. 
Mathematically:
\bea
\label{eq:distMAP1a}
\!\!\prob{\vepsilon (i)} = \kappa_\epsilon \exp\left(\!-\frac{w_i}{2} 
\|\vepsilon (i)\|^2\!
\right), \\
\label{eq:distMAP1b}
\prob{\vc} = 
\kappa_c \exp\left( \!-\frac{\lambda}{2} 
\|\vc\|^2\!
\right),
\eea
where $\kappa_\epsilon$ and $\kappa_c$ are suitable normalization constants that are irrelevant for the following derivation.

A MAP estimator for the unknown parameters $\vxx \triangleq \{\MR, \vt, \vc\}$ (belonging to a suitable domain $\domainX$) given measurements~$\vy(i)$ ($i=1,\ldots,N$) is defined as the maximum of the posterior distribution 
$\prob{ \vxx | \vy(1)\;\ldots\; \vy(N)}$:
\beq
\label{eq:MAP}
\argmax_{\vxx \in \domainX} \prob{ \vxx | \vy(1)\;\ldots\; \vy(N)}
= \\ 
\argmax_{\vxx \in \domainX} \prod_{i=1}^N \prob{ \vy(i) | \vxx} \prob{ \vxx }
\eeq 
where on the right we applied Bayes rule and used the standard assumption of independent measurements.
Using~\eqref{eq:distMAP1a} and~\eqref{eq:generativeModel} we obtain:
\bea
\label{eq:pyx}
\prob{ \vy(i) | \vxx}  = \kappa_\epsilon \exp\left( -\frac{w_i}{2} 
\left\| \vy(i) \!-\! \MR \sum_{k=1}^{\nrShapes} c_{k} \vb_{k}(i) \!-\! \vt \right\|^{2} 
\right).
\eea
Moreover, assuming we only have a prior on $\vc$:
\bea
\label{eq:px}
\prob{ \vxx } = \prob{\vc} = 
\kappa_c \exp\left( -\frac{\lambda}{2} 
\|\vc\|^2
\right).
\eea
Substituting~\eqref{eq:pyx} and~\eqref{eq:px} back into~\eqref{eq:MAP} and 
observing that the maximum of the posterior is the same as the minimum of the negative logarithm of the 
posterior:
\bea
\argmax_{\vxx \in \domainX} \prod_{i=1}^N \prob{ \vy(i) | \vxx} \prob{ \vxx } =  \\
\argmin_{\vxx \in \domainX} \sum_{i=1}^N -\log\prob{ \vy(i) | \vxx} -\log \prob{ \vxx } = \\
\argmin_{\substack{\MR \in \SOthree, \\ \vt \in \Real{3}, \vc \in \Real{\nrShapes}, \\ \ones\tran \vc  = 1 } } \sum_{i=1}^N \frac{w_i}{2} 
\left\| \vy(i) - \MR \sum_{k=1}^{\nrShapes} c_{k} \vb_{k}(i) - \vt \right\|^{2} \\
 +\frac{\lambda}{2} 
\|\vc\|^2 + \text{constants}
\eea 
which, after dropping constant multiplicative and additive factors, can be seen to match eq.~\eqref{eq:probOutFree-3D3Dcatlevel}, proving the claim.

\section{Problem~\eqref{eq:robust-3D3Dcatlevel} is a MAP Estimator when the Measurement Noise is Heavy-Tailed}
\label{sec:app-mapOutliers}

Here we prove that the optimization in eq.~\eqref{eq:robust-3D3Dcatlevel} 
with a truncated least square loss $\rho(r) = \min(r^2, \barcsq)$ is a 
\emph{maximum a posteriori} (MAP) estimator when the measurement noise $\vepsilon (i)$ in~\eqref{eq:generativeModel} 
follows a max-mixture distribution, where we replace the tails of a Gaussian with a uniform distribution ---a model
we borrow from~\cite{Antonante20arxiv-outlierRobustEstimation}.
Mathematically:
\beq\label{eq:normalWithUniformTail}
\prob{\vepsilon (i)} =\left\{\begin{array}{ll}
\kappa_\epsilon \exp\left(\!-\frac{1}{2} 
\|\vepsilon (i)\|^2\!
\right), & \|\vepsilon (i)\| < \inthr,\\
\kappa_\epsilon \exp\left(\!-\frac{1}{2} 
\inthr^2\!
\right), & \|\vepsilon (i)\| \in [\inthr,\alpha],\\
0, & \text{otherwise},
\end{array}\right.
\eeq
where $\inthr$ is the maximum noise for an inlier, $\kappa_\epsilon$ is a normalization constant, and $\alpha$ 
defines the support of the uniform distribution (both $\kappa_\epsilon$ and $\alpha$ are irrelevant for the  derivation); in~\eqref{eq:normalWithUniformTail} ---without loss of generality---  we assumed unit covariance for the Gaussian.
 Intuitively,~eq.~\eqref{eq:normalWithUniformTail} describes a Gaussian distribution for errors below $\inthr$, 
 but for errors larger than $\inthr$ the Gaussian tails %
 have been substituted by a uniform distribution (observe that  $\kappa_\epsilon \exp\left(\!-\frac{1}{2} 
\inthr^2\! \right)$ is a constant). 
Then the proof trivially follows from~\cite[Proposition 5]{Antonante20arxiv-outlierRobustEstimation} 
(the expression of the shape priors remains the same as Appendix~\ref{sec:app-mapOutlierFree}).

\section{Closed-form Shape Estimation: \\ Proof of Proposition~\ref{prop:shapeEstimation}}
\label{sec:app-shapeEstimation}
Fixing $\MR$, the Lagrangian of the linearly constrained linear least squares problem~\eqref{eq:clsofc} is:

\bea
\calL = \norm{\barMB \vc - (\eye_N \kron \MR\tran) \barvy}^2 + \lambda \norm{\vc}^2 + \gamma (\ones\tran \vc - 1)
\eea
where $\gamma \in \Real{}$ is the multiplier associated with the constraint $\ones\tran\vc = 1$~\cite{Boyd04book}. Observe that problem~\eqref{eq:clsofc} has a single equality constraint and trivially satisfies the linear independence constraint qualification (LICQ), therefore, any optimal solution must satisfy the following KKT conditions:
\bea
\hspace{-6mm} \nabla_{\vc}\calL = 2(\barMB\tran\barMB + \lambda \eye_K)\vc + \gamma \ones - 2\barMB\tran(\eye_N \kron \MR\tran)\barvy = \zero \\
\hspace{-6mm} \nabla_{\gamma} \calL = \ones\tran \vc - 1 = 0
\eea
which can be written compactly as the following linear system of equations:
\bea \label{eq:linearKKTsystemofc}
\hspace{-4mm} \bmat{cc}
2(\barMB\tran\barMB + \lambda \eye_K) & \ones \\
\ones\tran & 0
\emat
\!
\bmat{c}
\vc \\ \gamma \emat \!=\!
\bmat{c}
2\barMB\tran(\eye_N \kron \MR\tran)\barvy \\
1
\emat.
\eea
Now let
\bea
\MHtl \triangleq 2(\barMB\tran\barMB + \lambda \eye_K) \in \pd^\nrShapes, \\
\MH \triangleq \bmat{cc}
\MHtl & \ones \\
\ones\tran & 0
\emat \in \sym^{\nrShapes+1},
\eea
where $\pd^\nrShapes$ denotes the set of positive definite matrices of size $\nrShapes$.
{Note that the inverse of $\MH$ exists because $\MHtl$ is positive definite and invertible ($\MHtl\inv$ is also positive definite):}
\bea
\MH\inv = 
\bmat{cc}
\MHtl\inv - \frac{\MHtl\inv \ve \ones\tran \MHtl\inv }{\ones\tran \MHtl\inv \ones} & \frac{\MHtl\inv \ones}{\ones\tran \MHtl\inv \ones} \\
\frac{\ones\tran \MHtl\inv}{\ones\tran \MHtl\inv \ones} & - \frac{1}{\ones\tran \MHtl\inv \ones}
\emat.
\eea
Therefore the optimal $\vc$ can be obtained from~\eqref{eq:linearKKTsystemofc} as:
\bea 
\vc^{\star} (\MR) = 2\MG \barMB\tran (\eye_N \kron \MR\tran) \barvy + \vg,
\eea
where
\bea
\MG \triangleq \MHtl\inv - \frac{\MHtl\inv \ones \ones\tran \MHtl\inv }{\ones\tran \MHtl\inv \ones},\quad 
\vg \triangleq \frac{\MHtl\inv \ones}{\ones\tran \MHtl\inv \ones}
\eea
proving Proposition~\ref{prop:shapeEstimation}.

\section{Certifiably Optimal Rotation Estimation: \\ Proof of Proposition~\ref{prop:optRotation} 
and Corollary~\ref{cor:optRotation-relax}}
\label{sec:app-rotEst-details}

Let us first develop the cost function of problem~\eqref{eq:nonconvexR} as a quadratic function of $\vr \triangleq \vectorize{\MR}$:
\bea
\norm{\MM (\eye_N \kron \MR\tran)\barvy + \vh}^2 \\
= \norm{\MM \vectorize{\MR\tran \MY} + \vh}^2 \\
= \norm{\MM (\MY\tran \kron \eye_3) \vectorize{\MR\tran} + \vh}^2 \\
= \norm{\MM (\MY\tran \kron \eye_3) \MP \vr + \vh}^2 \\
= \vrhomo \tran \MQ \vrhomo
\eea
where $\MP \in \Real{9 \times 9}$ is the following permutation matrix
\bea
(1,1,1), 
(2,4,1),
(3,7,1), \\
(4,2,1),
(5,5,1),
(6,8,1), \\
(7,3,1),
(8,6,1),
(9,9,1),
\eea
with the triplet $(i,j,v)$ defining the nonzero entries of $\MP$ (\ie~$\MP_{ij} = v$), such that:
\bea
\vectorize{\MR\tran} \equiv \MP \vectorize{\MR}
\eea 
always holds, $\MY$ and $\vrhomo$ are defined as: 
\bea
\MY \triangleq \bmat{ccc}
\barvy(1) & \cdots & \barvy(N)
\emat \in \Real{3 \times N}, \\
\vrhomo \triangleq \bmat{cc} 1 & \vr\tran \emat\tran \in \Real{10},
\eea
and $\MQ \in \sym^{10}$ can be assembled as follows:
\bea
 \MQ \triangleq 
\bmat{cc}
\vh\tran \vh & \vh\tran\MM(\MY\tran\kron\eye_3)\MP \\
\star & \MP\tran(\MY \kron \eye_3)\MM\tran \MM (\MY\tran \kron \eye_3) \MP
\emat.
\eea

Now that the objective function of~\eqref{eq:nonconvexR} is quadratic in $\vr$ ($\MR$), we can write problem~\eqref{eq:nonconvexR} equivalently as the \emph{quadratically constrained quadratic program} (QCQP) in~\eqref{eq:categoryQCQP},
where $\MA_i \in \sym^{10}, i=1,\dots,15$, are the constant matrices that define the quadratic constraints associated with $\MR \in \SOthree$~\cite[Lemma 5]{Yang20cvpr-shapeStar}. For completeness, we give the expressions for $\MA_i$'s:
\bea
\MA_0: (1,1,1) \nonumber \\
\MA_1-\MA_3: \text{ columns have unit norm} \nonumber \\
\MA_1: (1,1,1),(2,2,-1),(3,3,-1),(4,4,-1) \nonumber \\
\MA_2: (1,1,1),(5,5,-1),(6,6,-1),(7,7,-1) \nonumber \\
\MA_3: (1,1,1),(8,8,-1),(9,9,-1),(10,10,-1) \nonumber \\
\MA_4-\MA_6: \text{ columns are mutually orthogonal} \nonumber \\
\MA_4: (2,5,1),(3,6,1),(4,7,1) \nonumber \\
\MA_5: (2,8,1),(3,9,1),(4,10,1) \nonumber \\
\MA_6: (5,8,1),(6,9,1),(7,10,1) \nonumber \\
\MA_7-\MA_{15}: \text{ columns form right-handed frame} \nonumber \\
\MA_7: (3,7,1),(4,6,-1),(1,8,-1) \nonumber \\
\MA_8: (4,5,1),(2,7,-1),(1,9,-1) \nonumber \\
\MA_9: (2,6,1),(1,10,-1),(3,5,-1) \nonumber \\
\MA_{10}: (6,10,1),(1,2,-1),(7,9,-1) \nonumber \\
\MA_{11}: (7,8,1),(5,10,-1),(1,3,-1) \nonumber \\
\MA_{12}: (5,9,1),(1,4,-1),(6,8,-1) \nonumber \\
\MA_{13}: (4,9,1),(3,10,-1),(1,5,-1) \nonumber \\
\MA_{14}: (2,10,1),(1,6,-1),(4,8,-1) \nonumber \\
\MA_{15}: (3,8,1),(2,9,-1),(1,7,-1) \nonumber
\eea
where the triplets $(i,j,v)$ define the \emph{diagonal and upper triangular} nonzero entries of a symmetric matrix (\ie~$\MA_{ij} = \MA_{ji} = v$ with $i \leq j$).

\section{Shor's Semidefinite Relaxation and \\ Relative Duality Gap}
\label{sec:app-sdp-relaxation-gap}
To see why problem~\eqref{eq:categoryQCQPrelax} is a convex relaxation for problem~\eqref{eq:categoryQCQP}, let us first create a matrix variable
\bea \label{eq:MXfactor}
\MX = \vrhomo \vrhomo\tran \in \sym^{10},
\eea
and notice that $\MX$ satisfies
\bea
\MX \succeq 0, \quad \rank{\MX} = 1.
\eea
Moreover, if $\MX \succeq 0, \rank{\MX} = 1$ then $\MX$ must have a factorization of the form~\eqref{eq:MXfactor}. Therefore, the non-convex QCQP~\eqref{eq:categoryQCQP} is equivalent to the following rank-constrained matrix optimization problem:
\bea \label{eq:rankconstrained}
\min_{\MX \in \sym^{10}} & \trace{\MQ \MX} \\
\subject & \trace{\MA_0 \MX} = 1,\\
&  \trace{\MA_i \MX} = 0, \forall i=1,\dots,15,\\
& \MX \succeq 0, \\
& \rank{\MX} = 1, \label{eq:rankoneconstraint}
\eea
where $\MA_0 \in \sym^{10}$ is an all-zero matrix except the top-left entry being 1 (to enforce that the first entry of $\vrhomo$ is 1), and we have used the fact that
\bea
\vrhomo\tran \MA \vrhomo = \trace{\vrhomo\tran \MA \vrhomo} = \trace{\MA \vrhomo \vrhomo\tran} = \trace{\MA \MX}.
\eea 
Now observe that the only nonconvex constraint in problem~\eqref{eq:rankconstrained} is the rank constraint~\eqref{eq:rankoneconstraint}, and the SDP relaxation~\eqref{eq:categoryQCQPrelax} is obtained by simply removing the rank constraint. 

In practice, we solve the convex problem~\eqref{eq:categoryQCQPrelax} and obtain an optimal solution $\MX^\star$, if $\rank{\MX^\star} = 1$, then the optimal solution of problem~\eqref{eq:categoryQCQPrelax} is unique (the rationale behind this is that interior-point methods converge to a maximum rank solution~\cite{DeKlerk06book-IPMSDP}) and it actually satisfies the rank constraint that has been dropped. Therefore, in this situation, we say the convex relaxation is tight and the global optimal solution to the nonconvex problem~\eqref{eq:categoryQCQP} can be obtained from the rank-one factorization of $\MX^\star$. 

{\bf Relative duality gap}. Checking if the solution is rank one can sometimes be sensitive to numerical thresholds, therefore, an alternative way to check the quality of the relaxation is to compute the relative duality gap. Let $\MX^\star$ be a solution of the SDP relaxation~\eqref{eq:categoryQCQPrelax} and let $\fsdp \triangleq \trace{\MQ \MX^\star}$ be the optimal cost. Let $\hat{\vr} \in \SOthree$ be a rounded solution from $\MX^\star$ (the rounding can be done by closed-form projection to $\SOthree$~\cite{Yang20neurips-certifiablePerception}), and let $\fest \triangleq [1,\hat{\vr}\tran] \MQ [1,\hat{\vr}\tran]\tran$ be the cost of the non-convex problem~\eqref{eq:categoryQCQP} evaluated at the rounded solution $\hat{\vr}$, then we have:
\bea
\fsdp \leq f^\star \leq \fest,
\eea
where $f^\star$ is the true global optimum of the nonconvex problem~\eqref{eq:categoryQCQP}, the first inequality follows from the fact that problem~\eqref{eq:categoryQCQPrelax} is a convex relaxation and the second inequality follows from the fact that $f^\star$ is the global minimum. We then compute the relative duality gap
\bea
\eta \triangleq \frac{\fest - \fsdp}{\fest},
\eea
which is informative of the suboptimality of the rounded solution. In particular, if $\eta \approx 0$, then $\hat{\vr}$ is certified to be the globally optimal solution.

\vspace{5mm}
\section{Minimum and Maximum Distances \\ between Convex Hulls}
\label{sec:app-bmin-bmax}
Recall from eq.~\eqref{eq:definebminbmax} the definitions of $b_{ij}^{\min}$ and $b_{ij}^{\max}$:
\bea
b_{ij}^{\min} = \min_{\vc \geq 0, \ones\tran\vc=1} \norm{\sum_{k=1}^K c_k (\vb_k(j) - \vb_k(i))}, \label{eq:definebmin}\\
b_{ij}^{\max} = \max_{\vc \geq 0, \ones\tran\vc=1} \norm{\sum_{k=1}^K c_k (\vb_k(j) - \vb_k(i))}, \label{eq:definebmax}
\eea
and let us use the following shorthand:
\bea
\vb_{k,ij} \triangleq \vb_k(j) - \vb_k(i), \\
\MB_{ij} \triangleq \bmat{ccc} \vb_{1,ij} & \cdots & \vb_{K,ij} \emat \in \Real{3 \times K},
\eea
to write problems~\eqref{eq:definebmin} and~\eqref{eq:definebmax} compactly as:
\bea
\hspace{-4mm} b_{ij}^{\min} = \min_{\vc \geq 0, \ones\tran\vc=1} \norm{\MB_{ij} \vc}, \quad b_{ij}^{\max} = \max_{\vc \geq 0, \ones\tran\vc=1} \norm{\MB_{ij} \vc}.
\eea

{\bf Compute $b^{\max}_{ij}$}. Because $\norm{\MB_{ij} \vc}$ is a convex function of $\vc$, and the maximum of a convex function over a polyhedral set (in our case, the standard simplex $\Delta_K \triangleq \{\vc \in \Real{K}: \vc \geq 0, \ones\tran \vc=1 \}$) is always obtained at one of the vertices of the polyhedron~\cite[Corollary 32.3.4]{Rockafellar70book-convexanalysis}, we have:
\bea
b_{ij}^{\max} = \max_{k} \norm{\vb_{k,ij}},
\eea
since the vertices of $\Delta_K$ are the vectors $\ve_k,k=1,\dots,K$, where $\ve_k$ is one at its $k$-th entry and zero anywhere else.

{\bf Compute $b^{\min}_{ij}$}. Observe that computing the minimum of $\norm{\MB_{ij} \vc}$ is equivalent to computing 
the minimum of $\norm{\MB_{ij} \vc}^2 = \vc\tran (\MB_{ij}\tran \MB_{ij}) \vc$
because the quadratic function $f(x) = x^2$ is monotonically increasing in the interval $[0,\infty]$, and hence we first solve the following convex quadratic program (QP):
\bea \label{eq:QPofbmin}
\min_{\vc \in \Real{K}} & \vc\tran (\MB_{ij}\tran \MB_{ij}) \vc \\
\subject & \vc \geq 0, \quad \ones\tran \vc = 1
\eea
and then compute $b_{ij}^{\min} = \norm{\MB_{ij} \vc^\star}$ from the solution $\vc^\star$ of the QP. Note that the QP~\eqref{eq:QPofbmin} can be solved in milliseconds for large $K$, so pre-computing $b_{ij}^{\min}$ for all $1\leq i < j \leq N$ is still tractable even when $N$ is large.

\section{Alternation Approach}
\label{sec:app-alternationApproach}

In Sections~\ref{sec:shape-cat-level}-\ref{sec:rotation-cat-level} of the main paper, we presented a certifiably optimal solver to solve the joint shape and rotation $(\vc,\MR)$ problem~\eqref{eq:translation-free-problem} (after eliminating the translation $\vt$). Here we describe a baseline method that solves problem~\eqref{eq:translation-free-problem} using \emph{alternating minimization} (\altern), a heuristic that is popular in related works on 3D shape reconstruction from 2D landmarks~\cite{Lin14eccv-modelFitting,Gu06cvpr-faceAlignment,Ramakrishna12eccv-humanPose}, but offers no optimality guarantees. Towards this goal, let us denote the cost function of~\eqref{eq:translation-free-problem} as $f(\MR,\vc)$; the \altern method starts with an initial guess $(\MR^{(0)}, \vc^{(0)})$ (default $\MR^{(0)} = \eye_3, \vc^{(0)} = \zero$), and performs the following two steps at each iteration $\tau$:
\begin{enumerate}
	\item Optimize $\vc$:
	\bea
	\vc^{(\tau)} = \argmin_{\vc \in \Real{\nrShapes}, \ones\tran \vc = 1} f(\MR^{(\tau-1)},\vc),
	\eea
	which is a linearly constrained linear least squares problem and can be solved by the closed-form solution~\eqref{eq:optimalvcofR}.

	\item Optimize $\MR$:
	\bea
	\MR^{(\tau)} = \argmin_{\MR \in \SOthree} f(\MR, \vc^{(\tau)}),
	\eea
	which can be cast as an instance of Wahba's problem~\cite{Yang19iccv-QUASAR} and can be solved in closed form using singular value decomposition~\cite{markley1988jas-svdAttitudeDeter}.
\end{enumerate}
The \altern method stops when the cost function converges,~\ie~$|f(\MR^{(\tau)},\vc^{(\tau)}) - f(\MR^{(\tau-1)},\vc^{(\tau-1)})| < \epsilon$ for some small threshold $\epsilon > 0$, or when $\tau$ exceeds the maximum number of iterations (\eg~$1000$).

\section{Extra Experimental Results}
\label{sec:app-experiments}
In Section~\ref{sec:exp-optimality-robustness}, we demonstrated the robustness of \nameRobust to $92\%$ outlier rates when $N=100$, $K=10$ and $r=0.1$. Here we show extra results when $K$ and $r$ are increased. Fig.~\ref{fig:app-simulation-robust}(a) shows the results for $N=100$, $K=10$ and $r=0.2$. One can see that as the intra-class variation radius $r$ is increased, the compatibility check becomes less effective, leading to a slight decrease in the robustness of \nameRobust against outliers --- \nameRobust is still robust up to $90\%$ outlier rate while has two failures at $91\%$ outlier rate. However, \nameRobust still outperforms \irlstls and \irlsgm by a large margin. Fig.~\ref{fig:app-simulation-robust}(b) shows the results for $N=100$, $K=50$ and $r=0.1$. We see that \nameRobust is robust up to $91\%$ outlier rate while encounters two failures at $92\%$ outlier rate. Finally, when $K=50$ and $r=0.2$ (Fig.~\ref{fig:app-simulation-robust}(c)), \nameRobust is robust to $80\%$ outlier rate.

\renewcommand{\mpwfour}{4.6cm}
\renewcommand{\myhspace}{\hspace{-3.5mm}}
\begin{figure*}[!h]
	\begin{center}
	\begin{minipage}{\textwidth}
	\begin{tabular}{cccc}%
		\myhspace \hspace{-3mm}
			\begin{minipage}{\mpwfour}%
			\centering%
			\includegraphics[width=\columnwidth]{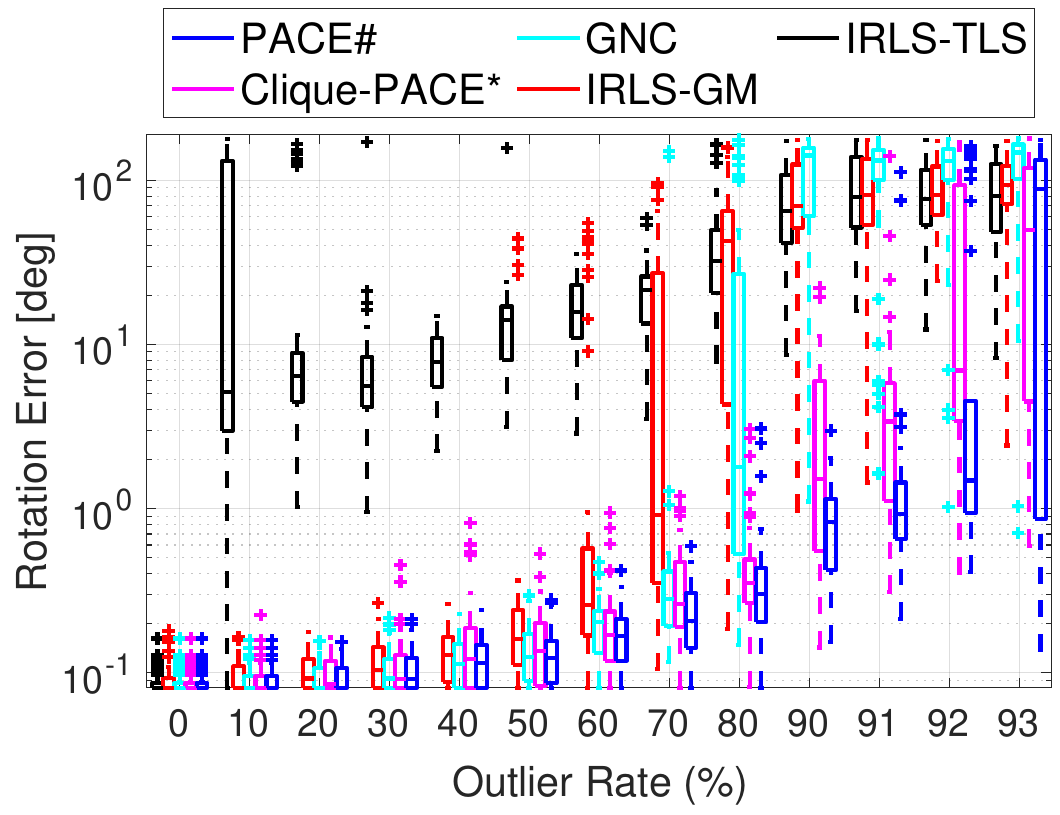}
			\end{minipage}
		&   \myhspace
			\begin{minipage}{\mpwfour}%
			\centering%
			\includegraphics[width=\columnwidth]{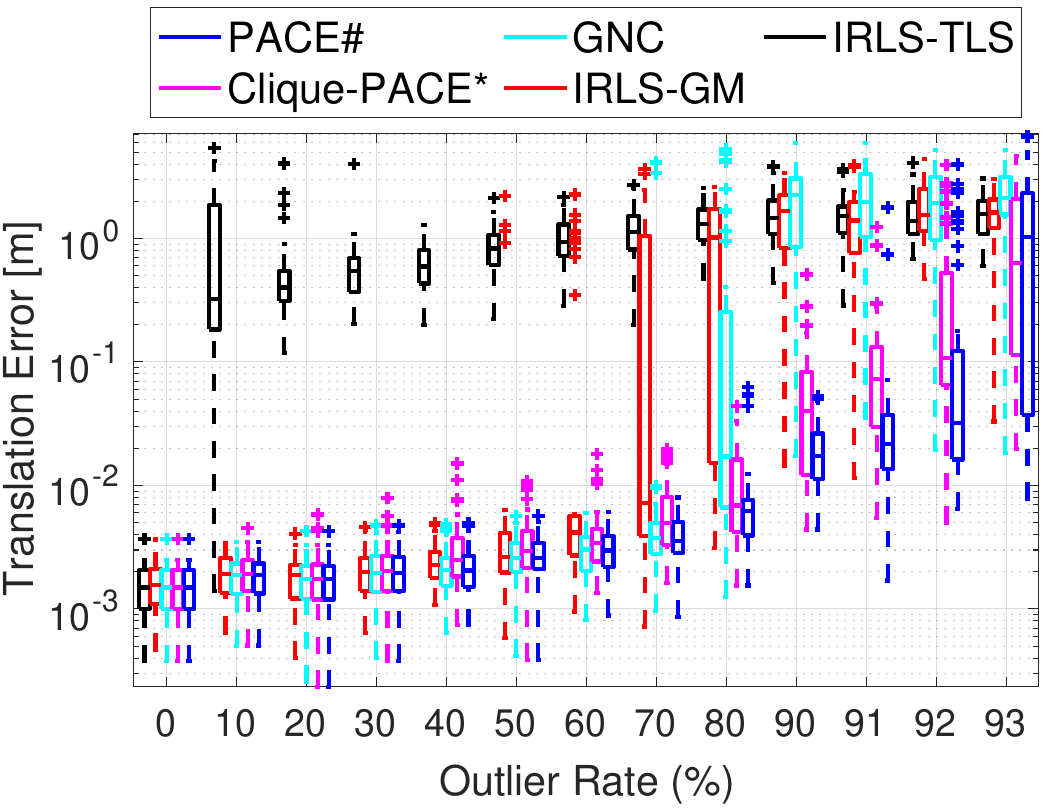}
			\end{minipage}
		&   \myhspace
			\begin{minipage}{\mpwfour}%
			\centering%
			\includegraphics[width=\columnwidth]{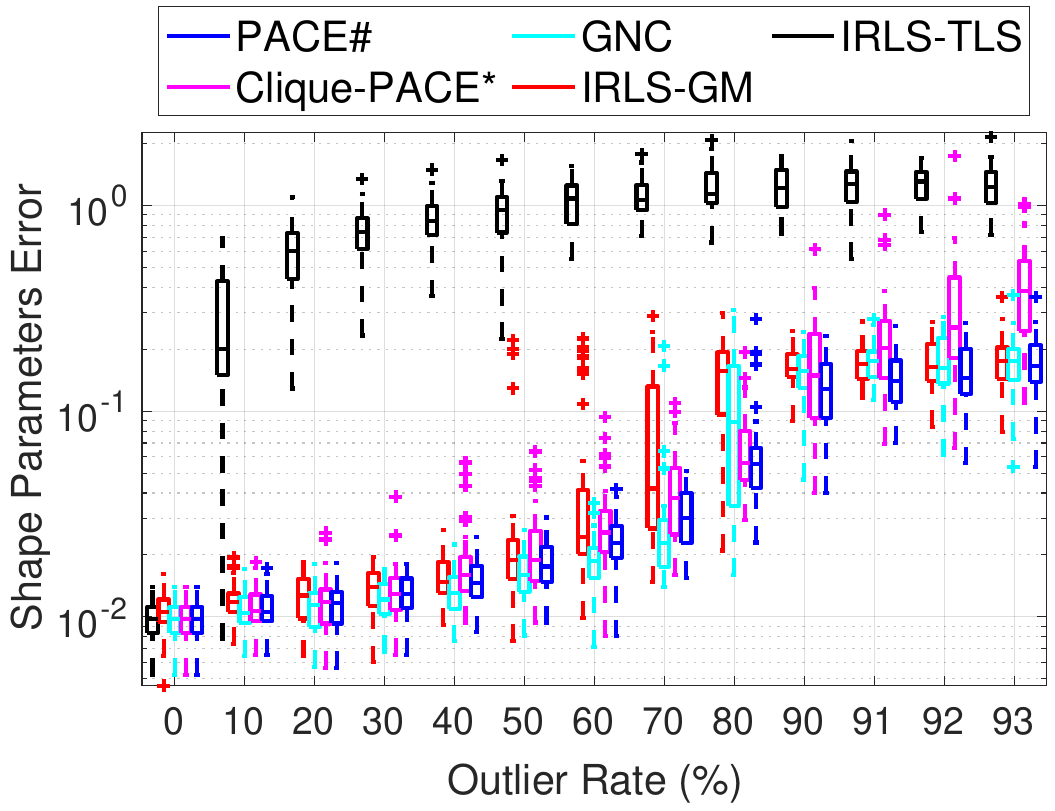}
			\end{minipage}
		&   \myhspace
			\begin{minipage}{\mpwfour}%
			\centering%
			\includegraphics[width=\columnwidth]{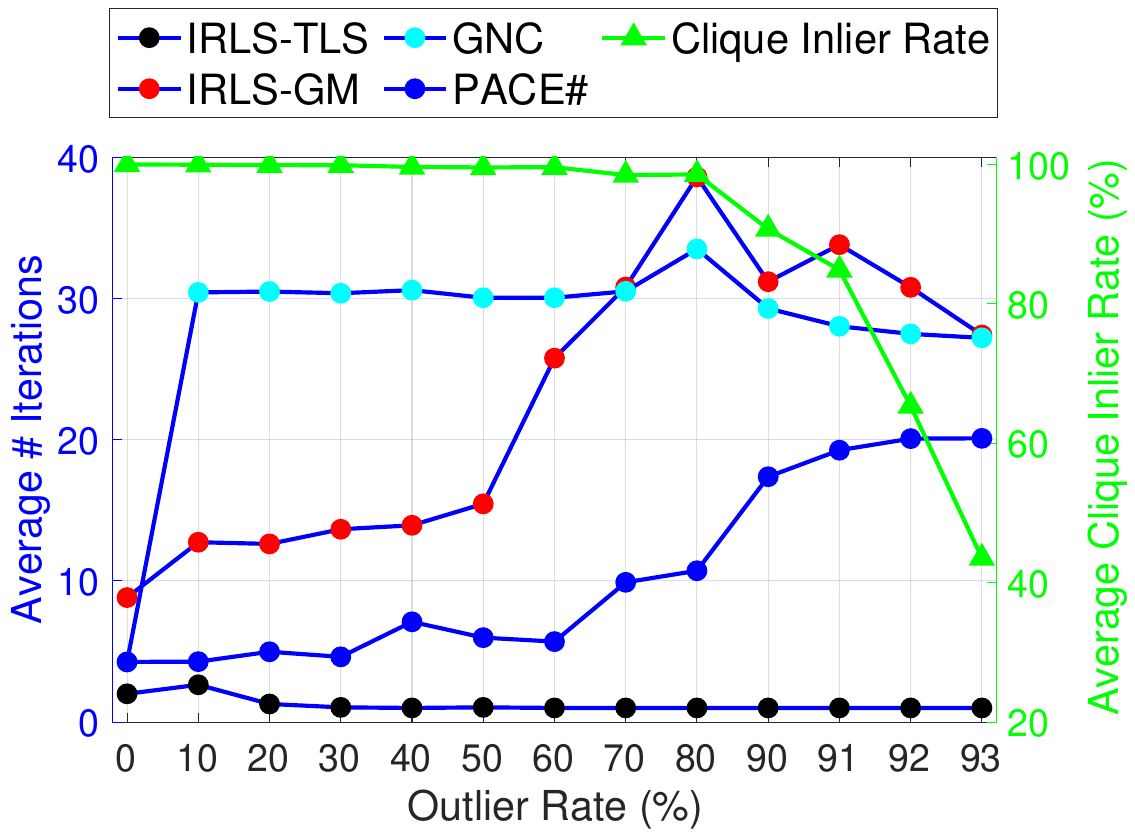}
			\end{minipage}
		\\
		\multicolumn{4}{c}{\smaller (a) Robustness of \nameRobust against increasing outliers on random simulated data: $N=100$, $K=10$, $r=0.2$. \vspace{1mm}}
		\\
		\myhspace \hspace{-3mm}
			\begin{minipage}{\mpwfour}%
			\centering%
			\includegraphics[width=\columnwidth]{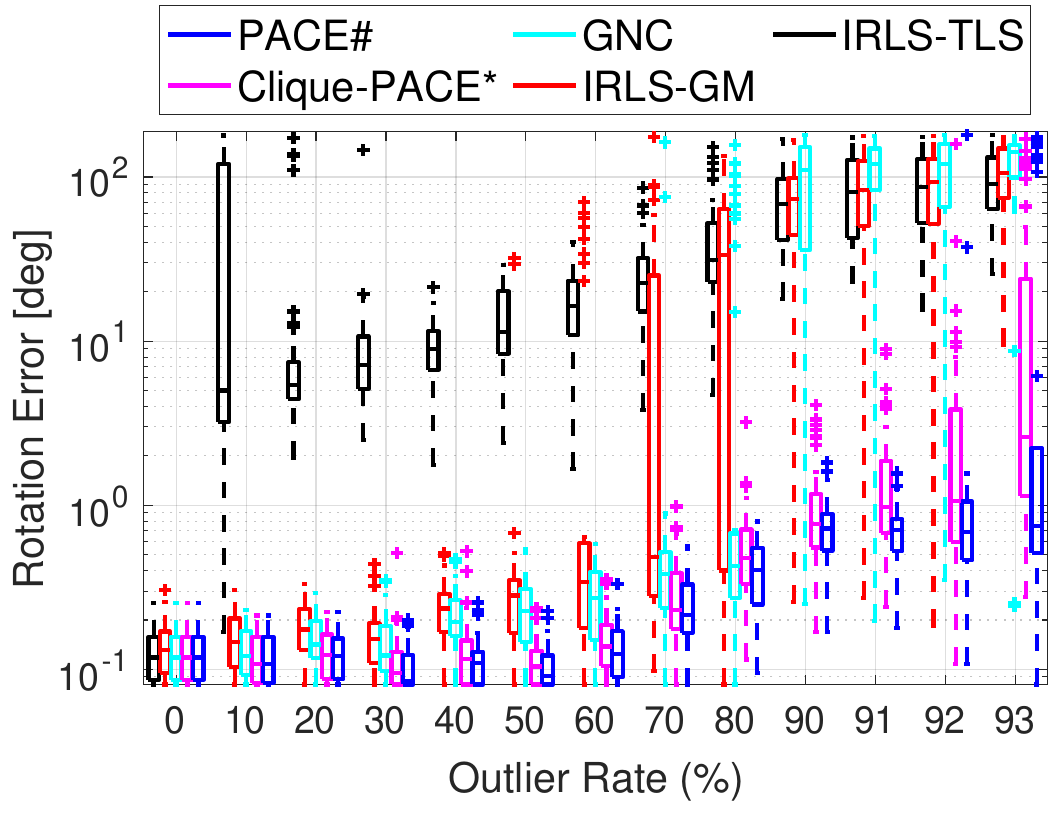}
			\end{minipage}
		&   \myhspace
			\begin{minipage}{\mpwfour}%
			\centering%
			\includegraphics[width=\columnwidth]{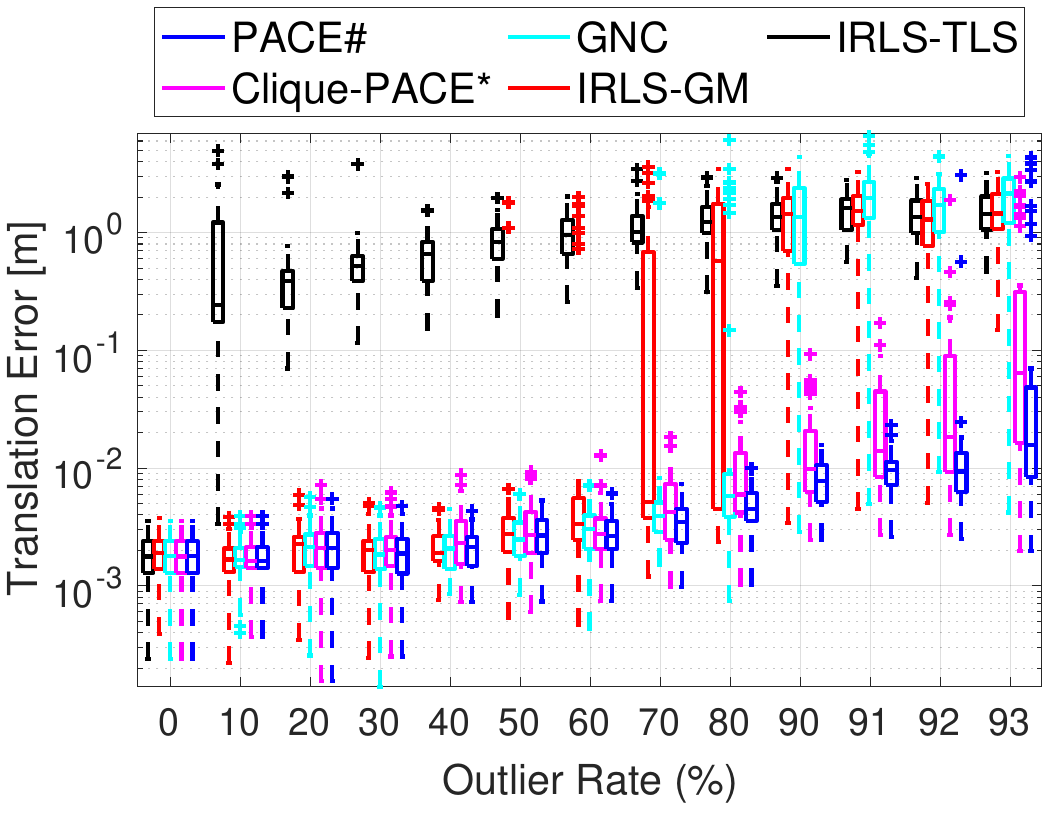}
			\end{minipage}
		&   \myhspace
			\begin{minipage}{\mpwfour}%
			\centering%
			\includegraphics[width=\columnwidth]{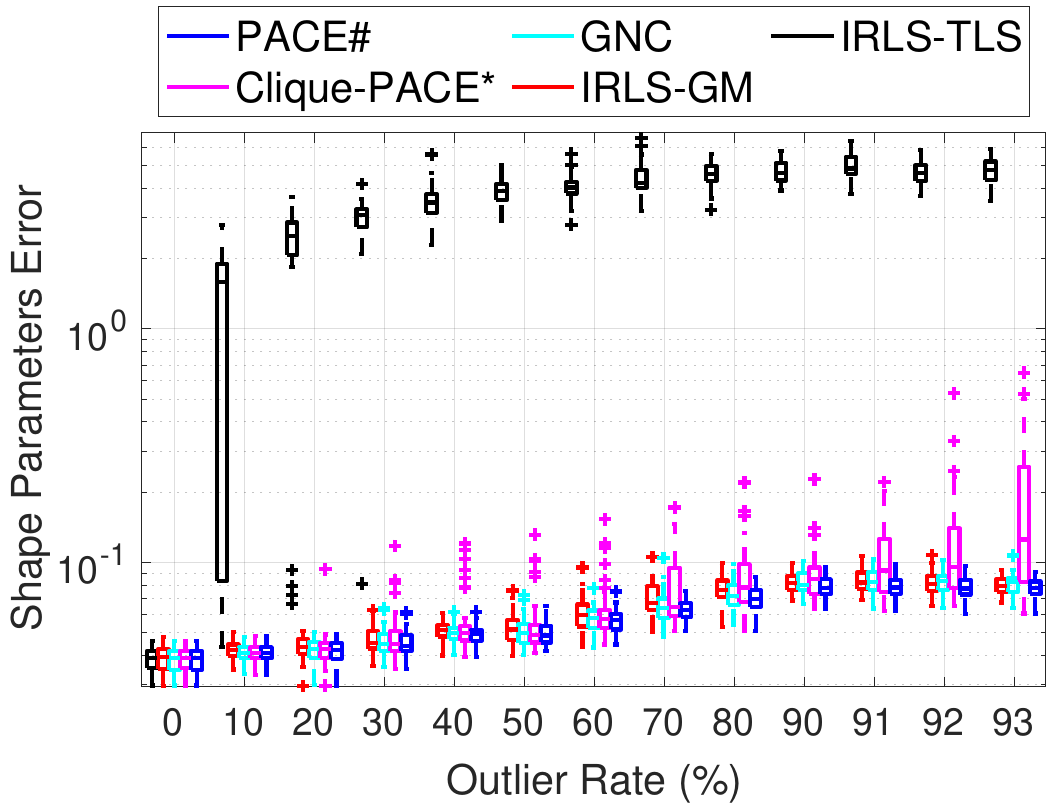}
			\end{minipage}
		&   \myhspace
			\begin{minipage}{\mpwfour}%
			\centering%
			\includegraphics[width=\columnwidth]{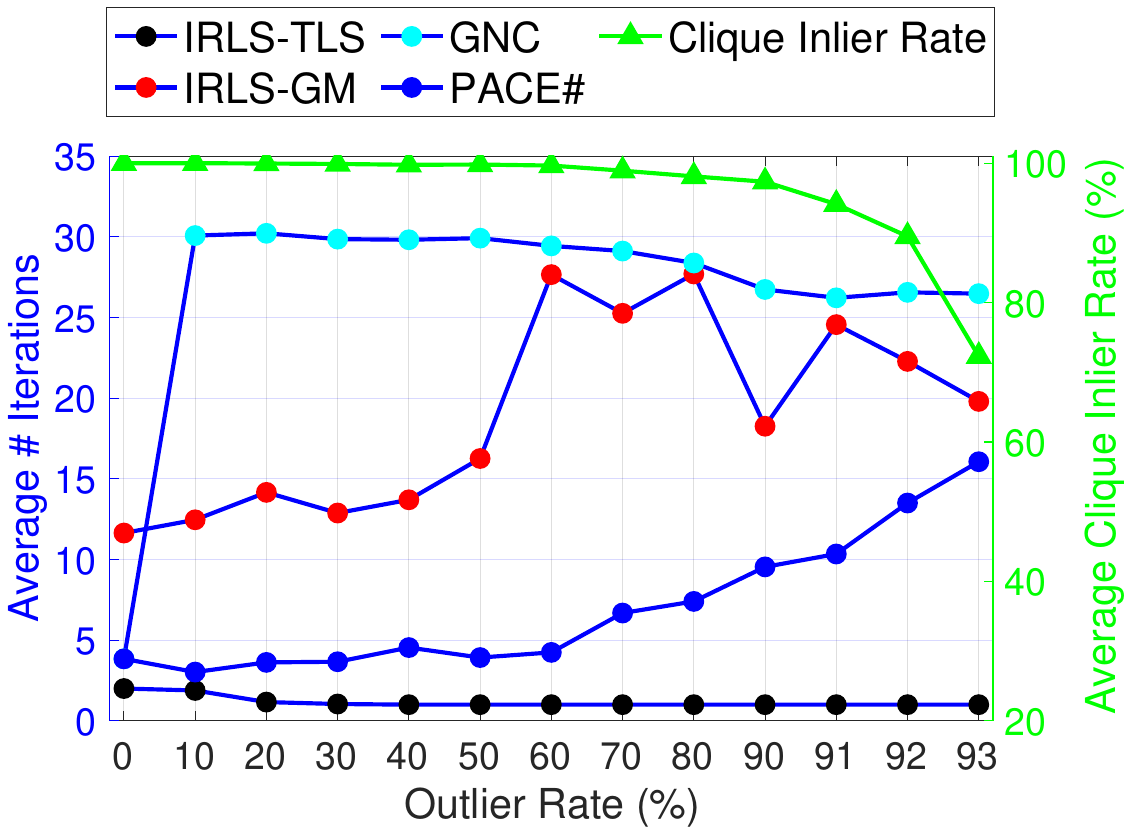}
			\end{minipage}
		\\
		\multicolumn{4}{c}{\smaller (b) Robustness of \nameRobust against increasing outliers on random simulated data: $N=100$, $K=50$, $r=0.1$. \vspace{1mm}}
		\\
		\myhspace \hspace{-3mm}
			\begin{minipage}{\mpwfour}%
			\centering%
			\includegraphics[width=\columnwidth]{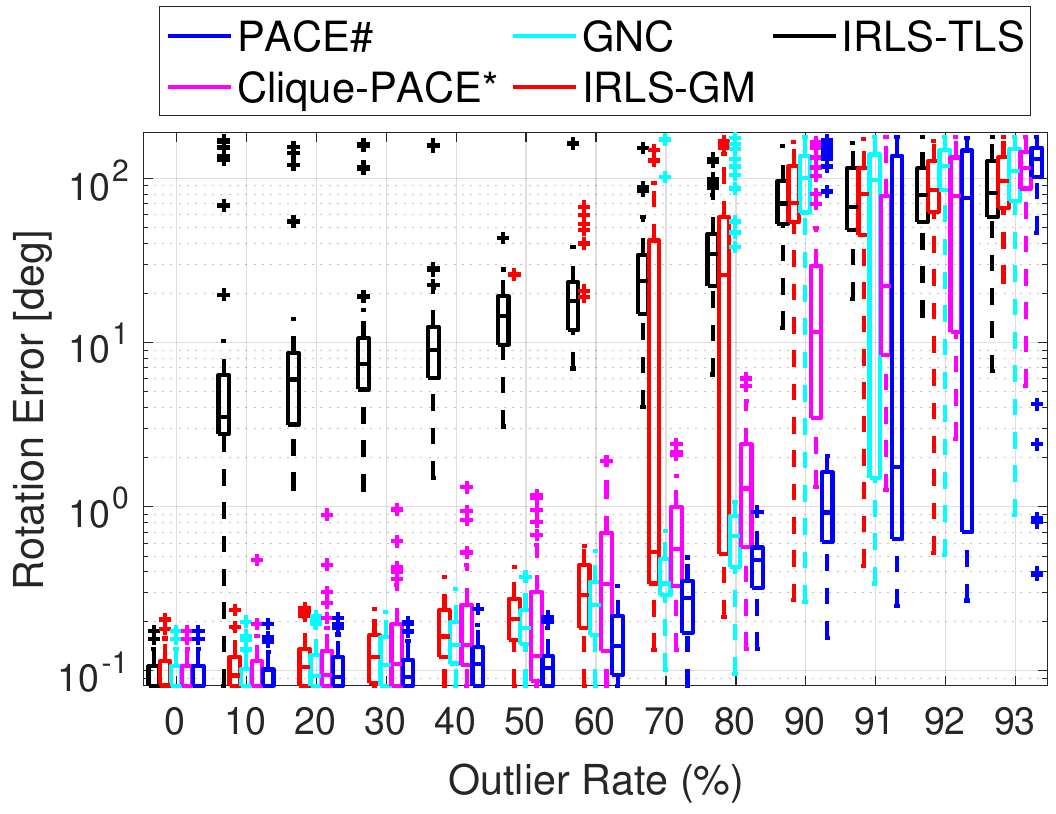}
			\end{minipage}
		&   \myhspace
			\begin{minipage}{\mpwfour}%
			\centering%
			\includegraphics[width=\columnwidth]{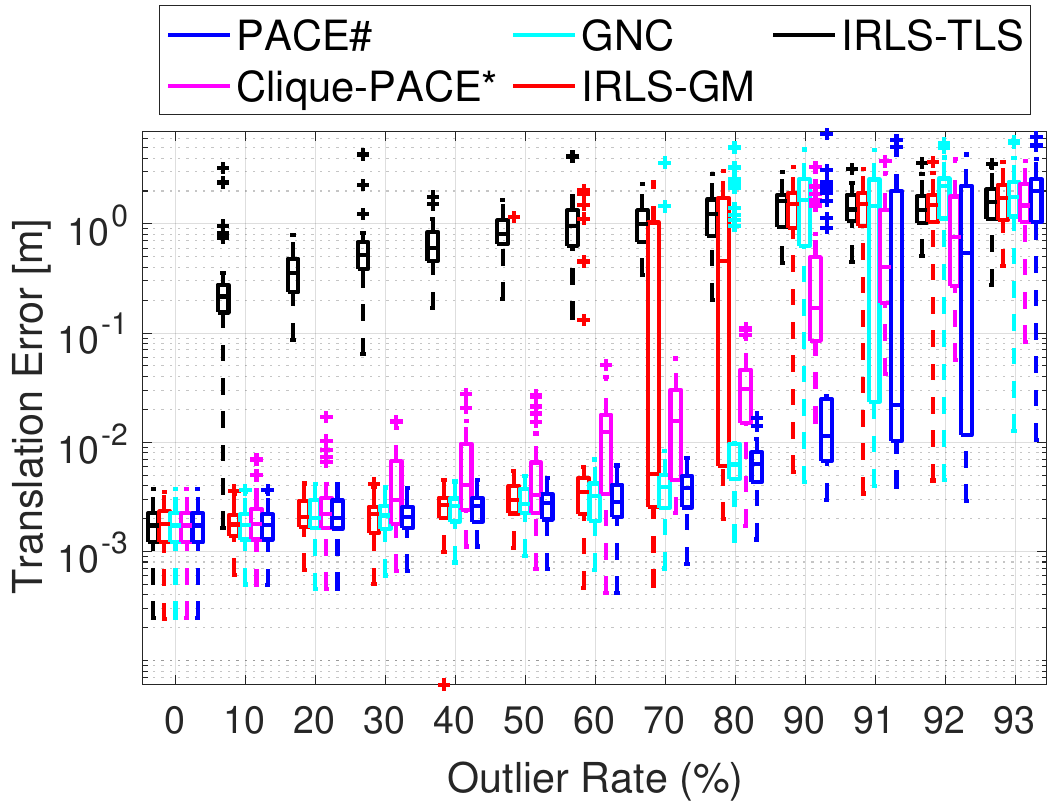}
			\end{minipage}
		&   \myhspace
			\begin{minipage}{\mpwfour}%
			\centering%
			\includegraphics[width=\columnwidth]{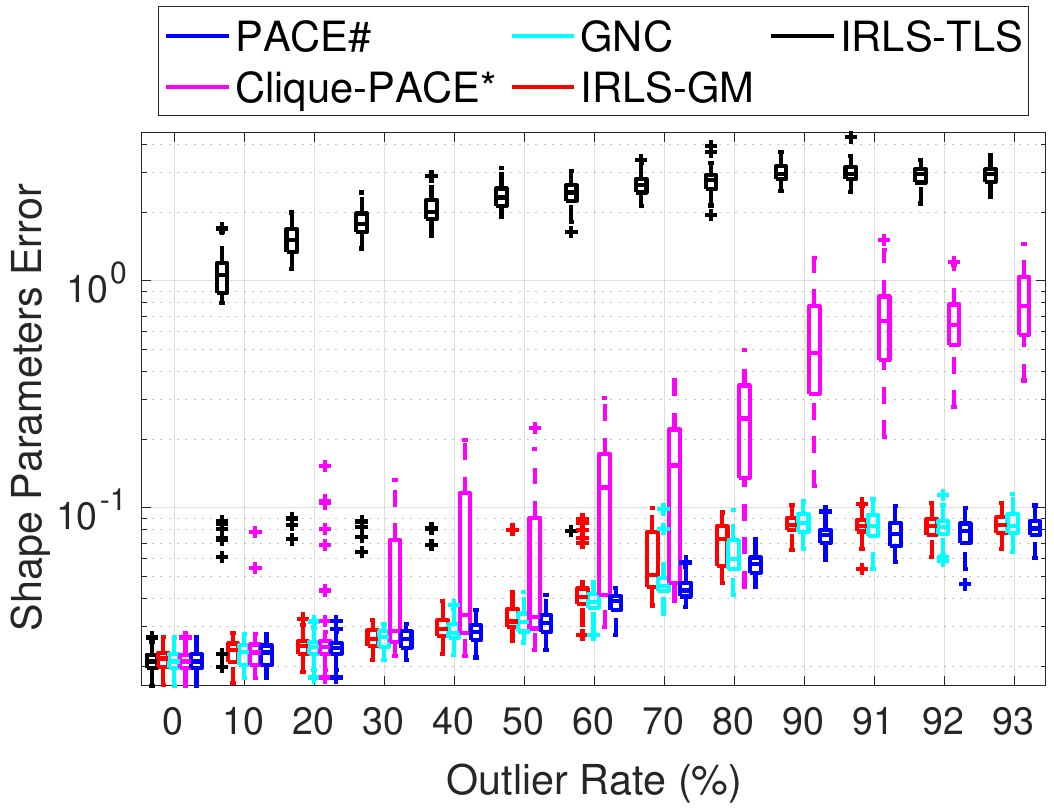}
			\end{minipage}
		&   \myhspace
			\begin{minipage}{\mpwfour}%
			\centering%
			\includegraphics[width=\columnwidth]{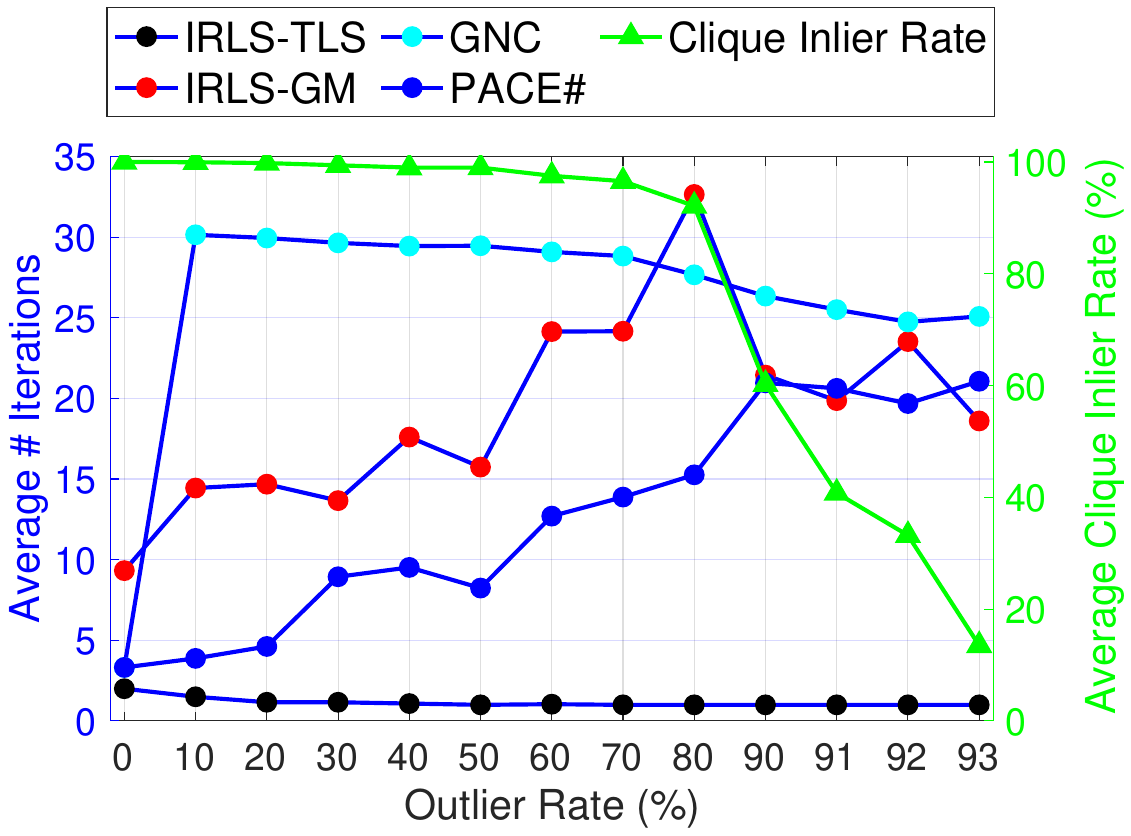}
			\end{minipage}
		\\
		\multicolumn{4}{c}{\smaller (c) Robustness of \nameRobust against increasing outliers on random simulated data: $N=100$, $K=50$, $r=0.2$. \vspace{-2mm}}
	\end{tabular}
	\end{minipage} 
	\caption{Performance of \nameRobust compared to baselines in simulated experiments with different 
	number of CAD models $\nrShapes$ and variation radius $r$.
	(a) The intra-class variation radius is increased to $r=0.2$. (b) The number of CAD models is increased to $K=50$. (c) $K=50$ and $r=0.2$. Each boxplot (and lineplot) reports statistics computed over  50 Monte Carlo runs.
	\label{fig:app-simulation-robust}} 
	\vspace{-4mm} 
	\end{center}
\end{figure*}

\newcommand{\mpwthree}{5.8cm}
\renewcommand{\myhspace}{\hspace{-3.5mm}}

\begin{figure*}[h]\ContinuedFloat
	\begin{center}{}
	\begin{minipage}{\textwidth}
	\begin{tabular}{cccc}%
		
& 
\myhspace
	\begin{minipage}{\mpwthree}%
	\centering%
	\includegraphics[width=\columnwidth]{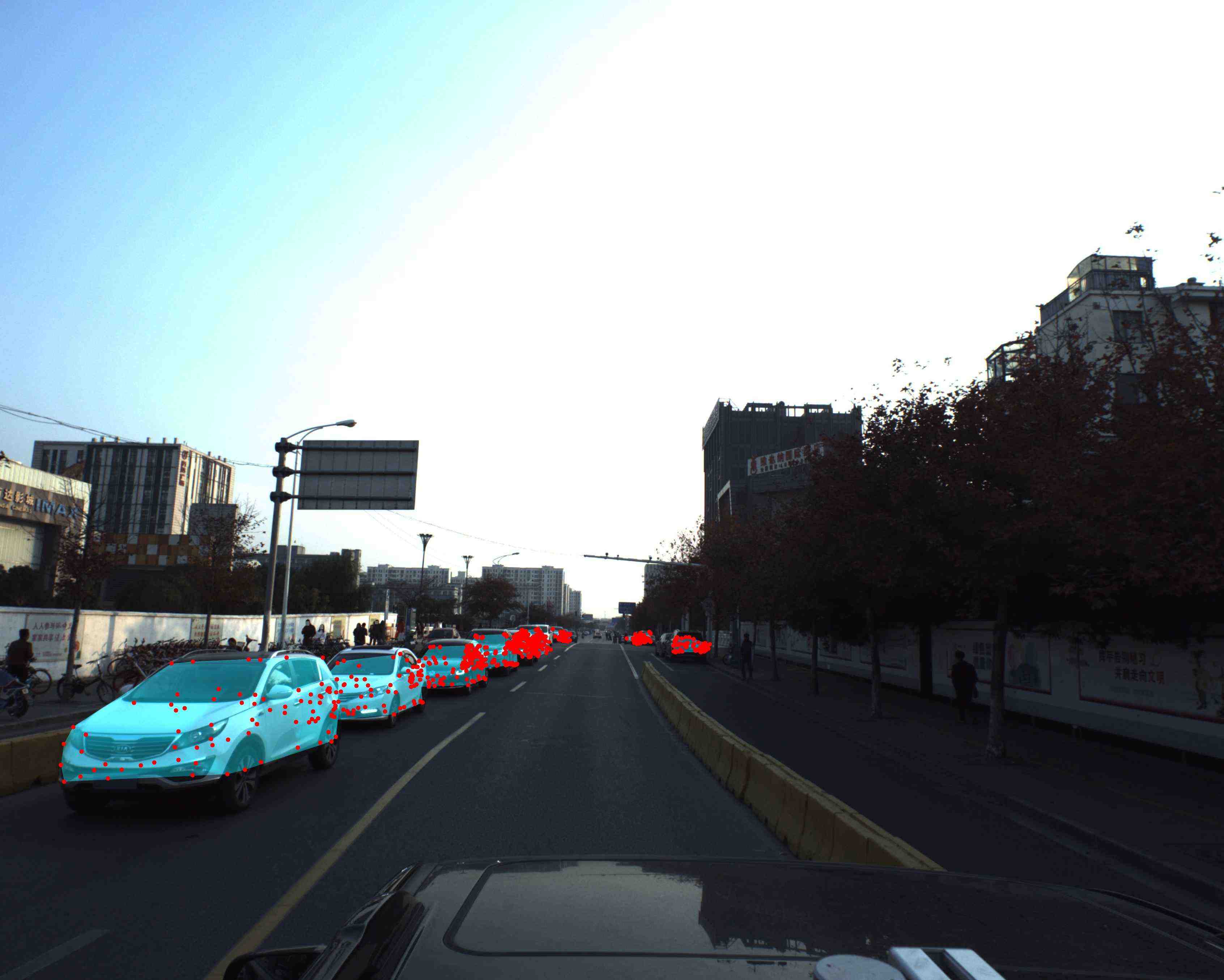} \\
	\vspace{1mm}
	\end{minipage}
& \myhspace
	\begin{minipage}{\mpwthree}%
	\centering%
	\includegraphics[width=\columnwidth]{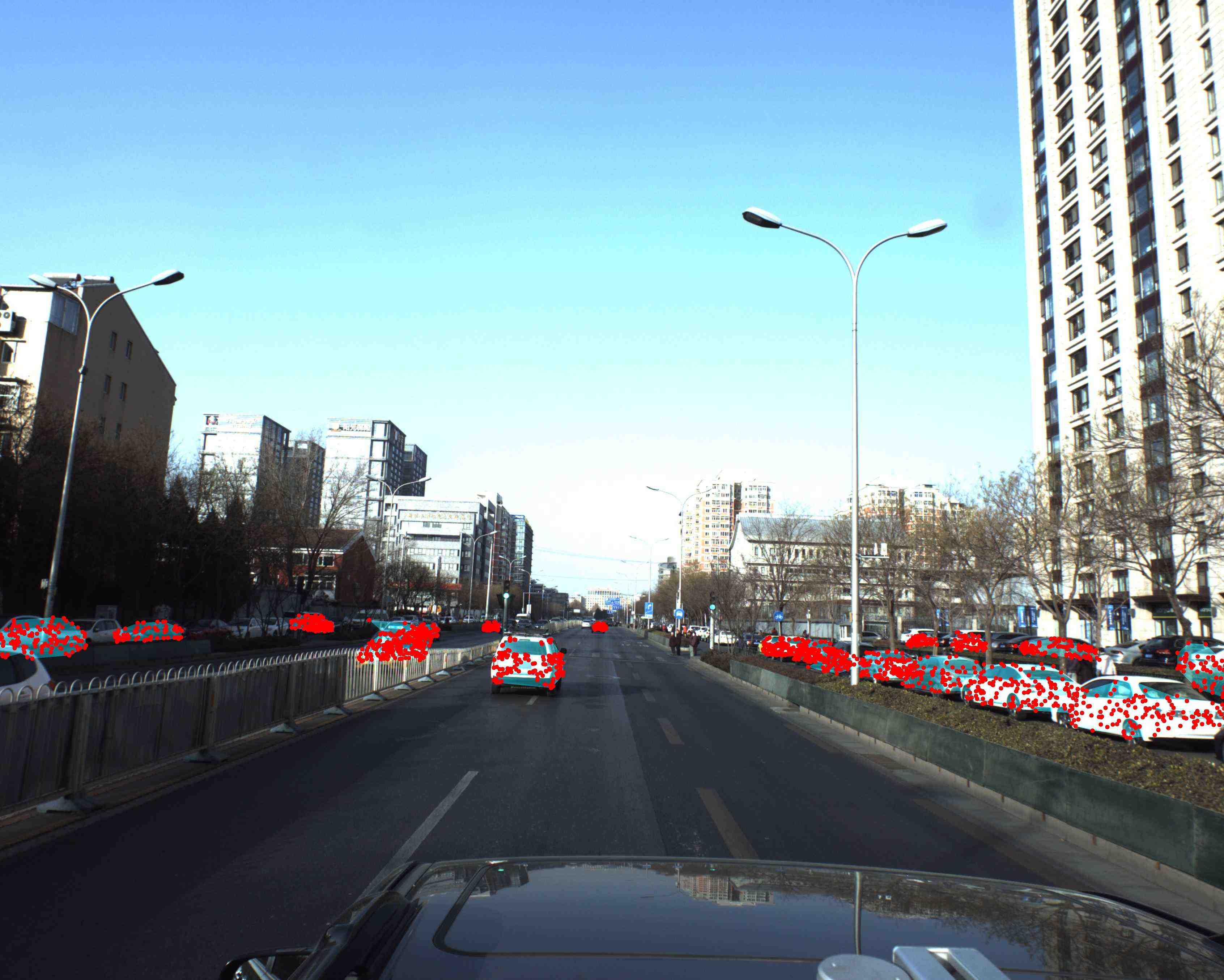} \\
	\vspace{1mm}
	\end{minipage}
& \myhspace
	\begin{minipage}{\mpwthree}%
	\centering%
	\includegraphics[width=\columnwidth]{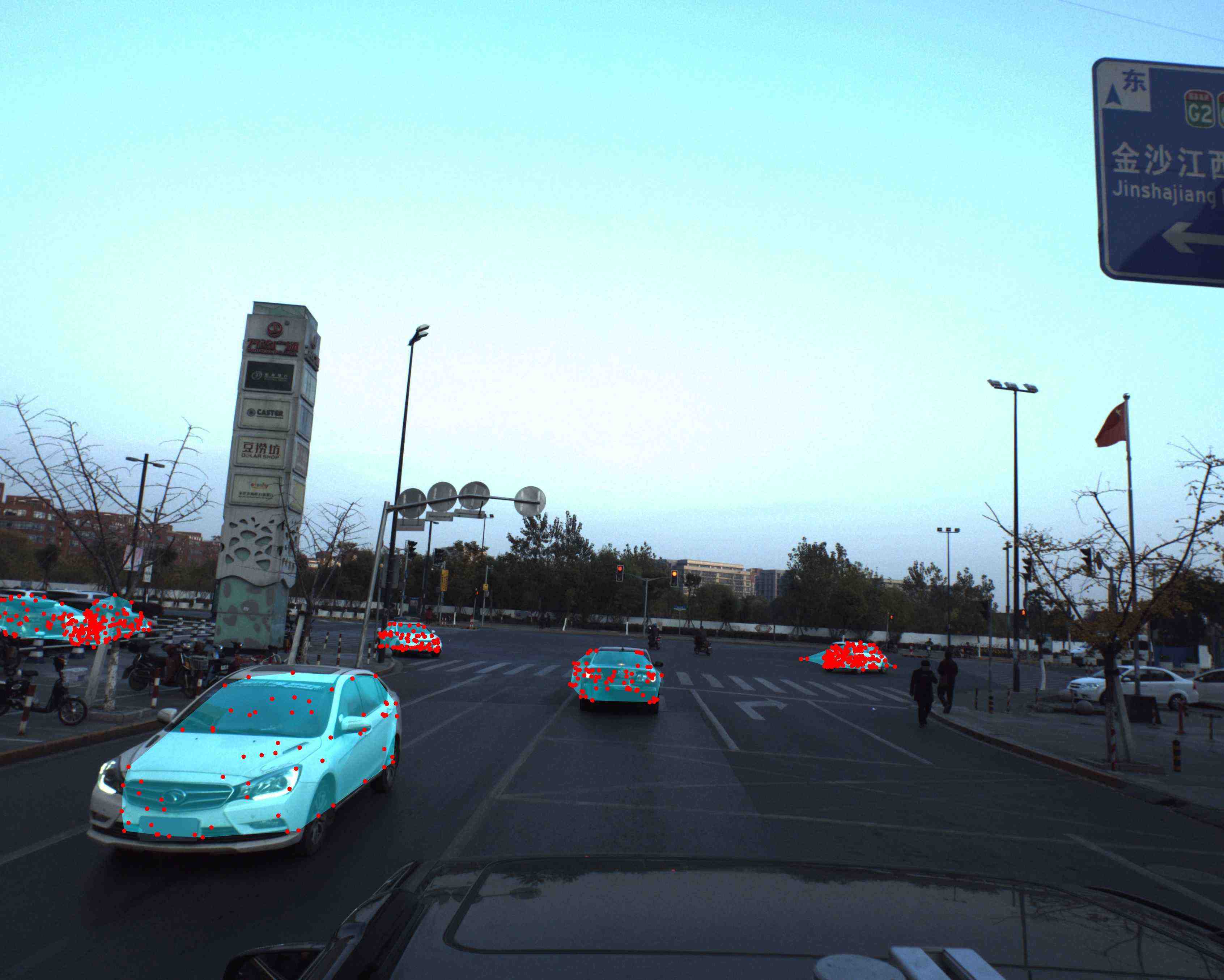} \\
	\vspace{1mm}
	\end{minipage} 
\\
&
\myhspace
	\begin{minipage}{\mpwthree}%
	\centering%
	\includegraphics[width=\columnwidth]{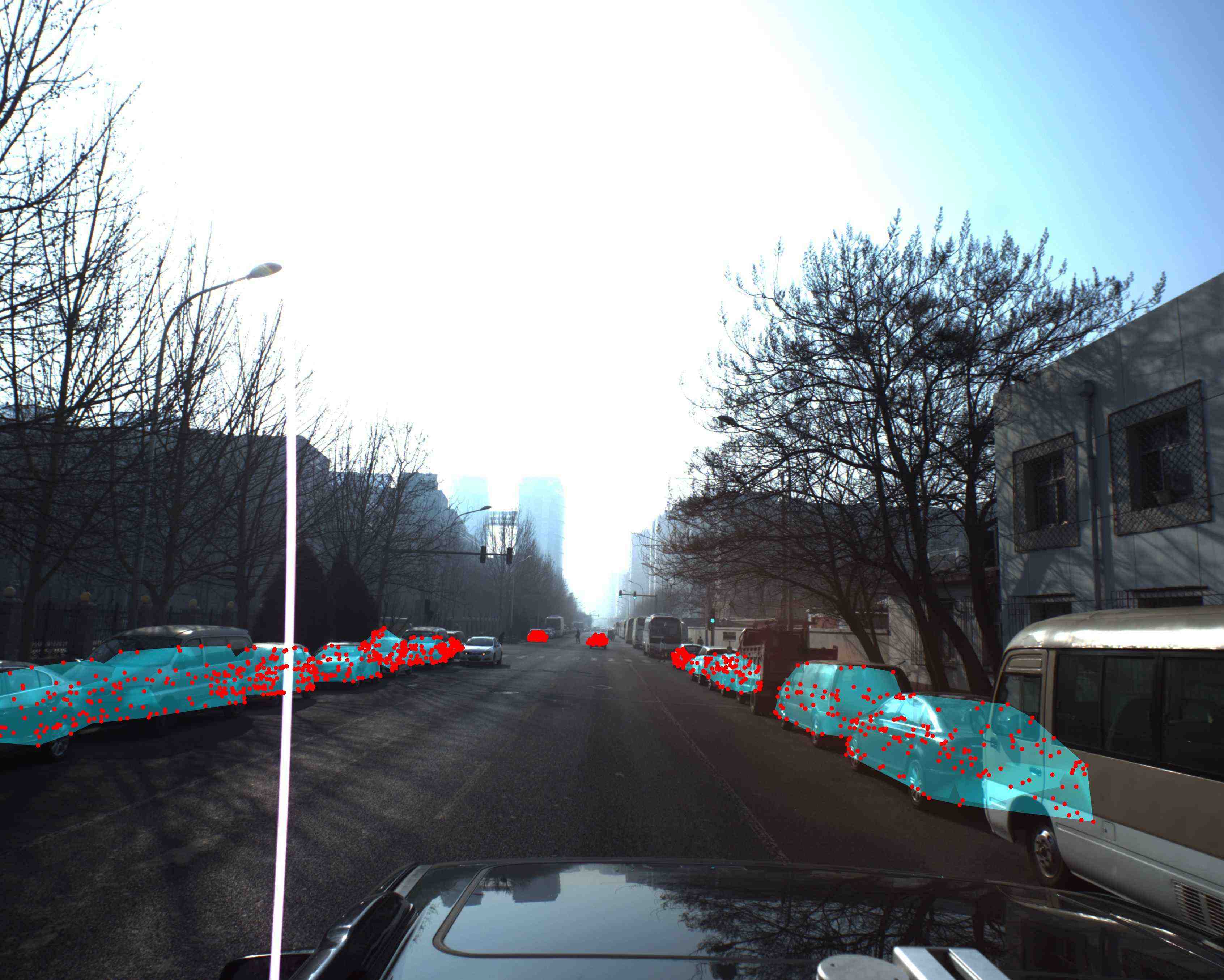} \\
	\vspace{1mm}
	\end{minipage}
& \myhspace
	\begin{minipage}{\mpwthree}%
	\centering%
	\includegraphics[width=\columnwidth]{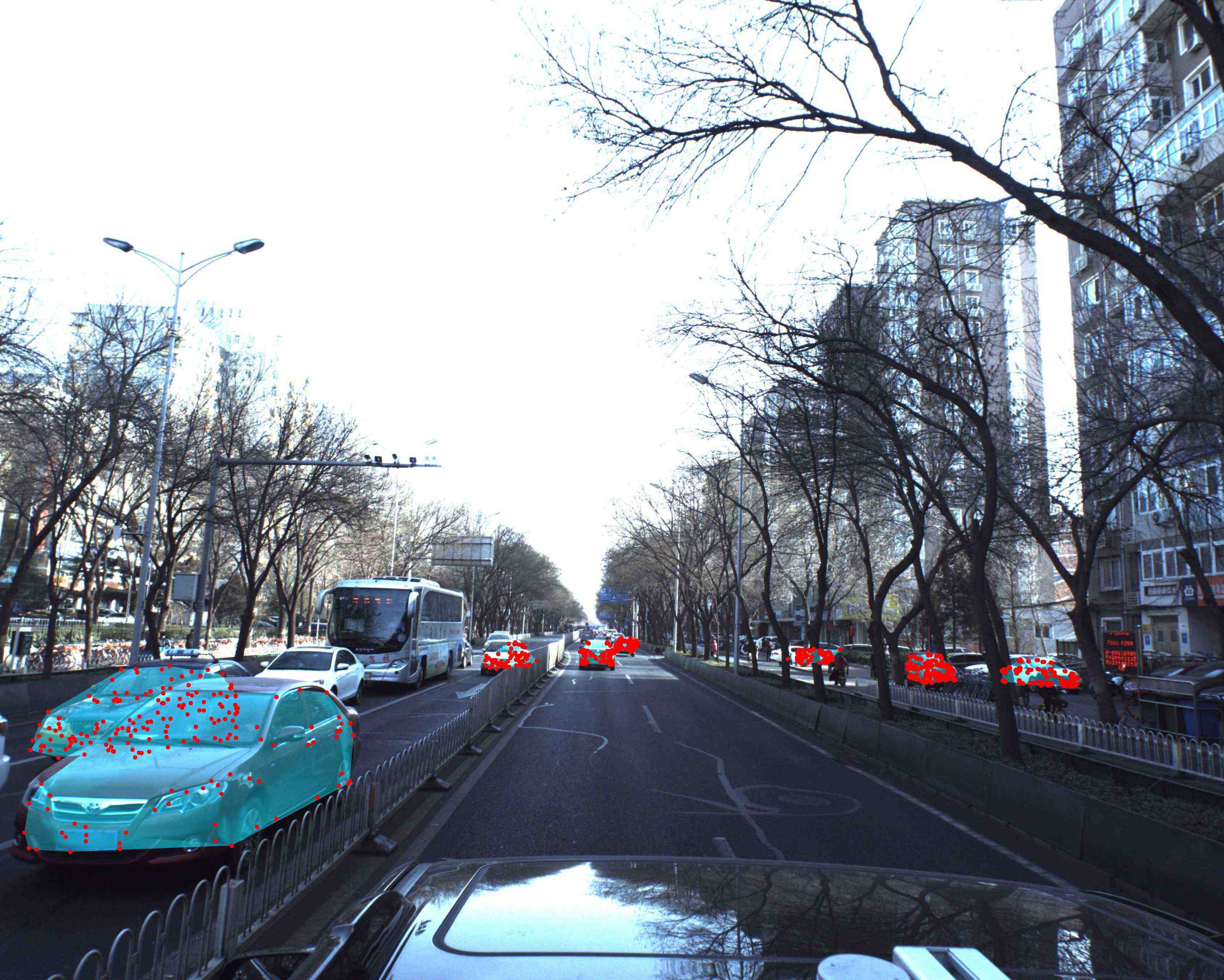} \\
	\vspace{1mm}
	\end{minipage}
& \myhspace
	\begin{minipage}{\mpwthree}%
	\centering%
	\includegraphics[width=\columnwidth]{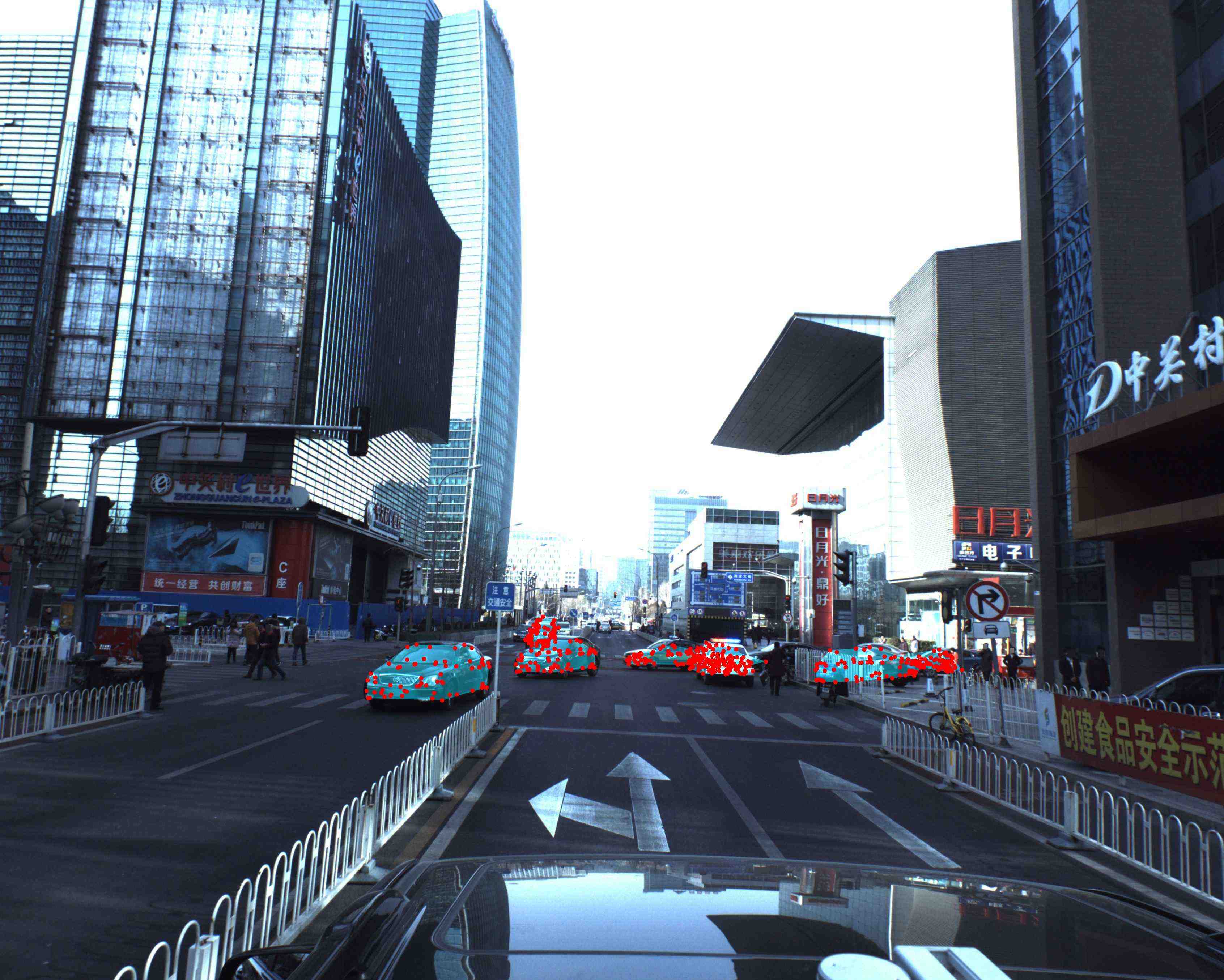} \\
	\vspace{1mm}
	\end{minipage} 
\\
&
\myhspace
	\begin{minipage}{\mpwthree}%
	\centering%
	\includegraphics[width=\columnwidth]{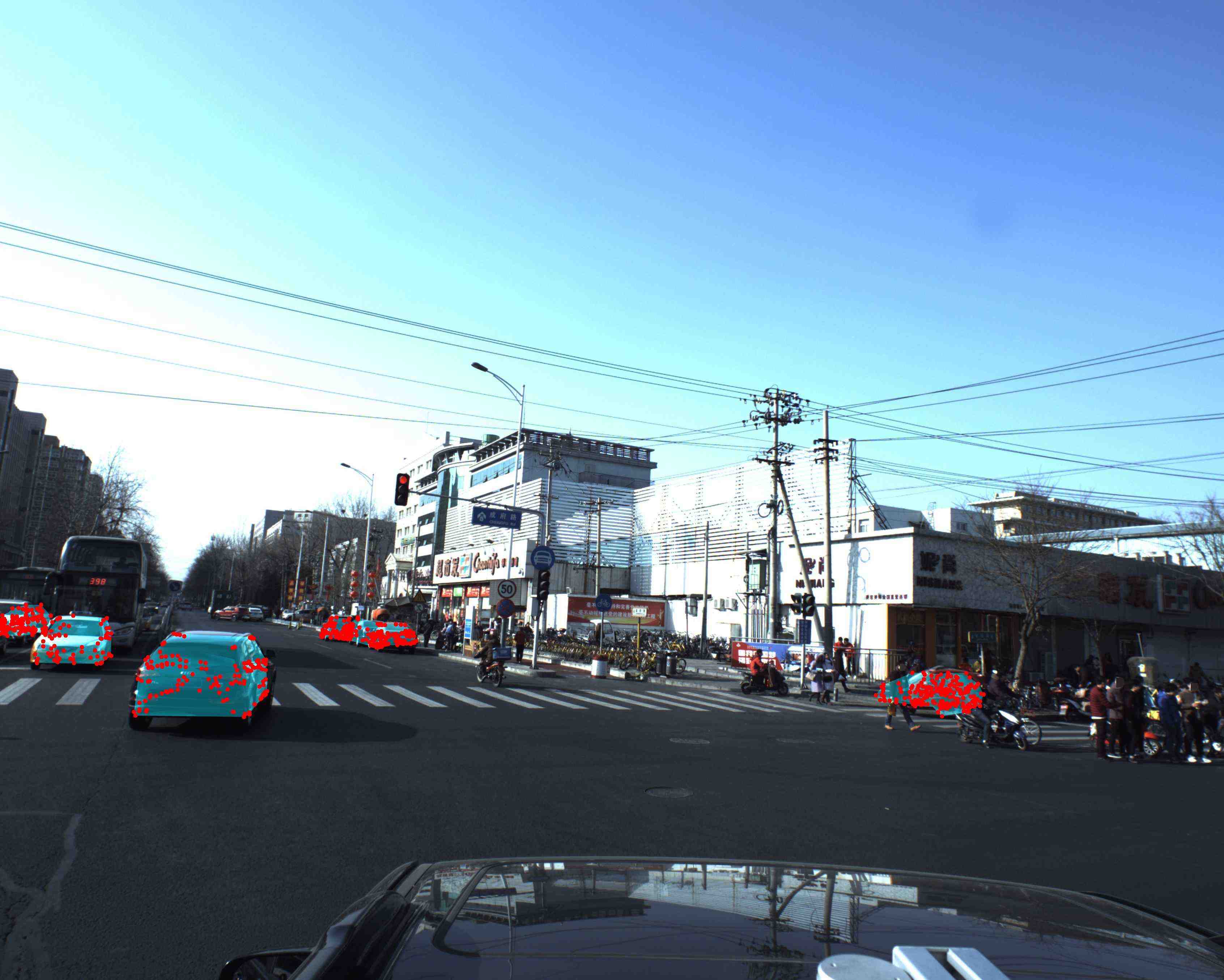} \\
	\vspace{1mm}
	\end{minipage}
& \myhspace
	\begin{minipage}{\mpwthree}%
	\centering%
	\includegraphics[width=\columnwidth]{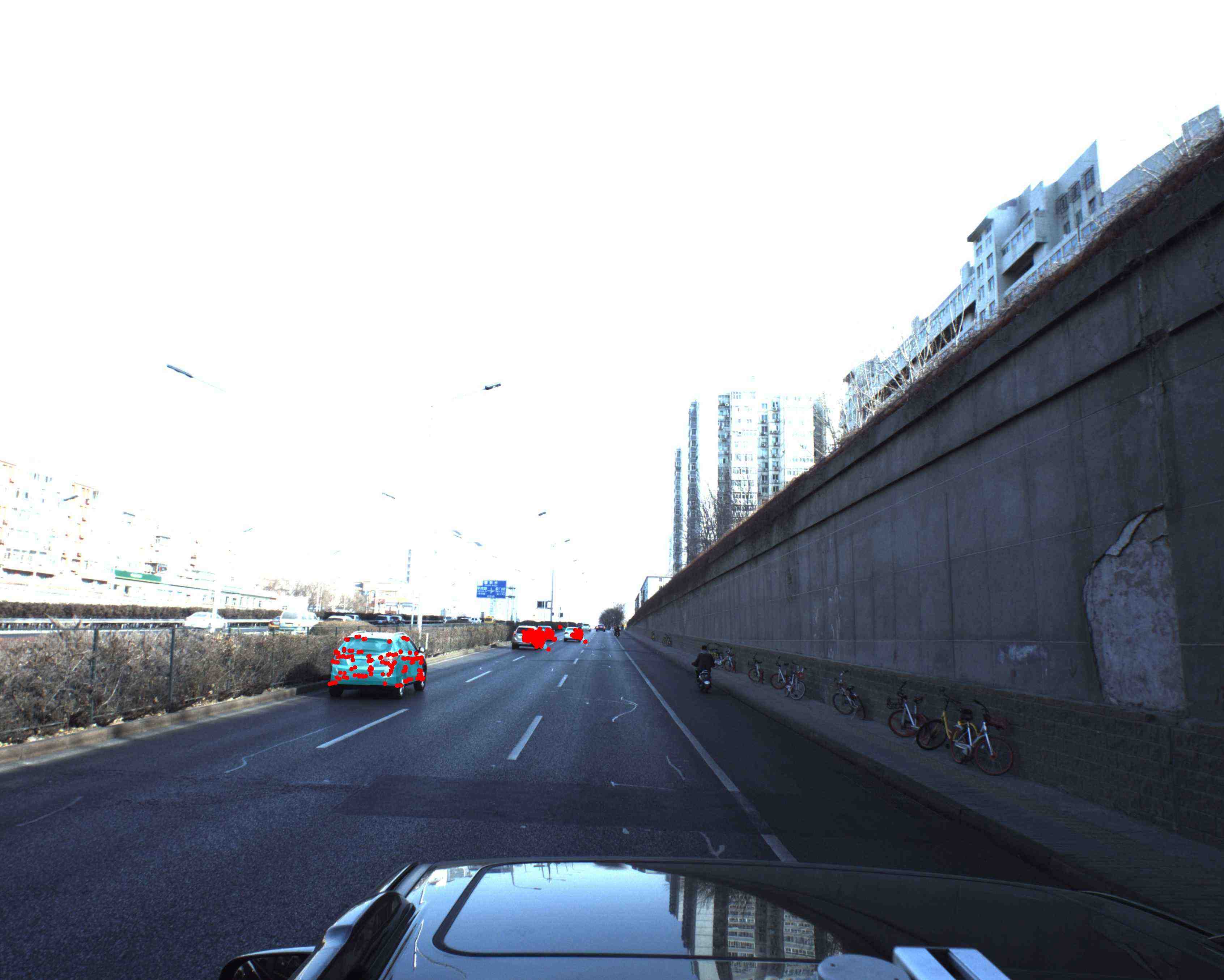} \\
	\vspace{1mm}
	\end{minipage}
& \myhspace
	\begin{minipage}{\mpwthree}%
	\centering%
	\includegraphics[width=\columnwidth]{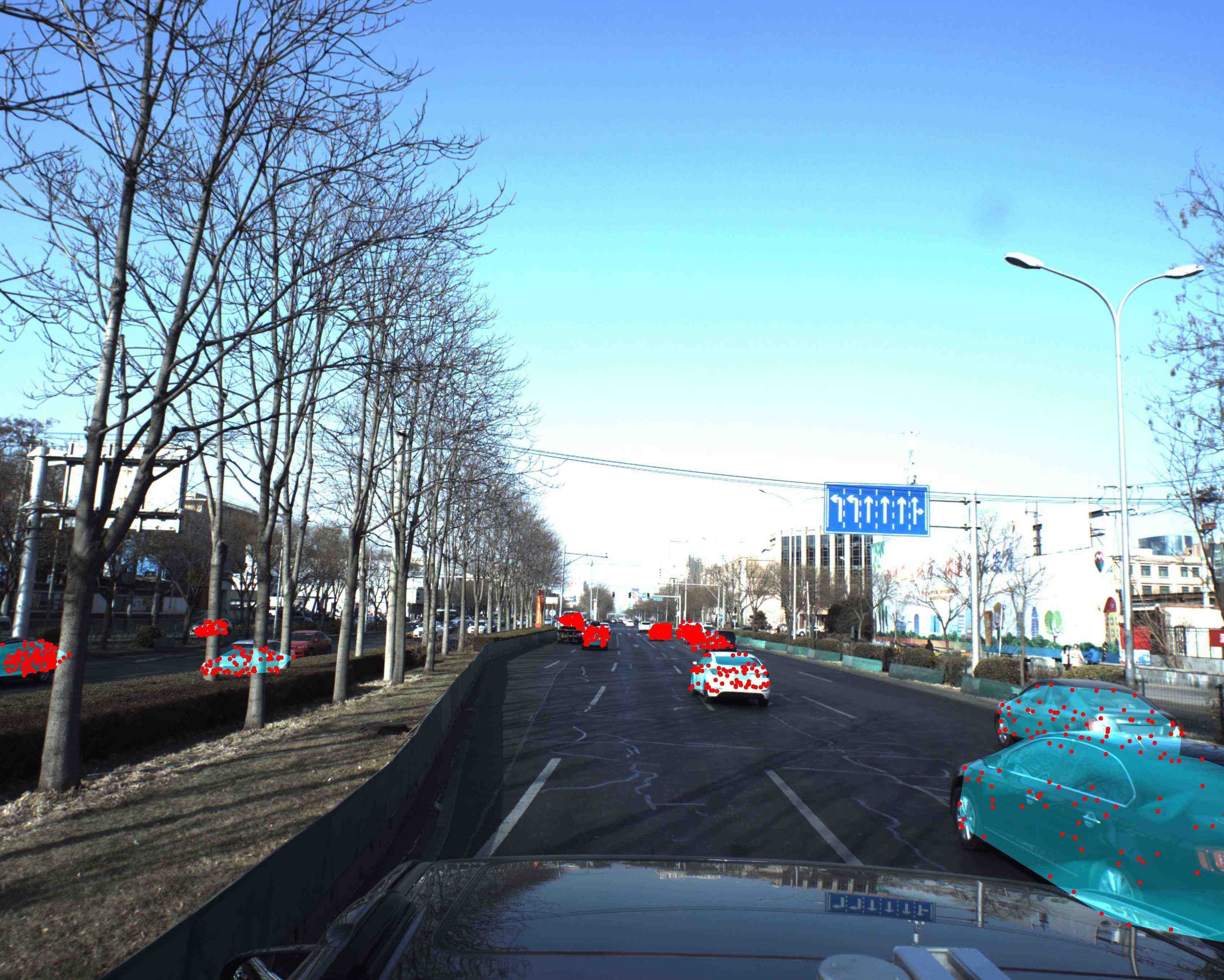} \\
	\vspace{1mm}
	\end{minipage} 
\\
&
\myhspace
	\begin{minipage}{\mpwthree}%
	\centering%
	\includegraphics[width=\columnwidth]{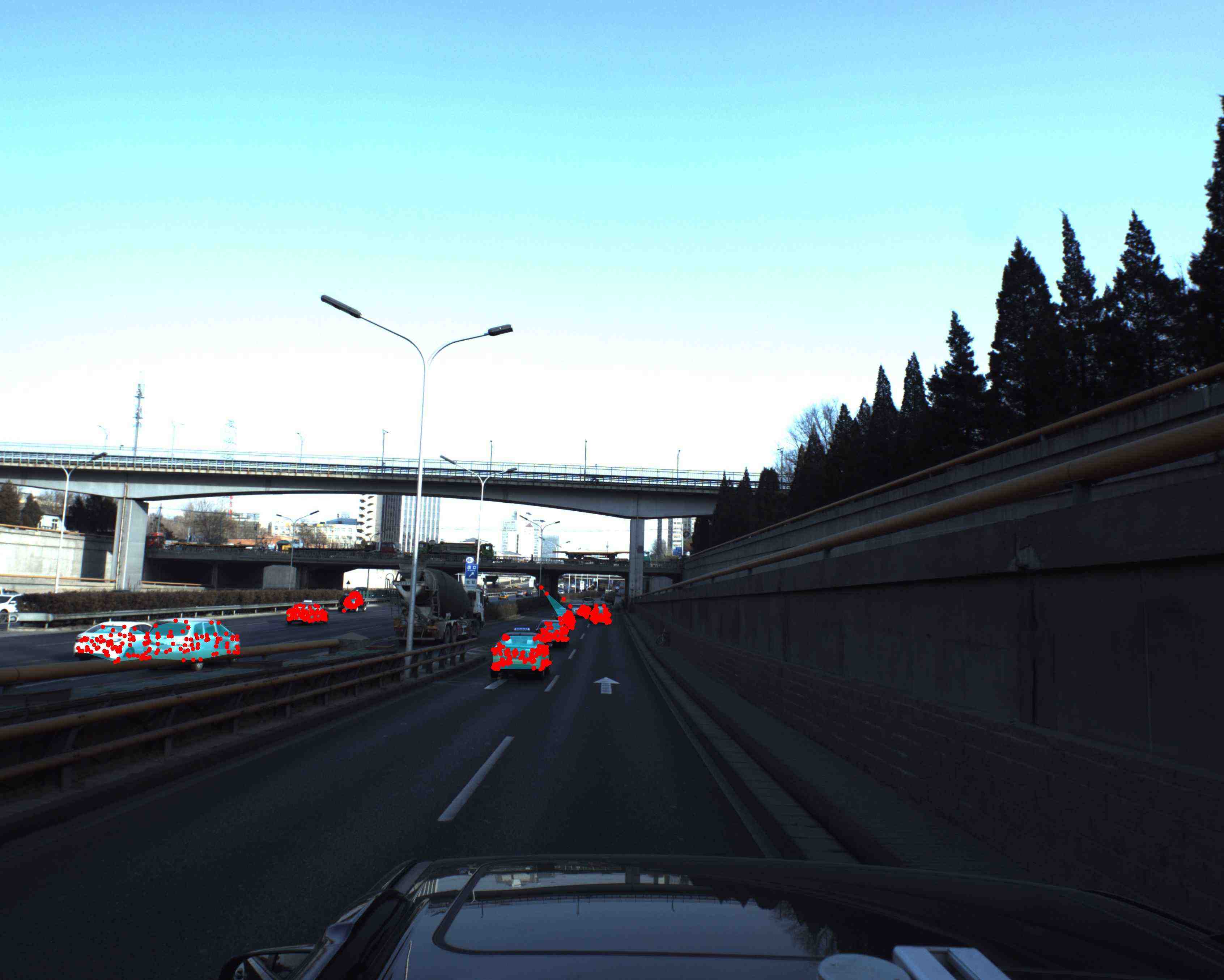} \\
	\vspace{1mm}
	\end{minipage}
& \myhspace
	\begin{minipage}{\mpwthree}%
	\centering%
	\includegraphics[width=\columnwidth]{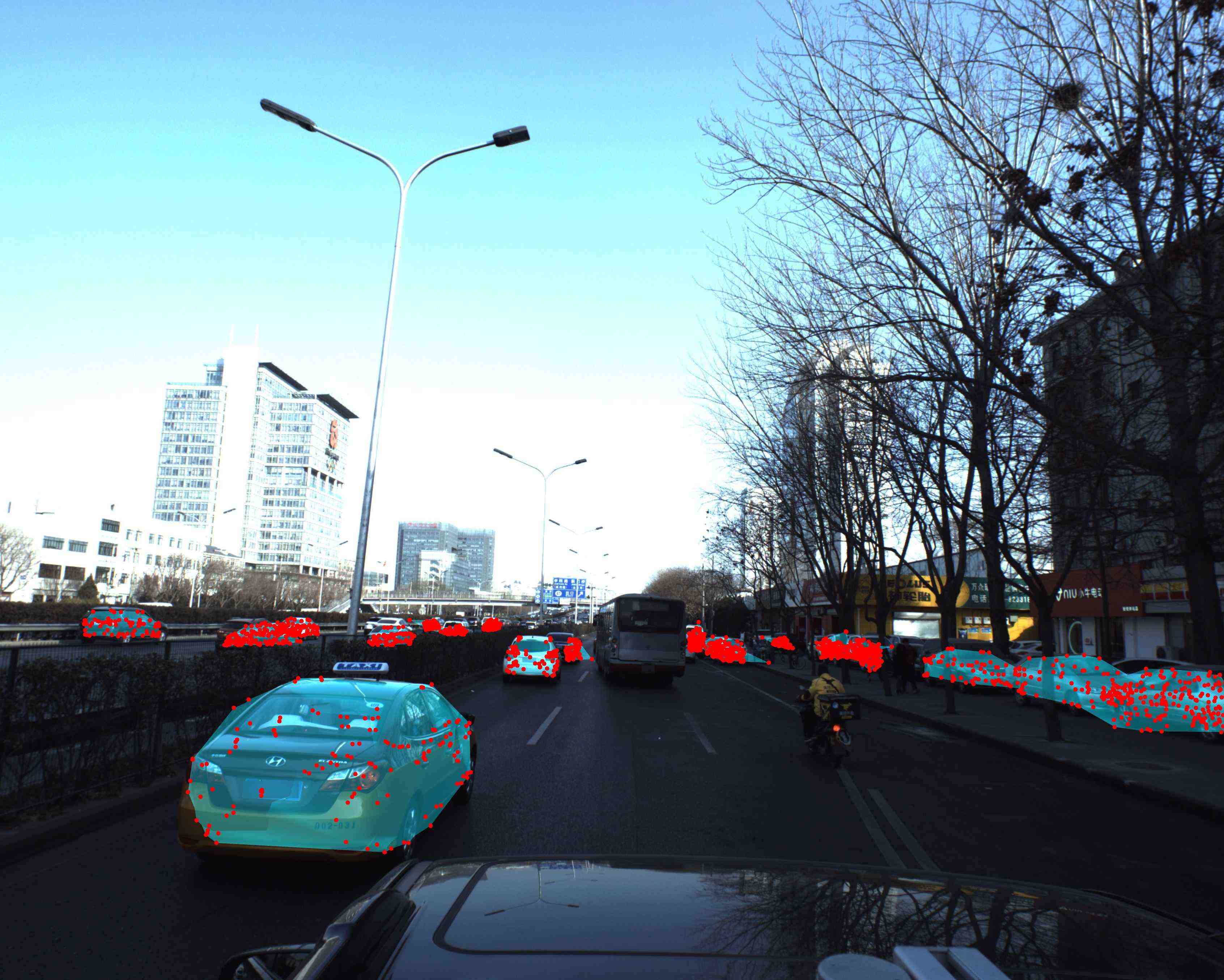} \\
	\vspace{1mm}
	\end{minipage}
& \myhspace
	\begin{minipage}{\mpwthree}%
	\centering%
	\includegraphics[width=\columnwidth]{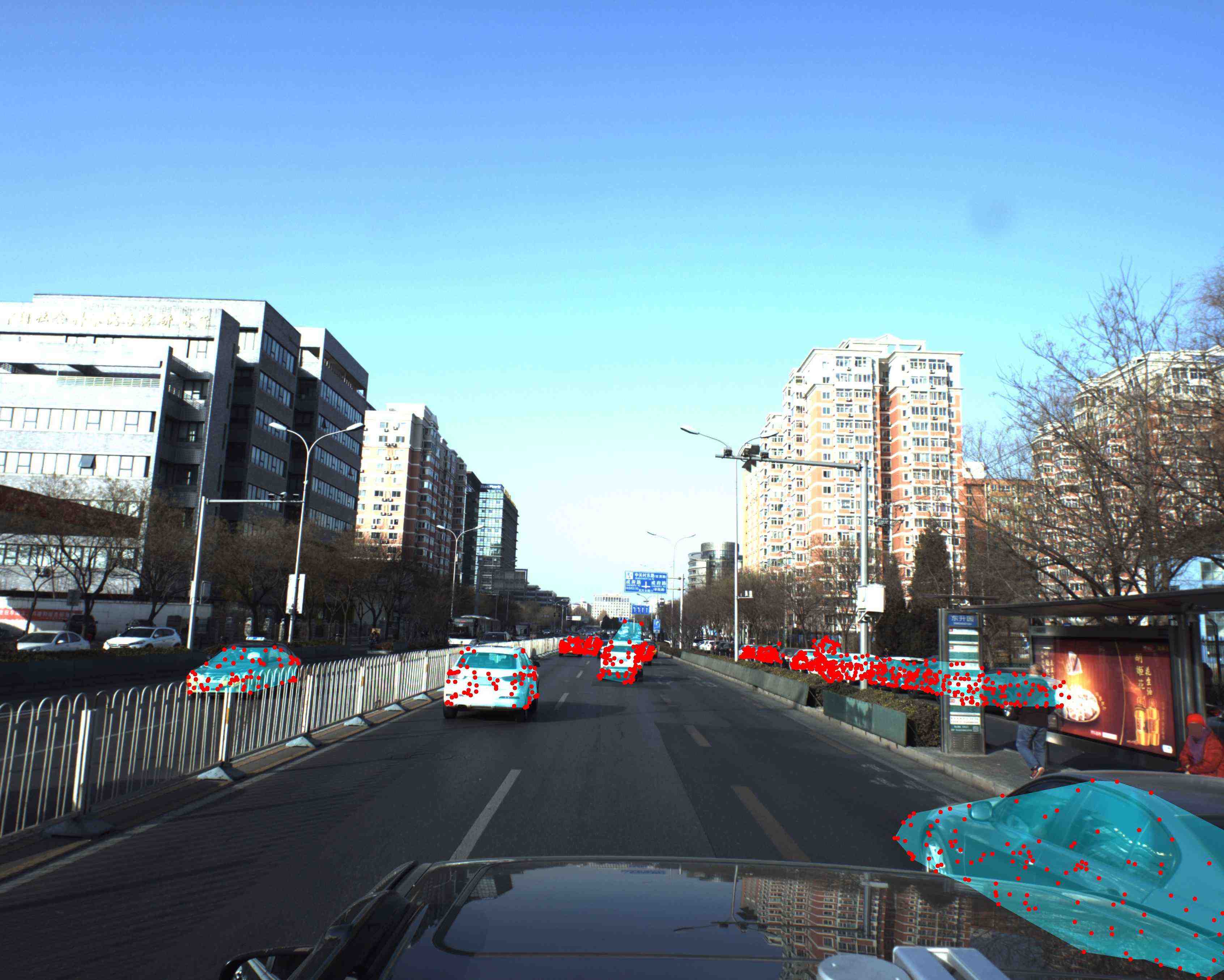} \\
	\vspace{1mm}
	\end{minipage} 
\\
\hline
\vspace{-2mm}
\\
	\myhspace \myhspace %
	\rotatebox{90}{\hspace{-7mm} \red{failures} } 
	&
\myhspace
	\begin{minipage}{\mpwthree}%
	\centering%
	\includegraphics[width=\columnwidth]{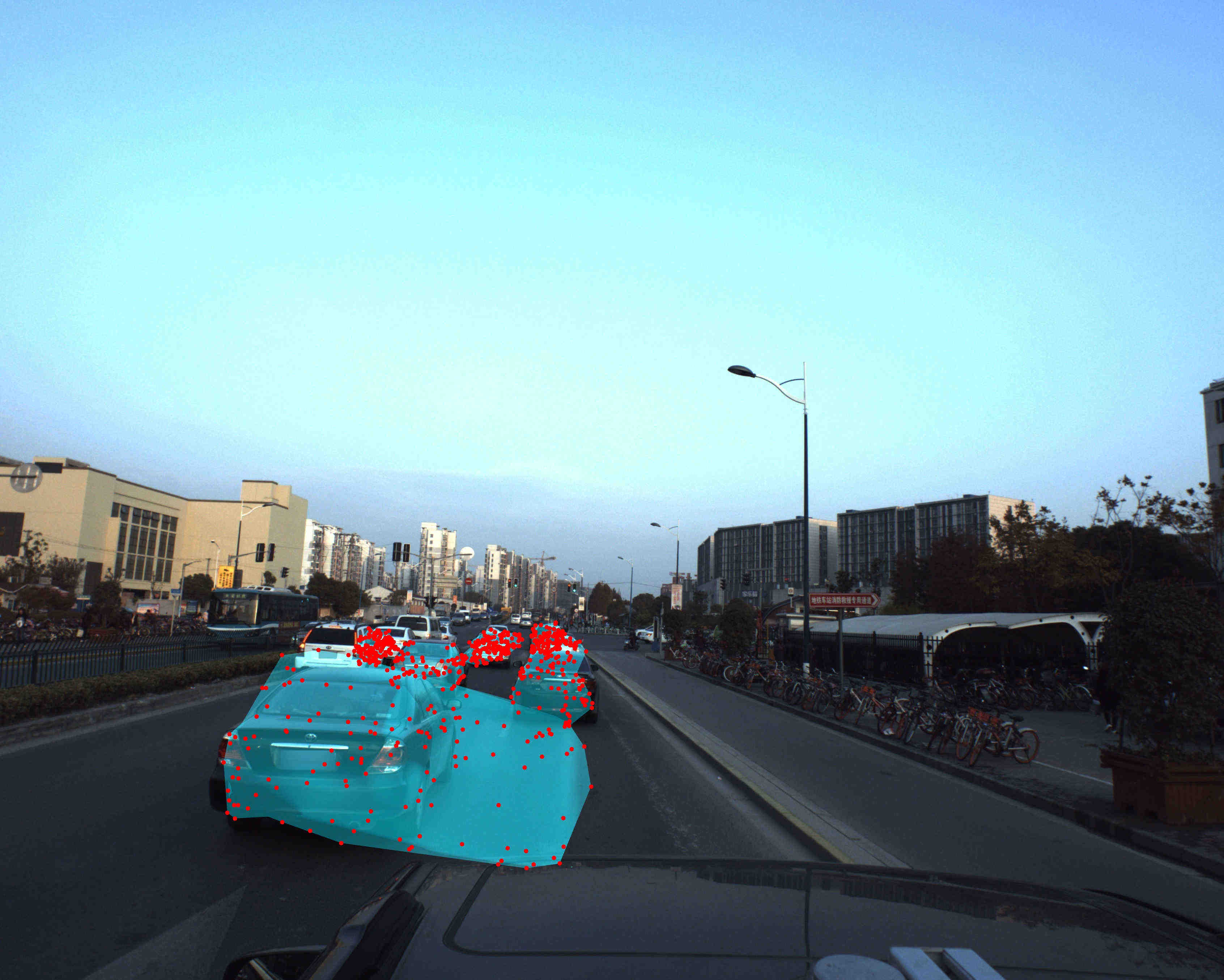} \\
	\vspace{1mm}
	\end{minipage}
& \myhspace
	\begin{minipage}{\mpwthree}%
	\centering%
	\includegraphics[width=\columnwidth]{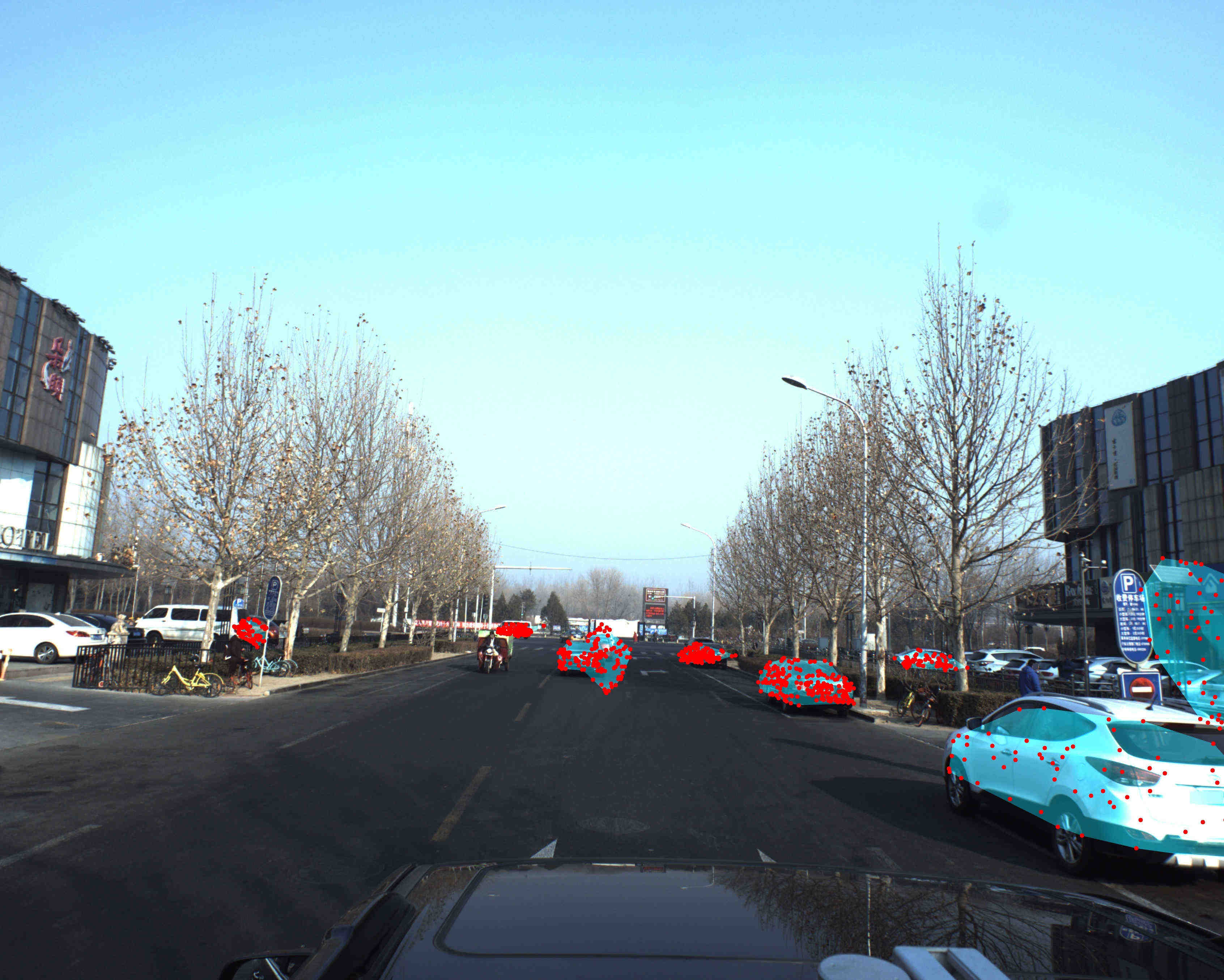} \\
	\vspace{1mm}
	\end{minipage}
& \myhspace
	\begin{minipage}{\mpwthree}%
	\centering%
	\includegraphics[width=\columnwidth]{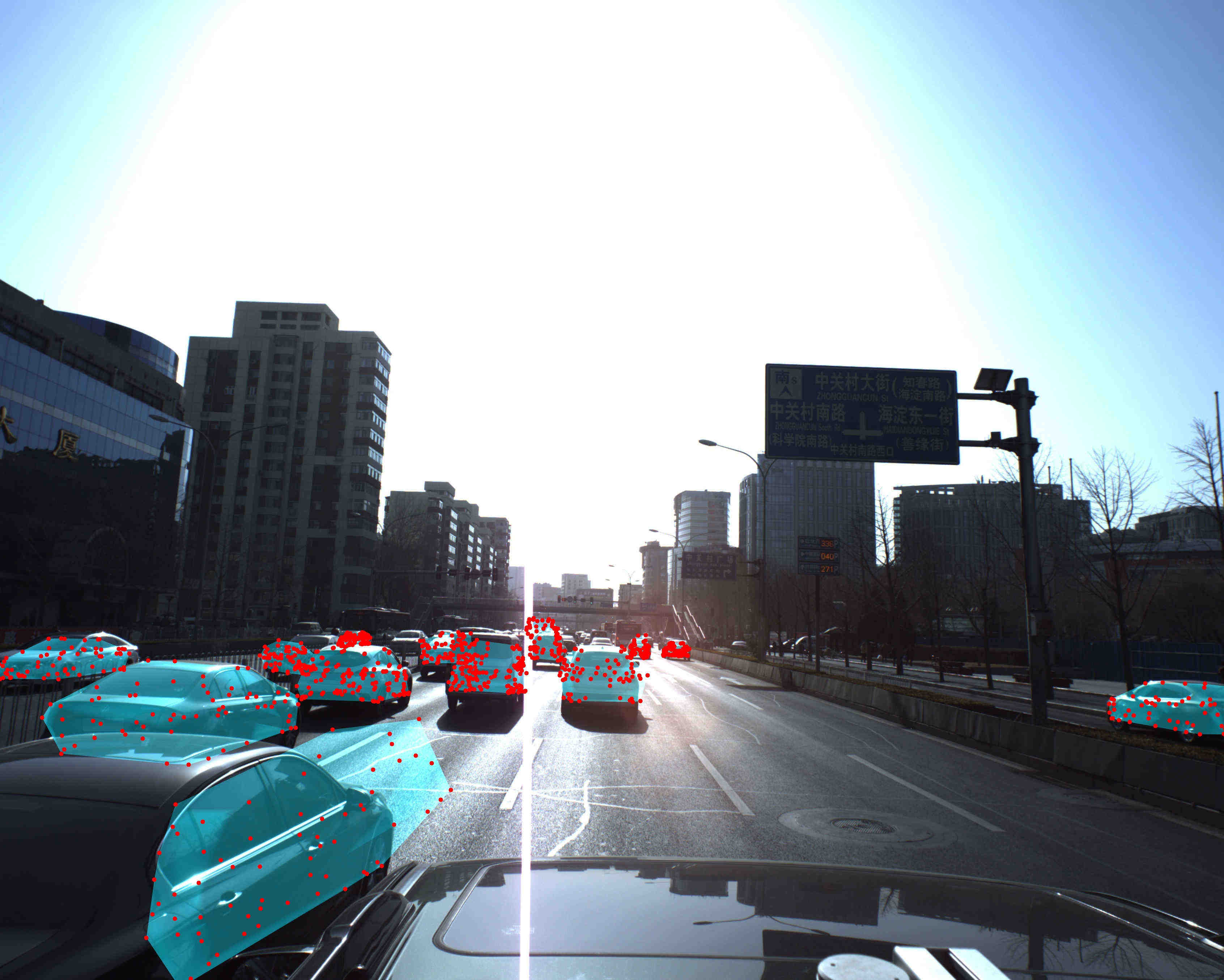} \\
	\vspace{1mm}
	\end{minipage} 
	\end{tabular}
	\end{minipage}
	\vspace{-2mm} 
	\caption{Qualitative results: overlay of estimated vehicle pose and shape on the images from the \apollo dataset. 
	The images are manually selected out of the 5277 images in the dataset to showcase successful vehicle 
	localization (top 4 rows) as well as failure cases (last row). }
	\end{center}
\end{figure*}
}{
}

\end{document}